\documentclass[10pt,journal,compsoc]{IEEEtran}
\ifCLASSOPTIONcompsoc
  \usepackage[nocompress]{cite}
\else
  \usepackage{cite}
\fi

\usepackage{epsfig}
\usepackage{epstopdf}
\usepackage{graphicx}
\usepackage{amsmath}
\usepackage{bm}
\usepackage{amssymb}
\usepackage{color, colortbl}
\usepackage{xcolor}
\usepackage[english]{babel}

\usepackage{algorithm} 
\usepackage{algpseudocode} 
\usepackage{booktabs}
\usepackage{verbatim}
\usepackage{gensymb}
\usepackage{pgffor}

\usepackage{xspace}
\usepackage{multirow}
\usepackage{enumitem}
\usepackage{rotating}
\usepackage{makecell}
\usepackage{ragged2e}
\usepackage[normalem]{ulem}
\usepackage[breaklinks=true,colorlinks,bookmarks=false]{hyperref}
\usepackage{pifont}

\RequirePackage{fontawesome}

\makeatletter
\let\MYcaption\@makecaption
\makeatother

\usepackage[font=footnotesize,labelformat=simple]{subcaption}

\makeatletter
\let\@makecaption\MYcaption
\makeatother

\makeatletter
\def\zapcolorreset{\let\reset@color\relax\ignorespaces}
\def\colorrows#1{\noalign{\aftergroup\zapcolorreset#1}\ignorespaces}
\makeatother

\makeatletter
\DeclareRobustCommand\onedot{\futurelet\@let@token\@onedot}
\def\@onedot{\ifx\@let@token.\else.\null\fi\xspace}

\def\ie{{i.e}\onedot} 
 
\def\etc{{etc}\onedot} 
\def\wrt{w.r.t\onedot} 
\def\etal{{et al}\onedot}
\makeatother

\def \rvcolor {black}
\newcommand{\rv}[1]{\textcolor{\rvcolor}{#1}}

\renewcommand{\paragraph}[1]{\textbf{#1}}

\newcommand{\cmark}{\ding{51}}%
\newcommand{\xmark}{\ding{55}}%

\newcommand{\shape}{\boldsymbol{\beta}}
\newcommand{\pose}{\boldsymbol{\theta}}
\newcommand{\reals}{\mathbb{R}}

\newif\ifarxiv
\arxivtrue

\begin{document}

\title{Recovering 3D Human Mesh from \\Monocular Images: A Survey}

\author{Yating Tian,
        Hongwen Zhang,
        Yebin Liu,~\IEEEmembership{Member,~IEEE},
        and Limin Wang,~\IEEEmembership{Member,~IEEE}
\IEEEcompsocitemizethanks{\IEEEcompsocthanksitem Yating Tian and Limin Wang are with the Department of Computer Science and Technology, Nanjing University, China.
Email: yatingtian@smail.nju.edu.cn; lmwang@nju.edu.cn
(Corresponding author: Limin Wang)
\IEEEcompsocthanksitem Hongwen Zhang and Yebin Liu are with the Department of Automation, Tsinghua University, China.
Email: zhanghongwen@mail.tsinghua.edu.cn; liuyebin@mail.tsinghua.edu.cn\\
(Yating Tian and Hongwen Zhang are co-first authors.)
}}

\markboth{~}%
{Tian \MakeLowercase{\textit{et al.}}: Recovering 3D Human Mesh from Monocular Images: A Survey}

\IEEEtitleabstractindextext{%
\begin{abstract}
\justifying
Estimating human pose and shape from monocular images is a long-standing problem in computer vision. Since the release of statistical body models, 3D human mesh recovery has been drawing broader attention. With the same goal of obtaining well-aligned and physically plausible mesh results, two paradigms have been developed to overcome challenges in the 2D-to-3D lifting process: i) an optimization-based paradigm, where different data terms and regularization terms are exploited as optimization objectives; and ii) a regression-based paradigm, where deep learning techniques are embraced to solve the problem in an end-to-end fashion. Meanwhile, continuous efforts are devoted to improving the quality of 3D mesh labels for a wide range of datasets. Though remarkable progress has been achieved in the past decade, the task is still challenging due to flexible body motions, diverse appearances, complex environments, and insufficient in-the-wild annotations. To the best of our knowledge, this is the first survey that focuses on the task of monocular 3D human mesh recovery. We start with the introduction of body models and then elaborate recovery frameworks and training objectives by providing in-depth analyses of their strengths and weaknesses. We also summarize datasets, evaluation metrics, and benchmark results. Open issues and future directions are discussed in the end, hoping to motivate researchers and facilitate their research in this area. A regularly updated project page can be found at \href{https://github.com/tinatiansjz/hmr-survey}{https://github.com/tinatiansjz/hmr-survey}.
\end{abstract}

}

\maketitle

\IEEEdisplaynontitleabstractindextext
\IEEEpeerreviewmaketitle

\IEEEraisesectionheading{\section{Introduction}\label{sec:introduction}}

\IEEEPARstart{U}nderstanding humans from monocular images is one of the fundamental tasks in computer vision.
Over the past two decades, the research community has focused on predicting 2D contents such as keypoints~\cite{cao2019openpose,fang2017rmpe,kreiss2021openpifpaf}, silhouettes~\cite{chen2018semantic}, and part segmentations~\cite{zhao2018understanding} from RGB images.
With these advances, researchers further seek to estimate human pose in 3D space~\cite{grauman2003inferring,agarwal2005recovering, martinez_2017_3dbaseline, pavlakos2017coarse, sun2018integral, mehta2020xnect,Weinzaepfel2020_dope}.
Although simple movements can be represented relatively clearly by 2D contents or a few sparse 3D joints, complex human behaviors require more descriptions of the human body with a finer granularity.
Moreover, it is critical to reason about body shape, contact, gesture, and expression since we interact with the world using our surface skin instead of unobserved joints.

\begin{figure}[t]
    \centering
	\includegraphics[width=\linewidth]{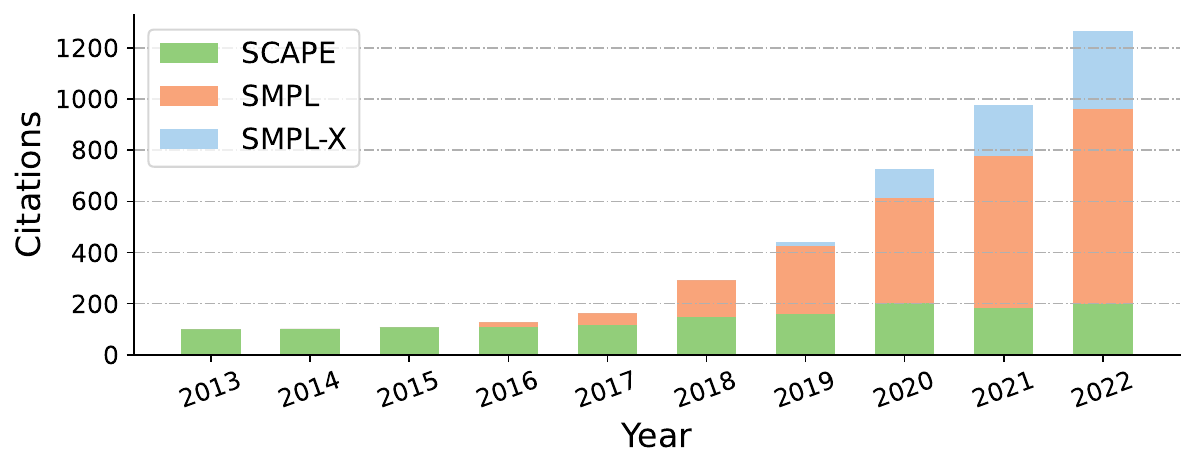}
	\vspace{-6mm}
	\caption{The annual citations of three representative 3D statistical human models, \ie, SCAPE~\cite{anguelov2005scape}, SMPL~\cite{loper2015smpl}, and SMPL-X~\cite{pavlakos2019expressive}.
	}
	\label{fig:citations}
	\vspace{-5mm}
\end{figure}

\begin{figure}[t]
    \centering
    \includegraphics[width=\linewidth]{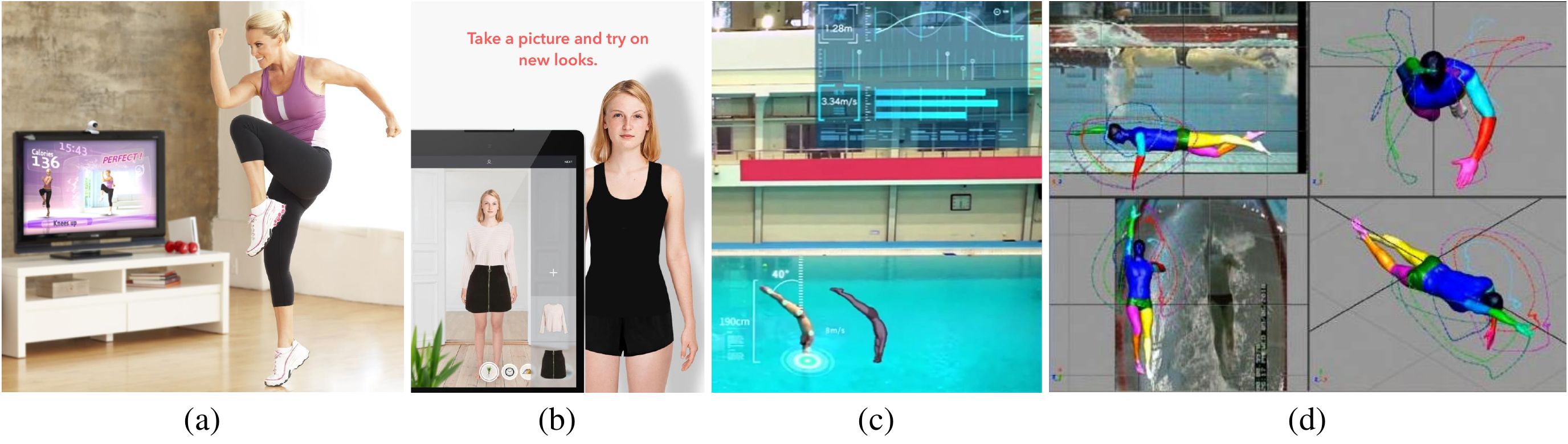}
    \vspace{-6mm}
    \caption{Real-world applications of human mesh recovery: (a) a video game for fitness \href{https://www.spokesman.com/stories/2009/dec/29/video-games-turn-attention-to-fitness/}{\faExternalLink}; (b) virtual try-on \href{https://www.reactivereality.com/}{\faExternalLink}; (c) a 3D+AI coaching system for diving \href{https://www.youtube.com/watch?v=PdzfC6OqIkg}{\faExternalLink}; (d) dynamic simulations during swimming \href{https://research.csiro.au/digitalhuman/}{\faExternalLink}.}
	\vspace{-5mm}
	\label{fig:application}
\end{figure}

In recent years, the community has shifted its interests towards 3D mesh recovery of human bodies~\cite{bogo2016keep,huang2017towards,zanfir2018monocular,kanazawa2018end,pavlakos2018learning,omran2018neural,zhang2021pymaf,kocabas2021pare,joo2021exemplar} along with expressive face and hands~\cite{pavlakos2019expressive, choutas2020monocular, feng2021collaborative, moon2022accurate,zhang2021lightweight}.
This trend is inseparable from the success of statistical human models. As shown in Fig.~\ref{fig:citations},
since the release of the SMPL model~\cite{loper2015smpl} in 2015 and the SMPL-X model~\cite{pavlakos2019expressive} in 2019, they have gained increasing interest as their annual citations grow rapidly year by year.
The recovery of human body meshes plays a key role in facilitating the downstream tasks such as clothed human reconstruction~\cite{yu2018doublefusion,zheng2021pamir,zheng2021deepmulticap,li2021image,feng2022fof,xiu2022icon,xiu2023econ}, rendering~\cite{peng2021neural,hu2022hvtr}, and avatar modeling~\cite{huang2020arch,ma2021power,zheng2022structured,zheng2023avatarrex}.
It is also involved in widespread applications such as VR/AR content creation, virtual try-on, and computer-assistant coaching, as depicted in Fig.~\ref{fig:application}.

Recovering 3D human mesh from monocular images is quite challenging, owing to the issues such as inherent ambiguities in lifting 2D observations to 3D space, flexible body kinematic structures, complex intersections with the environments, and insufficient annotated 3D data.
To address these issues, two different paradigms have been investigated in this field for the recovery of well-aligned
and physically plausible results.
Following the optimization-based paradigm~\cite{bogo2016keep,lassner2017unite,zanfir2018monocular}, methods explicitly fit body models to 2D observations in an iterative manner. Various data terms and regularization terms are explored as optimization objectives.
Alternatively, the regression-based paradigm~\cite{tung2017self, kanazawa2018end, pavlakos2018learning, omran2018neural, kolotouros2019learning} takes advantage of the powerful nonlinear mapping capability of neural networks and directly predicts model parameters from raw image pixels. Different network architectures and regression targets are designed to achieve better performances. 
Meanwhile, significant efforts have also been devoted to creating various datasets to facilitate the research of this task. Despite the remarkable progress achieved in recent years, the research community still faces challenges toward the ultimate goal of robust, accurate, and efficient human mesh recovery.

\subsection{Scope}
This survey mainly focuses on approaches to monocular 3D human mesh recovery (a.k.a. 3D human pose and shape estimation) in the deep learning era. Single RGB images and monocular RGB videos (or ``monocular images" to refer to them collectively) as input are considered. 
In addition to single-person recovery from monocular images, we also take multi-person recovery into account.
As for the reconstruction target, statistical human models are used to estimate body shape under clothing.
RGBD and multi-view inputs are beneficial to resolve ambiguities, but they are not in the scope of this review.
We simply ignore the modeling of clothes, which is a step towards photorealism. We refer readers to \cite{chen2021towards} for clothed human reconstruction.
We also do not cover work on neural rendering~\cite{tewari2020state,peng2021neural} that focuses on the appearance modeling instead of geometry.
This survey is also complementary to existing survey papers focusing on 2D/3D human pose estimation~\cite{chen2020monocular,zheng2020deep,liu2021recent}.

\subsection{Organization}
The rest of the survey is organized as follows. In Section~\ref{sec:human_models}, we give a brief introduction of the development history of human models and provide detailed information on the SMPL model~\cite{loper2015smpl}, the most widely used template for human reasoning. Section~\ref{sec:single_person} describes approaches to body recovery and whole-body recovery with hands and face. Methods are categorized 
into an optimization-based paradigm or a regression-based paradigm.
In Section~\ref{sec:multi-person} and \ref{sec:video}, we sort out novel modules that help to deal with videos or multi-person recovery.
However, results may be physically unreasonable and suffer from visual defects if we merely supervise the human body with regular data terms. Thus, in Section~\ref{sec:plausibility}, we discuss the strategies used to enhance physical plausibility by involving realistic camera models, contact constraints, and human priors.
The commonly used datasets and evaluation criteria, along with the benchmark leaderboard, are summarized in Sections~\ref{sec:datasets} and~\ref{sec:evaluation}.
Finally, we draw conclusions and point out worthwhile future directions in Section~\ref{sec:conclusion}.

\section{Human Modeling}
\label{sec:human_models}

The human body can be abstracted as a stick figure~\cite{lee1985determination}, simply marking the keypoints in body, hands, and face and connecting them with sticks, as shown in Fig.~\ref{fig:model_skeleton}. However, we interact with the world through surface contacts and facial expressions, which requires the modeling of both body pose and shape.
In early work~\cite{nevatia1977description,ju1996cardboard}, a wide variety of geometric primitives have been studied to approximate body shapes.
Later, inspired by the breakthrough~\cite{blanz1999morphable} in face modeling, researchers derive body shape constraints from 3D scanned data and create body models~\cite{anguelov2005scape, loper2015smpl} from a statistical viewpoint.
Based on modeling details, we classify the modeling process into two classes: methods that represent the human body with \emph{geometric primitives}, and methods that use subject-specific \emph{body scanned data} to build a statistical 3D mesh model.

\subsection{Geometric Primitives}
\begin{figure}[t]
    \begin{subfigure}[b]{0.11\textwidth}
        \centering
		\includegraphics[height=48mm]{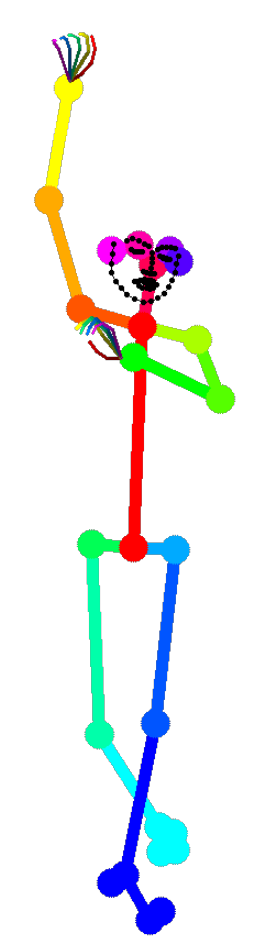}
		\caption{}
		\label{fig:model_skeleton}
    \end{subfigure}
    \begin{subfigure}[b]{0.11\textwidth}
        \centering
		\includegraphics[height=48mm]{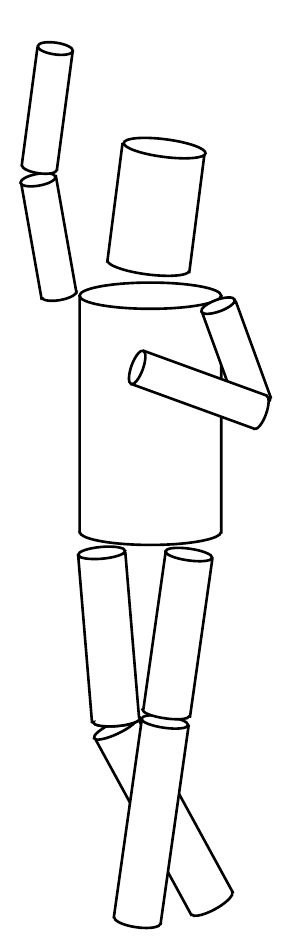}
		\caption{}
		\label{fig:model_cylinder}
    \end{subfigure}
    \begin{subfigure}[b]{0.11\textwidth}
        \centering
		\includegraphics[height=48mm]{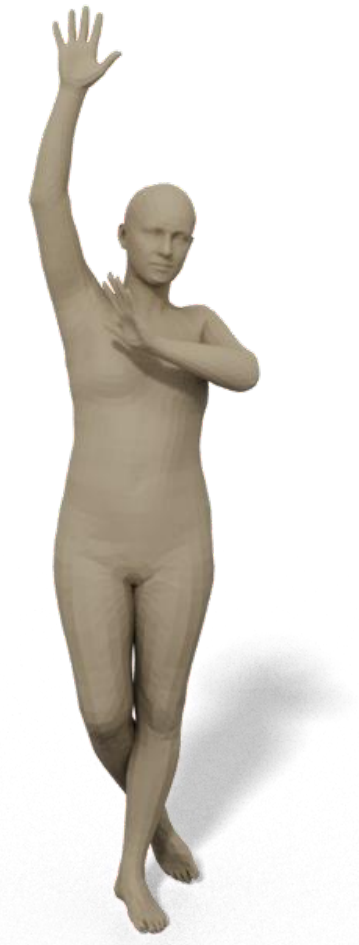}
		\caption{}
		\label{fig:model_smpl}
    \end{subfigure}
    \begin{subfigure}[b]{0.11\textwidth}
        \centering
		\includegraphics[height=48mm]{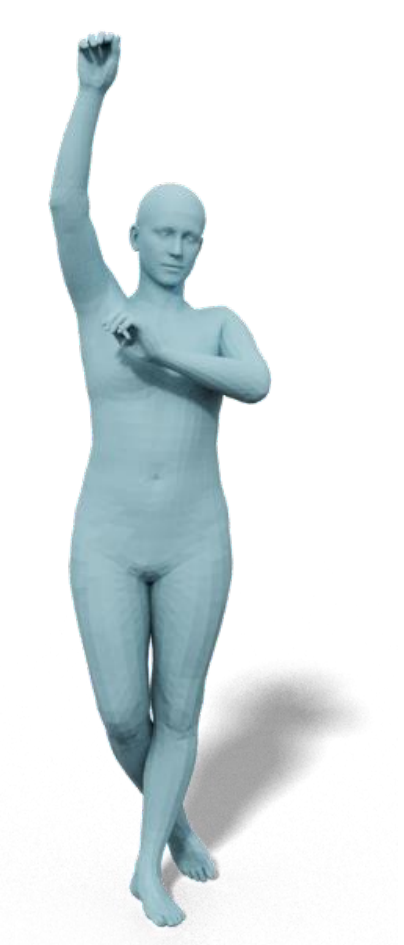}
		\caption{}
		\label{fig:model_smplx}
    \end{subfigure}
	\caption{Typical 2D and 3D human models representing the same posing human. (a) 2D skeletons~\cite{cao2019openpose}, formed from the keypoints of body, hands and face; (b) a cylindrical body model; (c) SMPL~\cite{loper2015smpl}; (d) SMPL-X~\cite{pavlakos2019expressive}.}
	\vspace{-3mm}
	\label{fig:human_models}
\end{figure}

Body modeling starts by manipulating a bunch of geometric primitives, including planar rectangles~\cite{ju1996cardboard}, cylinders~\cite{nevatia1977description, marr1978representation, rohr1994towards, wachter1999tracking, sidenbladh2000stochastic, sigal2010humaneva}, and epllisoids~\cite{wang2020monocular}, as shown in Fig.~\ref{fig:model_cylinder}. Nevatia~\etal~\cite{nevatia1977description} use generalized cylinders to fit range data.
Marr~\etal~\cite{marr1978representation} propose a general, compositional 3D shape representation.
Pentlan~\etal~\cite{pentland1991recovery} attempt to track a jumping man using a model with spring-like connections between body parts.
Later, more sophisticated primitives were proposed, such as superquadric
ellipsoids~\cite{metaxas1993shape, gavrila1996vision, sminchisescu2003estimating}, metaballs~\cite{plankers2001tracking} and customized graphical model~\cite{kakadiaris2000model, wang2020monocular}. 
By then, human body models were hand-crafted, unrealistic, and tended to be brittle.

\subsection{Statistical Modeling}
Compared to primitives-based models, full-body 3D scans offer more detailed measurements of the body surface, but the modeling process is much more complicated. To convert a dense point cloud and a triangulated mesh from 3D scans to a watertight and animatable 3D human body mesh, three main pre-processing steps are taken~\cite{pons-moll_rosenhahn_2011}: (i) \emph{template mesh registration}: fit a template mesh to the 3D point cloud to deal with holes that the triangulated mesh contains; (ii) \emph{skeleton fitting}: determine the number of joints and the location and axis orientations of rotations for each joint; (iii) \emph{skinning}: bind every vertex in the surface to the skeleton for animation.

\subsubsection{Body Modeling}
Statistical body modeling refers to learning a statistical body model by exploiting an extensive collection of 3D body scans and simply ignoring hand articulation or facial expression.
There has been a lot of research~\cite{allen2003space, anguelov2005scape, hasler2009statistical, chen2013tensor, freifeld2012lie, hirshberg2012coregistration, pons2015dyna, allen2006learning, hasler2010learning, wang2020blsm} on learning highly realistic human body models from scanning data like CAESAR~\cite{CAESAR}. Among them, SCAPE~\cite{anguelov2005scape} and SMPL~\cite{loper2015smpl} are two representative models that factor body deformations into identity-dependent and pose-dependent shape deformations.

SCAPE~\cite{anguelov2005scape} is a deformable human body model that represents the individual shape and the pose-dependent shape via triangle deformations.
During processing, Anguelov~\etal combine static scans of several people with the scans of a single person in various poses. SCAPE is one of the most successful human models. Many models~\cite{hasler2009statistical, chen2013tensor, freifeld2012lie, hirshberg2012coregistration, zuffi2015stitched, pons2015dyna} are built upon SCAPE.
The stitched puppet model~\cite{zuffi2015stitched} combines the realism of statistical models with the advantages of part-based representations. Dyna~\cite{pons2015dyna}, an extension of SCAPE,
relates soft-tissue deformations to motion and body shape and enables itself to produce a wide range of realistic soft-tissue motions.

SMPL~\cite{loper2015smpl} is a vertex-based linear model depicting minimally-clothed humans in natural poses, which is currently the most widely used human body model in the research community. It is compatible with existing rendering engines. Like SCAPE~\cite{anguelov2005scape}, SMPL factors deformations into shape and pose deformations. Two basic sets of parameters control pose deformation $\pose$ and shape variation $\shape$, respectively.
The pose parameters $\pose=[\boldsymbol{w}_0^T,...,\boldsymbol{w}_K^T]^T$ are defined by a standard skeletal rig at $K=23$ joints, where $\boldsymbol{w}_k\in \reals^3$ denotes the relative rotation of part $k$ \wrt its parent in the kinematic tree and $\boldsymbol{w}_0$ refers to the root orientation.
The shape parameters $\shape\in \reals^m$ are coefficients of the top-$m$ principal components in a low-dimensional shape space after principal component analysis (PCA).
SMPL can be represented as a function $M(\cdot)$ mapping pose parameters $\pose$ and shape parameters $\shape$ to a triangulated mesh with $N=6890$ vertices. It is formulated as an additive model in the vertex space. Specifically, a posed human body instance can be obtained as follows:
\begin{align}
M(\shape, \pose) &= W(T(\shape, \pose), J(\shape), \pose; \mathcal{W}), \\
T(\shape, \pose) &= \bar{\mathbf{T}}+B_{s}(\shape)+B_{p}(\pose), \label{eq:smpl_f}
\end{align}
where a rest pose $T(\shape, \pose)$ is first generated by learning corrective blend shapes, \ie, pose-dependent deformations $B_{P}(\pose): \reals^{|\pose|} \mapsto \reals^{3 N}$ and shape-dependent deformations $B_{S}(\shape): \reals^{|\shape|} \mapsto \reals^{3 N}$, in order to deal with standard LBS artifacts~\cite{mohr2003building}. A blend shape is a vector of displacements in the mean template shape $\bar{\mathbf{T}}$. Secondly, linear blend skinning function $W$ with a set of blend weights $\mathcal{W}\in \reals^{N\times K}$ and the pose parameters $\pose$ allow to pose the T-shape mesh $T(\shape, \pose)$ based on its skeleton joints locations $J(\shape): \reals^{|\shape|} \mapsto \reals^{3 K}$. Moreover, SMPL can be extended to capture soft-tissue dynamics~\cite{pons2015dyna}. Dynamic deformations of the resulting DMPL model is parameterized by coefficients $\boldsymbol{\delta}$.

The SMPL family has been growing. FLAME face model~\cite{li2017learning}, MANO hands model~\cite{romero2017embodied}, and SMIL infant body model~\cite{hesse2018learning} have been proposed, which are overall a linear blend skinning with shape and pose blend-shapes.
Despite the success in the application, SMPL still has its limitation. First, its global blend shapes capture spurious long-range correlations and result in non-local deformation artifacts. Second, SMPL ignores correlations between body shape and pose-dependent shape deformation. In addition, SMPL relies on a linear PCA subspace to represent soft-tissue deformations, struggling to reproduce highly nonlinear deformations.
Many researchers seek an improvement for descriptive capability~\cite{santesteban2020softsmpl, osman2020star}.
STAR~\cite{osman2020star} is a drop-in replacement for SMPL. It factorizes pose-dependent deformation into a set of sparse and spatially local pose-corrective blend-shape functions. SoftSMPL~\cite{santesteban2020softsmpl} defines a highly efficient nonlinear subspace to encode tissue deformations, compared to the linear descriptors~\cite{pons2015dyna, loper2015smpl}.
\rv{Recently, learning-based solutions are also explored to represent the body model in implicit~\cite{deng2020nasa,mihajlovic2021leap,chen2021snarf,mihajlovic2022coap} or explicit~\cite{sun2023learning} manners.
}

\subsubsection{Whole Body Modeling}
Recently, much progress has been made in modeling the human body together with hands~\cite{romero2017embodied}, or with hands and face~\cite{joo2018total, pavlakos2019expressive}.
Romero~\etal~\cite{romero2017embodied} attach MANO to SMPL to obtain a new articulated model (SMPL+H) with hands and body interaction.
Frankenstein Model~\cite{joo2018total} combines a simplified version of SMPL~\cite{loper2015smpl} with an artist-designed hand rig and the FaceWarehouse face model~\cite{cao2013facewarehouse}. These disparate models are integrated together, resulting in a model that is slightly out of proportion. A simpler parameterized model, Adam, is also introduced, which is more capable of body motion capture.
Pavlakos~\etal~\cite{pavlakos2019expressive} learn a new, holistic model named SMPL-X that jointly models the human body, face, and hands. They extend SMPL with the FLAME head model~\cite{li2017learning} and the MANO hand model~\cite{romero2017embodied} and then register this combined model to CAESAR~\cite{CAESAR} scans to curate for quality. SMPL-X has several parameters representing the body, hand, and face. Initially, there are 75 rotational parameters for the global rotation and \{body, eyes, jaw\} poses; 24 low-dimensional PCA coefficient or 90 rotational parameters for hand poses; 10 for the body shape and 10 for the facial expressions.
\rv{Following SMPL-X, SUPR~\cite{osman2022supr} is proposed recently for more expressive and accurate modeling of head, hand, and foot.}
Besides a series of linear models based on SMPL, some attempts are devoted to different modeling strategies. For example,
GHUM and GHUML shape spaces~\cite{xu2020ghum} are based on variational auto-encoders (VAE), which are nonlinear. All the model parameters, including shape spaces, pose-space deformation correctives, skeleton joint center predictors, and blend skinning functions, are trained end-to-end in a single consistent learning loop.

\begin{figure*}[t]
    \centering
    \includegraphics[width=\linewidth]{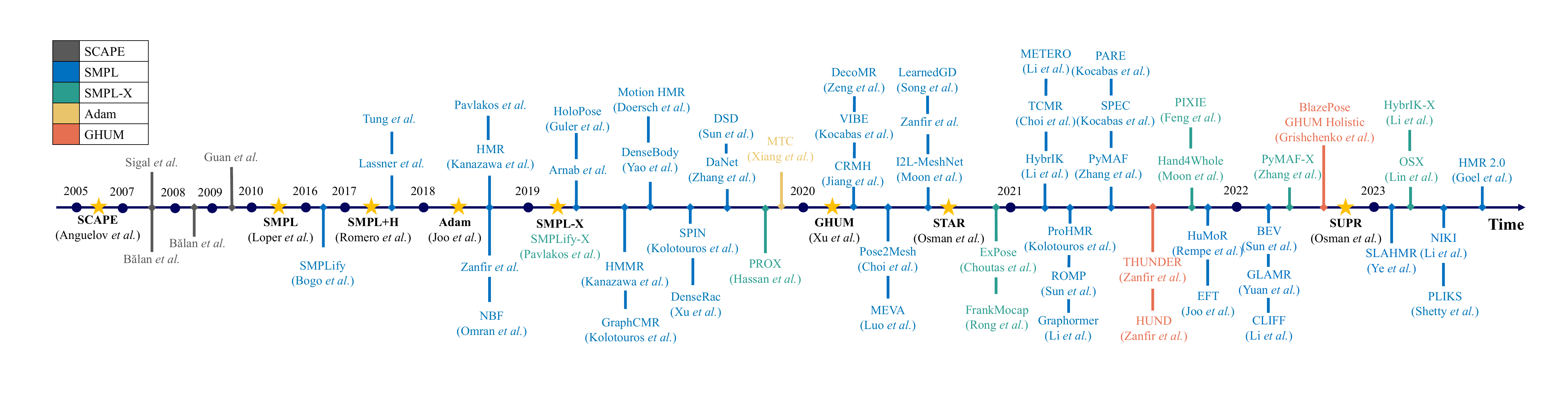}
    \vspace{-9mm}
    \caption{Chronological overview of the most relevant parametric human models and 3D human mesh recovery methods.}
	\vspace{-5mm}
	\label{fig:timeline}
\end{figure*}

\section{Human Mesh Recovery}
\label{sec:single_person}
Since the release of statistical body models, researchers have used them to estimate the shape and pose from monocular images.
Balan~\etal~\cite{balan2007detailed} pioneer in estimating the parameters of SCAPE~\cite{anguelov2005scape} from images. Nowadays, SMPL~\cite{loper2015smpl} has been prevailing in academia for 3D body shape recovery. The credit goes to SMPL's open-source nature and its fast-developing community around it: the ground-truth acquisition methods~\cite{loper2014mosh, bogo2016keep}, datasets with extended SMPL annotations~\cite{ionescu2014human3, von2018recovering, varol2017learning, lassner2017unite, mahmood2019amass}, and milestone works~\cite{kanazawa2018end, kolotouros2019learning, kocabas2020vibe}.
This section will sort out articles on human mesh recovery based on predefined body models~\cite{anguelov2005scape, loper2015smpl, pavlakos2019expressive}. Body models capture the shape and pose variability but do not account for clothing or hair. Thus, to put it more precisely, approaches estimate the shape and pose of the body under clothing or in tight clothing. In Fig.~\ref{fig:timeline}, we demonstrate some representative methods. We categorize them based on the human models they adopt. In Table~\ref{tab:objectives}, we provide a summary of two paradigms, i.e., optimization and regression methods, for the goal of better alignment and physical plausibility.

According to the level of detail in the reconstruction, related approaches are categorized into body recovery (Section~\ref{sec:body_recovery})  and whole-body recovery (Section~\ref{sec:whole_body_recovery}) with expressive hands and face.
In each case, we further divide them into two paradigms.
\emph{Optimization-based} or \emph{fitting-based} approaches explicitly fit a parametric human model to 2D observations in an iterative manner.
On the contrary, \emph{regression-based} methods make use of a deep neural network to regress the representation from image pixels directly.

\subsection{Body Recovery} \label{sec:body_recovery}
Algorithms dealing with body recovery are expected to yield a mesh that reflects the body pose and shape, without considering the detailed recovery of hands and face.

\subsubsection{Optimization-based Paradigm}
Optimization-based approaches attempt to estimate a 3D body mesh consistent with 2D image observations. The objective function typically contains two parts: data terms and regularization terms. Data terms are the measure of alignment between 2D cues and the re-projection of a mesh. To obtain physically plausible body mesh, it is important to introduce regularization terms to favor probable poses over improbable ones.
Before deep learning became all the rage, optimization-based approaches were the leading paradigm for model-based human reconstruction. In the early work, silhouette cues are crucial in fitting a 3D body model, SCAPE~\cite{anguelov2005scape} in most cases, to images. The objective function penalizes pixels in non-overlapping regions~\cite{balan2007detailed, bualan2008naked, guan2009estimating, hasler2010multilinear, zhou2010parametric}. Some literature also requires manually clicked 2D keypoints~\cite{guan2009estimating} or correspondences~\cite{hasler2010multilinear} for a rough fit or camera estimation as initialization.

With the advances in 2D detection in the deep learning era, Bogo~\etal~\cite{bogo2016keep} proposed SMPLify that iteratively fits the SMPL model~\cite{loper2015smpl} to detected 2D keypoints of an unconstrained image.
They adopt an off-the-shelf 2D pose Convolutional Network (ConvNet) to detect the keypoints and perform gradient-based optimization. The objective function is the sum of a joint-based data term and several regularization terms, including an interpenetration error term, two pose priors, and a shape prior.
Specifically, the data term penalizes the distance between detected 2D joints and the projected SMPL joints. The pose priors consist of a penalty on unnatural rotations of elbows and knees, and a mixture of Gaussians trained on CMU marker data~\cite{cmu_mocap}. The shape prior is a quadratic penalty on the shape coefficients estimated via PCA. The interpenetration error term exploits capsule approximations and penalizes the capsule intersections.
The 3D pose generated by SMPLify is relatively well-aligned. However, the shape remains highly unconstrained since the connection length between two keypoints is the only indicator that can be used to estimate the body shape.
To further add constraints, instead of relying solely on one geometric term, \cite{guan2009estimating, lassner2017unite, zanfir2018monocular, xiang2019monocular, guler2019holopose} combine multiple cues for optimization, including 2D keypoints, silhouettes, and segmentations. 
For example,
\cite{zanfir2018monocular} leverage a multi-task neural network that estimates multiple cues to guide a joint multi-person optimization under constraints. In the refinement stage of HoloPose~\cite{guler2019holopose}, the FCN-based estimates of DensePose~\cite{guler2018densepose}, 2D, and 3D keypoints drive the regressed 3D models to better align with image evidence.

Moreover, deep learning techniques can be embedded into the gradient-based optimization process as a powerful tool to enhance robustness and plausibility~\cite{joo2021exemplar, song2020human}.
Given 2D keypoint annotations, Exemplar Fine-Tuning (EFT)~\cite{joo2021exemplar} leverages a fully-trained 3D pose regressor and carries out optimization in the neighborhood of the pre-trained parameters. After the fitting is completed for one sample, the regressor's parameters are re-initialized for a new round. EFT optimizes all body parts without any external regularization terms since the pre-trained regressor implicitly embodies a strong prior. 
Song~\etal~\cite{song2020human} resort to neural networks to generate the parameter update rule. Current parameters, target 2D joints, and the gradient are passed into the network to get the updated term for the next iteration.

Besides, inverse kinematics has also been studied. 
\rv{Forward kinematics (FK) computes the positions of each body joint from specified joint rotations. Conversely, inverse kinematics (IK) calculates body joint rotations that match the given body joints or vertices.}
Iqbal~\etal~\cite{iqbal2021kama} calculate rotations for every joint accordingly based on the number of children. They follow SMPLify to refine the pose and estimate the shape.
Differential IK module in \cite{yu2021skeleton2mesh} relies on a set of kinematics prior knowledge to infer 3D rotations from estimated 3D skeletons.
HybrIK~\cite{li2021hybrik} decompose relative rotations into twist and swing. An adaptive IK algorithm is designed to recover swing angles. The shape and twist angles are learned in a regression-based manner.
\rv{Li~\etal~\cite{li2023niki} later propose NIKI~\cite{li2023niki}. It combines the FK and IK processes using an invertible neural network to explicitly decouple errors from plausible poses. }
\rv{PLIKS~\cite{shetty2023pliks} approximate the rotations based on the UV position map inputs $\rm{X}_{uvd}$ and then solve IK from 2D pixel-aligned vertex inputs $\rm{X}_{uv}$.}

\begin{figure*}[t]
	\begin{center}
		\includegraphics[width=\linewidth]{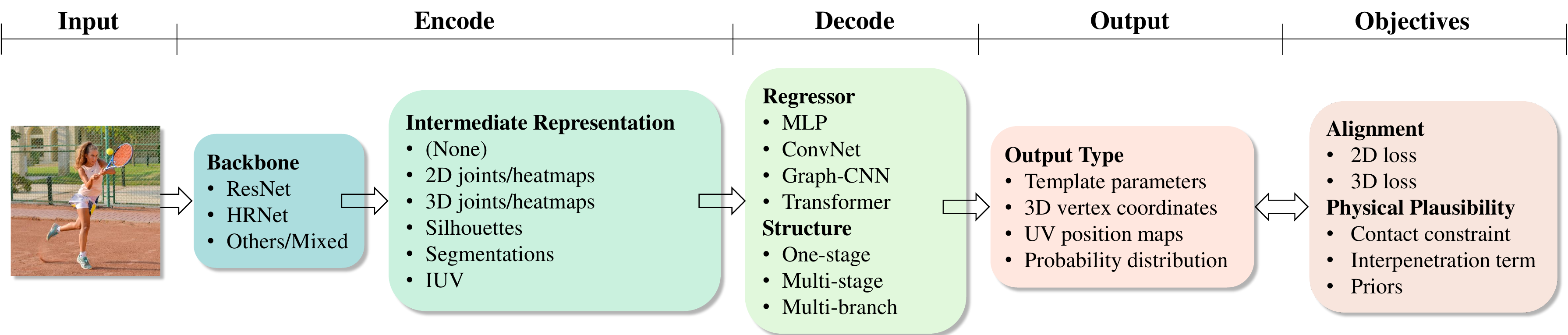}
		\vspace{-7mm}
		\caption{The pipeline of regression-based methods for human mesh recovery.}
		\vspace{-6mm}
		\label{fig:pipeline}
	\end{center}
\end{figure*}

\subsubsection{Regression-based Paradigm}
\begin{figure*}[t]
	\begin{center}
		\includegraphics[width=\linewidth]{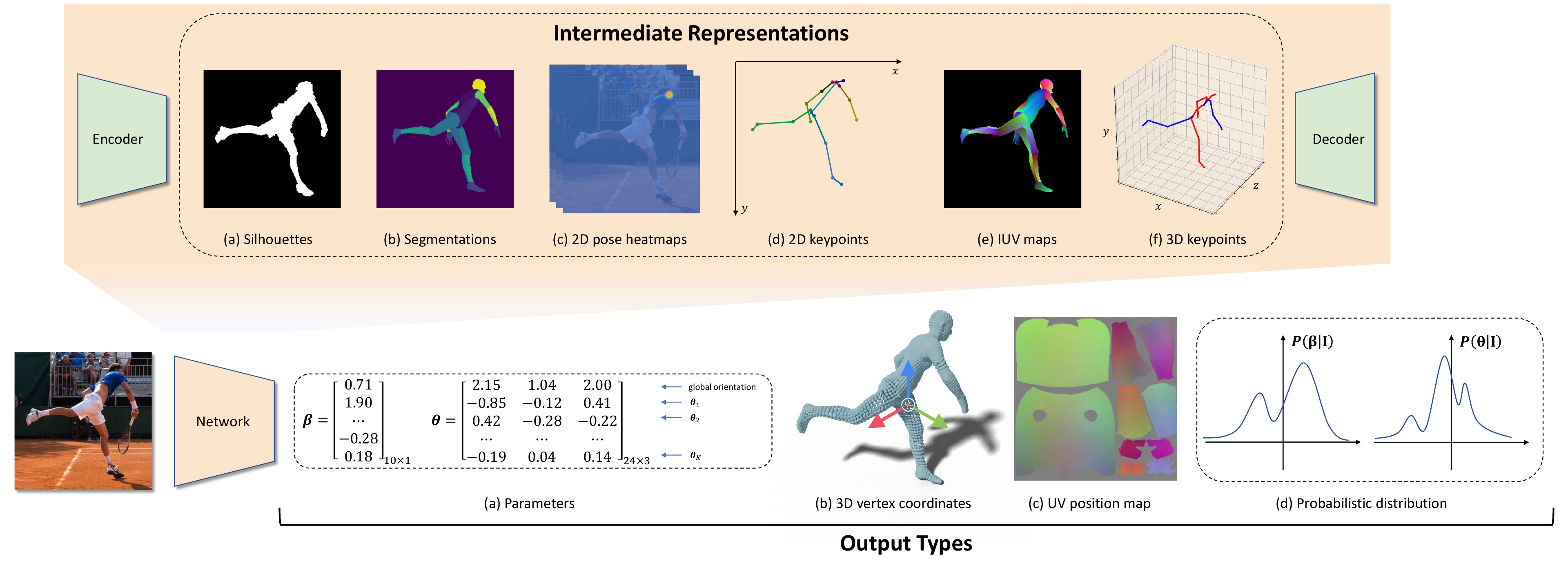}
		\vspace{-9mm}
		\caption{Illustration of various output types and intermediate representations in the regression networks. We investigate four output types: (a) parametric output; (b) 3D coordinates of the mesh vertices; (c) UV position maps; (d) probability distribution over pose and/or shape parameters. Intermediate representations adopted in the multi-stage frameworks include (a) silhouettes; (b) segmentations; (c) 2D pose heatmaps; (d) 2D keypoint coordinates; (e) IUV maps; (d) 3D keypoint coordinates, which can serve as a simplification of the inputs or a guidance.}
		\vspace{-6mm}
		\label{fig:output_intermediate}
	\end{center}
\end{figure*}

Regression-based methods take advantage of thriving deep learning techniques to process pixels directly. Here, we take a step further by breaking the networks apart and going through the similarities and differences. We examine the output types to represent a human mesh, and their motivations and setbacks. Then, we talk about the intermediate representations embedded in the networks as well as various ways to supervise in 2D and 3D space. Finally, we elaborate on the network architectures, which reflect researchers' observations and insights into this task. Fig.~\ref{fig:pipeline} summarizes a typical pipeline of regression-based methods. Moreover, Fig.~\ref{fig:output_intermediate} provides an illustration of various output types and intermediate representations in the regression networks. We also summarize representative regression-based methods in Table~\ref{tab:body_summary}.

\paragraph{Output Type}.
Outputs are mainly divided into two groups: parametric outputs and non-parametric outputs.

\emph{Parametric Output}. The majority of image-based human mesh recovery methods~\cite{kanazawa2018end, omran2018neural, pavlakos2018learning, kolotouros2019learning, zhang2019danet, rockwell2020full, sengupta2020synthetic, xu20203d, zanfir2020weakly, moon2022accurate, li2021hybrik, kocabas2021pare, kocabas2021spec, zhang2021pymaf, li2022cliff} choose to regress the parameters of the parametric models directly.
They are also categorized as ``model-based" approaches.
Since this representation is embedded in a latent space, it is highly abstract. Networks simply need to output a low-dimensional vector, which corresponds to a body with a specific pose and shape.

Pose parameters $\pose$ contain the angle-axis representation of relative rotations of body joints plus the root orientation. Intuitively, networks can directly regress a vector corresponding to joint rotations in axis angle~\cite{kanazawa2018end, pavlakos2018learning, kanazawa2019learning, xu2019denserac, sun2019human, georgakis2020hierarchical}. 
HoloPose~\cite{guler2019holopose} choose to use Euler angles as the regression target alternatively.
However, as demonstrated in \cite{zhou2019continuity}, axis angle and Euler angle representations are discontinuous in the three-dimensional Euclidean space. To overcome the discontinuity, rotation matrices are adopted as the learning objective~\cite{omran2018neural, zhang2019danet, zhang2020learning}.
Learning rotation matrices is beneficial in avoiding discontinuity, but the trade-off is increasing representational redundancy and consequently dimensionality.
Recently, there has been a growing trend to use a 6D representation~\cite{zhou2019continuity}, which is continuous in space, more compact than a matrix, and thus considered more suitable for deep learning~\cite{kolotouros2019learning, zhang2021pymaf, kocabas2020vibe, luo20203d, sengupta2020synthetic, zhou2021monocular, choi2021beyond, kocabas2021pare, moon2022accurate}.
To alleviate the error accumulation issue, SGRE~\cite{Wang2023SGRE} directly estimates the global rotation of each joint instead of relative rotations.

\emph{Non-parametric Output}. The key to the model-based paradigm may be a stumbling block. The template serves as a strong structure prior to handling severe occlusions or ambiguities and generating likely results. In the meantime, it gets stuck in the predefined embedded space, making it harder to align with 2D cues.
Researchers seek to relax this heavy reliance on the parameter space while still retaining the topology. Instead of predicting the template's parameters, some methods directly regress non-parametric body shapes in the form of voxels~\cite{varol2018bodynet, Zheng_2019_ICCV} or 3D positions of mesh vertices~\cite{kolotouros2019convolutional,moon2020i2l,lin2021mesh}.
Among them, BodyNet~\cite{varol2018bodynet} predicts a volumetric representation and then fits a SMPL model.
Kolotouros~\etal~\cite{kolotouros2019convolutional} pioneer in 3D mesh vertex coordinates regression. \cite{lin2021end, lin2021mesh} choose to predict 3D coordinates of mesh vertices and body joints in parallel.
Luan~\etal~\cite{luan2021pc} build up a non-rigid transformation with the guidance of a concise 3D target pose and apply it to every vertex to correct the results from HMR~\cite{kanazawa2018end}.
To model uncertainty and maintain the spatial relationship between pixels in images, I2L-MeshNet~\cite{moon2020i2l} uses lixel (line+pixel)-based 1D heatmap for dense mesh vertex localization. It's a memory-efficient version of voxel-based 3D heatmaps.
Recently released state-of-the-arts~\cite{kolotouros2019convolutional, moon2020i2l, zanfir2021thundr} show that evaluation on non-parametric results generally outperforms model-based ones due to their flexibility.

Apart from generating the locations of each vertex in 3D space, \cite{yao2019densebody, zeng20203d, zhang2020object} utilize UV map and turn the vertex inference problem into an image-to-image translation task, which fits well with the characteristic of convolutional layers. UV maps is a pixel-to-surface dense correspondence map, which are often used for texture rendering.
\rv{By storing the vertex coordinates as the $(R, G, B)$ color values in the UV map, the UV position map is obtained and used as a suitable regression objective for fully convolutional networks.}
In practice, \cite{yao2019densebody, zhang2020object} leverage the default UV map provided by the SMPL model. \cite{zeng20203d} propose a new UV map to maintain neighboring relations on the original mesh surface.

\emph{Probabilistic Output}. 
The above-mentioned are deterministic and uni-modal regression models, typically yielding a single estimate for one input.
Due to reconstruction ambiguity, we can also design a network to produce a set of plausible poses or a probabilistic distribution.
Biggs~\etal~\cite{biggs20203d} learn a multi-hypothesis neural network to generate multiple sets of parameters that are plausible estimates and consistent with the ambiguous views. Sengupta~\etal~\cite{sengupta2021probabilistic} assume simple multivariate Gaussian distributions over SMPL pose parameters $\pose$ and let the network to predict $\mu_{\pose}{(I)}$ and $\delta_{\pose}{(I)}$.
ProHMR~\cite{kolotouros2021probabilistic} models a conditional probability distribution $p{(\pose|I)}$ using Conditional Normalizing Flow, which is more powerful and expressive than Gaussian distributions.
Sengupta~\etal~\cite{sengupta2021hierarchical} estimate a hierarchical matrix-Fisher distribution over the relative 3D rotation matrix of each joint. This probability density function is conditioned on the parent joint along with the body's kinematic tree structure. The shape is still based on a Gaussian distribution.
\rv{
Fang~\etal~\cite{fang2023learning} propose to learn probability distributions for human joint rotations by leveraging the learned analytical posterior probability.
Sengupta~\etal~\cite{sengupta2023humaniflow} improve the consistency and diversity of predictions by modeling the ancestor-conditioned per-body-part pose distributions in an autoregressive manner.
}

\paragraph{Intermediate/Proxy Representation}.
Instead of directly lifting a raw RGB image to a 3D pose, plenty of approaches introduce intermediate representation into network architectures.
Intermediate representations are the outputs of generic human analysis ConvNets.
The benefits of involving 2D/3D cues in the intermediate stage can be summarized by two words: ``simplification" and ``guidance".

Intermediate representations can be viewed as a simplification over RGB inputs, ignoring illumination, clothing, or background clutter, which do not necessarily correlate with human pose and shape. Intermediate estimates take the place of RGB images to be the actual input to the regression network. In this case, they are also referred to as ``proxy representation", such as silhouettes~\cite{pavlakos2018learning, sengupta2020synthetic, sengupta2021probabilistic}, segmentations~\cite{omran2018neural, rueegg2020chained, zanfir2020weakly,zanfir2021neural}, 2D heatmaps~\cite{pavlakos2018learning, doersch2019sim2real, sengupta2020synthetic, zanfir2020weakly, sengupta2021probabilistic, zhou2021monocular}, 2D keypoint coordinates~\cite{choi2020pose2mesh}, optical flow~\cite{doersch2019sim2real, li2022deep}, IUV~\cite{xu2019denserac, zeng20203d, zhang2020learning}, 3D keypoint coordinates~\cite{li2021hybrik, choi2020pose2mesh, zhou2021monocular}, and surface markers~\cite{zanfir2021thundr}.
The introduction of proxy representation distinctly contributes to overcoming data scarcity. The initial stage processes the RGB inputs to proxy representations. However, we can involve synthetic instances in the following stages to make a difference in performance.
Compared to the synthesis of raw RGB images, proxy inputs lead to a smaller synthetic-to-real domain gap which is more readily bridged by data augmentation~\cite{pavlakos2018learning, doersch2019sim2real, xu2019denserac, sengupta2020synthetic,gong2022self}.

On the other hand, intermediate representations guide toward finer information for accurate prediction.
2D keypoint coordinates can be used to obtain part-based information to represent local body structure that is invariant to global image deformation. \cite{guler2019holopose, zhang2020learning} use the pose-guided pooling around keypoints to extract image features and partial IUV, and then adopt a multi-branch framework for individual part-based prediction.
Besides explicit extraction or cropping, features can also be ``purified" implicitly.
Tung~\etal~\cite{tung2017self} concatenate the RGB image and corresponding 2D heatmaps and feed to the network.
Sun~\etal~\cite{sun2019human} use the detected 2D keypoint coordinates and employ bilinear transformation to disentangle the skeleton from the rest details.
Hand4Whole~\cite{moon2022accurate} calculates 3D positional pose from 3D heatmaps and interpolates on the image feature map to obtain joint-level features.
3DCrowdNet~\cite{choi20213dcrowdnet} concatenates image features and 2D heatmaps along the channel dimension, which will be further processed to output a 2D human pose-guided feature with high activation on a target person.
PARE~\cite{kocabas2021pare} predicts part attention masks to model the likelihood of a pixel belonging to a particular joint. The attention masks and image feature maps are fused to aggregate information from attended regions.

\paragraph{Supervision}.
Supervision signals are categorized based on the dimension of space where they play a role.
3D supervision matches the task better as the output is defined in 3D space.
We can supervise the pose parameters $\pose$ in the form of axis angle~\cite{kanazawa2018end, georgakis2020hierarchical, pavlakos2022human, xu20203d}, rotation matrix~\cite{pavlakos2018learning, omran2018neural, zhang2019danet, rong2019delving, sun2019human, xu2019denserac, zhang2020learning, rockwell2020full}, or a 6D representation~\cite{kanazawa2018end, omran2018neural, pavlakos2018learning}.
Once mesh vertices are obtained, we can compute 3D joints using a pre-trained linear regressor and penalize the distance between regressed 3D joints and ground truth~\cite{kanazawa2018end, omran2018neural, pavlakos2018learning}.
Given predicted mesh vertices and the corresponding ground truth vertices, we can also supervise the network with an additional 3D per-vertex loss~\cite{pavlakos2018learning, kolotouros2019convolutional, sengupta2020synthetic, lin2021end}. Though 3D joints and vertices are fully determined by the parameters, the redundancy leads to more stable training and better performance empirically, as each supervision signal has a different granularity~\cite{sengupta2020synthetic}. In the approaches that directly regress vertices~\cite{moon2020i2l, choi2020pose2mesh}, surface normal loss and surface edge loss are included to improve surface smoothness and details.

When 3D annotations are unavailable, we can train networks in a weakly-supervised or unsupervised manner. The strategy of reprojection-and-compare or render-and-compare has been extensively studied to transform 3D outputs to a 2D plane for supervision. The most common form of 2D supervision is 2D joints.
Predicting camera parameters allows us to obtain corresponding 2D joints through re-projection and measure the displacement between ground truth and estimated 2D joints.
Results can also benefit from the supervision for silhouettes~\cite{tung2017self, yu2021skeleton2mesh}, segmentations~\cite{xu2019denserac, zanfir2020weakly, dwivedi2021learning} and dense correspondences~\cite{guler2019holopose, zhang2020learning, zeng20203d, zhang2021pymaf}. As pointed out in \cite{rong2019delving}, dense correspondences such as IUV map~\cite{guler2018densepose} are effective substitutes for 3D annotations.

\paragraph{Network Architecture}.
Generally speaking, network architectures follow an encoder-decoder paradigm.
The encoder is a convolutional backbone that extracts features of input images, while the decoder, or regressor, takes image features as input and outputs regressed results.
Therefore, the core issue is how to design a powerful encoder and an efficient decoder to capture more information from the input and parse it adequately to boost performance.
We review the existing network architectures, summarize design strategies, and organize them into three main categories:
\begin{enumerate}
\item [a.] One-stage frameworks that predict the pose and shape from a RGB image in a single path. No intermediate modalities are generated. 
\item [b.] Multi-stage frameworks that break down the estimation into a series of sub-tasks, then leverage intermediate cues to generate final 3D outputs.
\item [c.] Multi-branch frameworks that predict pose and shape, or each body part independently in different branches after feature disentanglement.
\end{enumerate}

\emph{One-stage Frameworks}.
In a one-stage framework, convolutional backbones like ResNet~\cite{he2016deep} and HRNet~\cite{wang2020deep} are employed as an encoder to generate a global feature~\cite{kanazawa2018end,kolotouros2019convolutional,kolotouros2019learning} or spatial feature~\cite{yao2019densebody,zhang2021pymaf}.
As for the decoders,
the Iterative Error Feedback (IEF) loop in HMR~\cite{kanazawa2018end} reduces the prediction risk compared to regressing in one go. However, it reuses the same global feature during iteration, making the regressor hardly perceive spatial information. 
PyMAF~\cite{zhang2021pymaf} proposes a mesh alignment feedback that leverages mesh-aligned evidence sampled from spatial feature maps to correct parameters in each loop.
\rv{HUND~\cite{zanfir2021neural} utilizes multiple RNN layers, with shared parameters and internal memory, to optimize the result stage by stage.}
GraphCMR~\cite{kolotouros2019convolutional} attaches the global feature to vertices and employs a Graph-CNN to parse neighborhood vertex-vertex interactions and then regress the 3D coordinates of each vertex.
The decoder in \cite{yao2019densebody,zeng20203d} comprises up-sampling and convolutional layers to generate UV position maps.
In~\cite{kolotouros2021probabilistic}, a global feature vector is fed to a conditional normalizing flow to decode the probability distribution over pose parameters.
\rv{
ImpHMR~\cite{cho2023implicit} introduces the neural feature fields and learns the 3D shape and pose with volume-rendered features.
HMR 2.0~\cite{goel20234d} uses ViT~\cite{dosovitskiy2020image} as the image encoder and a standard transformer decoder with multi-head self-attention to make predictions.
}
METRO~\cite{lin2021end} and Graphormer~\cite{lin2021mesh} leverage an encoder-based transformer as a decoder to model non-local intersections among mesh vertices and joints, which complements convolutional operations.
\rv{Following METRO~\cite{lin2021end}, the follow-up works improve the architecture from the aspects of reducing computational cost~\cite{cho2022cross,zheng2023potter,Dou2023TORE}, leveraging pixel-aligned features~\cite{zhang2021pymaf} in their architectures~\cite{kim2023sampling,yoshiyasu2023deformable}, or fusing 2D and 3D features~\cite{Li2023JOTR}.
}

\emph{Multi-stage Architecture}.
Existing methods have also investigated breaking down the process into multiple sub-tasks. The intermediate results gradually get close to the final representation.
An intermediate estimate provides a new starting point, which alleviates the reconstruction difficulty.
A direct strategy is regressing body model parameters on top of intermediate predictions, including 2D/3D joints~\cite{pavlakos2018learning, zanfir2020weakly,choi2020pose2mesh,sengupta2021probabilistic,moon2022accurate}, sihouettes~\cite{pavlakos2018learning, sengupta2021probabilistic}, semantic parts~\cite{omran2018neural,zanfir2021neural,zanfir2020weakly}, and IUV~\cite{xu2019denserac, zhang2020learning}.

\emph{Multi-branch Architecture}.
Pose parameters represent relative rotations of local body parts.
Shape parameters, however, reflect the holistic body figure.
Given the above observation, researchers seek to disentangle global shape features and local part-specific features, resulting in a multi-branch architecture.
Pavlakos~\cite{pavlakos2018learning} design a two-branch network. One branch takes 2D pose heatmaps as input to regress the pose, and the other processes the silhouettes to yield the shape.
HoloPose~\cite{guler2019holopose} pools convolutional features around each keypoint. The pooled local features are sent to a series of linear layers to infer the votes for putative joint angles.
DaNet~\cite{zhang2020learning} decomposes the prediction task into one global stream and multiple local streams. A global IUV map is produced for the camera and shape prediction. A set of partial IUV maps are estimated based on joint-centric RoI pooling for independent predictions of each joint.
HKMR~\cite{georgakis2020hierarchical} expresses the pose as a concatenation of six individual chains and estimate pose parameters on each kinematic chain with a network.
Kocabas~\etal~\cite{kocabas2021pare} use part attention maps to aggregate 3D body features. After obtaining the final feature, they use separate linear layers to predict each SMPL joint rotation.

\rv{
\paragraph{Regression with More Accurate Shape}.
Most of the regression-based methods focus on the accuracy of the poses and overlook the inaccurate shapes.
This issue becomes critical when the inputs contains humans with extreme shapes since the results of typical regression-based methods are close to mean shapes.
To predict more accurate body shapes, Sengupta~\etal~\cite{sengupta2020synthetic} leverage synthetic training data to overcome the lack of shape diversity in prevalent datasets.
SHAPY~\cite{choutas2022accurate} improves body shape estimation by exploiting the data labeled with anthropometric measurements and linguistic shape attributes.
Ma~\etal~\cite{ma20233d} propose to use virtual markers, which are learned from large-scale MoCap data, as intermediate representations for better capture of body shapes.
}

\begin{table*}[t]
    \renewcommand\arraystretch{1.4}
    \centering
    \footnotesize
    \caption{
    Summary of representative regression-based methods for human mesh recovery.}
    \vspace{-3mm}
    \label{tab:body_summary}
    \begin{tabular}{|c|c|c|p{12 cm}<{\raggedright}|}
        \hline
        \multirow{18}*{\raisebox{-1.5cm}[0pt]{\makecell[c]{Frame-\\based}}}
        & \multirow{12}*{\raisebox{-2.5cm}[0pt]{\makecell[c]{Single\\Person}}}
        & \multirow{5}*{\raisebox{-0.8cm}[0pt]{\makecell[c]{Output\\Type}}}
        & \textbf{a) Template paramters}: \cite{kanazawa2018end, omran2018neural, pavlakos2018learning, guler2019holopose, kolotouros2019learning, xu2019denserac, georgakis2020hierarchical, joo2021exemplar, luo20203d, pavlakos2022human, rockwell2020full, rueegg2020chained, sengupta2020synthetic, biggs20203d, xu20203d, zanfir2020weakly, zhang2020learning, fan2021revitalizing, kocabas2021pare, li2021hybrik, moon2022accurate, yu2021skeleton2mesh, zhang2021pymaf, kocabas2021spec}\\
        &
        &
        & \textbf{b) 3D vertex coordinates}: GraphCMR~\cite{kolotouros2019convolutional}, Pose2Mesh~\cite{choi2020pose2mesh}, I2L-MeshNet~\cite{moon2020i2l}, PC-HMR~\cite{luan2021pc}, METRO~\cite{lin2021end}, Graphormer~\cite{lin2021mesh}\\
        &
        &
        & \textbf{c) Voxels}: BodyNet~\cite{varol2018bodynet}, DeepHuman~\cite{Zheng_2019_ICCV}\\
        &
        &
        & \textbf{d) UV position maps}: DenseBody~\cite{yao2019densebody}, DecoMR~\cite{zeng20203d}, Zhang~\etal~\cite{zhang2020object}, PLIKS~\cite{shetty2023pliks}\\
        &
        &
        & \textbf{e) Probabilistic outputs}: Biggs~\etal~\cite{biggs20203d}, Sengupta~\etal~\cite{sengupta2021probabilistic, sengupta2021hierarchical}, ProHMR~\cite{kolotouros2021probabilistic}\\
        &
        &
        & \textbf{f) Whole-body}: SMPLify-X~\cite{pavlakos2019expressive}, ExPose~\cite{choutas2020monocular}, PIXIE~\cite{feng2021collaborative}, Hand4Whole~\cite{moon2022accurate}, PyMAF-X~\cite{zhang2023pymafx}\\
        \cline{3-4}
        &
        & \multirow{6}*{\raisebox{-0.6cm}[0pt]{\makecell{Intermediate/\\Proxy\\Representation}}}
        & \textbf{a) Silhouettes}: Pavlakos~\etal\cite{pavlakos2018learning}, STRAPS~\cite{sengupta2020synthetic}, Skeleton2Mesh~\cite{yu2021skeleton2mesh}\\
        &
        &
        & \textbf{b) Segmentations}: NBF~\cite{omran2018neural}, Rueegg~\etal~\cite{rueegg2020chained}, STRAPS~\cite{sengupta2020synthetic}, Zanfir~\etal~\cite{zanfir2020weakly}, HUND~\cite{zanfir2021neural}\\
        &
        &
        & \textbf{c) 2D pose heatmaps}: Tung~\etal~\cite{tung2017self}, Pavlakos~\etal\cite{pavlakos2018learning}, STRAPS~\cite{sengupta2020synthetic}, Zanfir~\etal~\cite{zanfir2020weakly}, HUND~\cite{zanfir2021neural}, Sengupta~\etal~\cite{sengupta2021hierarchical}\\
        &
        &
        & \textbf{d) 2D keypoint coordinates}: HoloPose~\cite{guler2019holopose}, Pose2Mesh~\cite{choi2020pose2mesh}, Skeleton2Mesh~\cite{yu2021skeleton2mesh}\\
        &
        &
        & \textbf{e) IUV maps}: DenseRac~\cite{xu2019denserac}, DaNet~\cite{zhang2020learning}, DecoMR~\cite{zeng20203d}, \rv{Wang~\etal~\cite{wang2022best}}\\
        &
        &
        & \textbf{f) 3D keypoint coordinates}: I2L-MeshNet~\cite{moon2020i2l}, Pose2Mesh~\cite{choi2020pose2mesh}, HybrIK~\cite{li2021hybrik}, Hand4Whole~\cite{moon2022accurate}, Skeleton2Mesh~\cite{yu2021skeleton2mesh}, \rv{Wang~\etal~\cite{wang2022best}, NIKI~\cite{li2023niki}}\\
        &
        &
        & \rv{\textbf{g) Markers / Dense vertices}}: \rv{THUNDR~\cite{zanfir2021thundr}, Ma~\etal~\cite{ma20233d}, PLIKS~\cite{shetty2023pliks}}\\
        \cline{3-4}
        &
        & \multirow{3}*{\raisebox{-1.0cm}[0pt]{\makecell[c]{Network\\Architecture}}}
        & \textbf{a) One-stage}: HMR~\cite{kanazawa2018end}, GraphCMR~\cite{kolotouros2019convolutional}, DenseBody~\cite{yao2019densebody}, SPIN~\cite{kolotouros2019learning}, PyMAF~\cite{zhang2021pymaf}, METRO~\cite{lin2021mesh}, Graphormer~\cite{lin2021mesh}, ProHMR~\cite{kolotouros2021probabilistic}, \rv{CLIFF~\cite{li2022cliff}, HMR 2.0~\cite{goel20234d}}\\
        &
        &
        & \textbf{b) Multi-stage}: Pavlakos~\etal\cite{pavlakos2018learning}, NBF~\cite{omran2018neural}, DenseRac~\cite{xu2019denserac}, DaNet~\cite{zhang2020learning}, Zanfir~\etal~\cite{zanfir2020weakly}, Pose2Mesh~\cite{choi2020pose2mesh}, STRAPS~\cite{sengupta2020synthetic}, I2L-MeshNet~\cite{moon2020i2l}, DecoMR~\cite{zeng20203d}, Zhang~\etal~\cite{zhang2020object}, PARE~\cite{kocabas2021pare}, THUNDR~\cite{zanfir2021thundr}, HUND~\cite{zanfir2021neural}, Skeleton2Mesh~\cite{yu2021skeleton2mesh}\\
        &
        &
        & \textbf{c) Multi-branch}: Pavlakos~\cite{pavlakos2018learning}, HoloPose~\cite{guler2019holopose}, DaNet~\cite{zhang2020learning}, HKMR~\cite{georgakis2020hierarchical}, PARE~\cite{kocabas2021pare}\\
        \cline{2-4}
        
        & \multirow{2}*{\raisebox{-0.2cm}[0pt]{\makecell{Multiple\\Person}}}
        &
        & \textbf{a) Top-down}: Jiang~\etal~\cite{jiang2020coherent}, 3DCrowdNet~\cite{choi20213dcrowdnet}, Ugrinovic~\etal~\cite{ugrinovic2021body}, \rv{REMIPS~\cite{fieraru2021remips}, Cha~\etal~\cite{cha2022multi}, OCHMR~\cite{khirodkar2022occluded}}\\
        &
        &
        & \textbf{b) Bottom-up}: MubyNet~\cite{zanfir2018deep}, ROMP~\cite{sun2021monocular}, BEV~\cite{sun2021putting}, \rv{PSVT~\cite{qiu2023psvt}}\\
        \hline
        
        \multirow{3}*{\raisebox{-0.5cm}[0pt]{Temporal}}
        & \multirow{1}*{\raisebox{-0.4cm}[0pt]{\makecell[c]{Single\\Person}}}
        & 
        & Tung~\etal~\cite{tung2017self}, HMMR~\cite{kanazawa2019learning}, Doersch~\etal~\cite{doersch2019sim2real}, Pavlakos~\etal~\cite{pavlakos2022human}, Arnab~\etal~\cite{arnab2019exploiting}, DSD-SATN~\cite{sun2019human}, VIBE~\cite{kocabas2020vibe}, MEVA~\cite{luo20203d}, TCMR~\cite{choi2021beyond}, Lee~\etal~\cite{lee2021uncertainty}, MAED~\cite{wan2021encoder}, SimPoE~\cite{yuan2021simpoe}, DTS-VIBE~\cite{li2022deep}, MPS-Net~\cite{wei2022capturing} \\
        \cline{2-4}
        & \multirow{2}*{\raisebox{-0.2cm}[0pt]{\makecell[c]{Multiple\\Person}}}
        & 
        & XNect~\cite{mehta2020xnect}, HMAR~\cite{rajasegaran2021tracking},
        GLAMR~\cite{yuan2022glamr}\\
        &
        &
        & \\
        \hline
    \end{tabular}
    \vspace{-5mm}
\end{table*}

\begin{table*}[t]
    \renewcommand\arraystretch{1.4}
    \centering
    \footnotesize
    \caption{Summary of optimization/regression objectives for better alignment and physical plausibility. For each term, representative methods are listed.}
    \vspace{-2mm}
    \label{tab:objectives}
    \begin{tabular}{|c|c|p{13.0cm}<{\raggedright}|} 
        \hline
        \multirow{9}*{\raisebox{-1.0cm}[0pt]{\makecell[c]{Alignment}}}
        & \multirow{5}*{\raisebox{-0.5cm}[0pt]{\makecell[c]{2D Loss}}}
        & \textbf{a) 2D keypoints}: \cite{bogo2016keep, tung2017self, kanazawa2018end, omran2018neural, pavlakos2018learning, xiang2019monocular, xu2019denserac, pavlakos2019expressive, kanazawa2019learning, kolotouros2019learning, kocabas2020vibe, zanfir2020weakly, choutas2020monocular, zhang2020learning, lin2021end, zhang2021pymaf, kocabas2021pare, kocabas2021spec, joo2021exemplar, sun2021monocular, moon2022accurate}\\
        &
        & \textbf{b) Silhouettes}: Balan~\etal~\cite{balan2007detailed, bualan2008naked}, Sigal~\etal~\cite{sigal2007combined}, Guan~\etal~\cite{guan2009estimating}, Tung~\etal~\cite{tung2017self}, Pavlakos~\etal~\cite{pavlakos2018learning}, Zhang~\etal~\cite{zhang2020object}, Skeleton2Mesh~\cite{yu2021skeleton2mesh}\\
        &
        & \textbf{c) Segmentations}: Lassner~\etal~\cite{lassner2017unite},
        Zanfir~\etal~\cite{zanfir2020weakly}, PARE~\cite{kocabas2021pare}\\
        &
        & \textbf{d) UV Position Map}: DenseBody~\cite{yao2019densebody}, DecoMR~\cite{zeng20203d}, Zhang~\etal~\cite{zhang2020object}\\
        &
        & \textbf{e) IUV}: HoloPose~\cite{guler2019holopose}, Rong ~\etal~\cite{rong2019delving}, DenseRac~\cite{xu2019denserac}, DaNet~\cite{zhang2020learning}, PyMAF~\cite{zhang2021pymaf}\\
        \cline{2-3}
        & \multirow{4}*{\raisebox{-0.4cm}[0pt]{\makecell{3D Loss}}}
        & \textbf{a) Parameters}: \cite{kanazawa2018end, omran2018neural, pavlakos2018learning, pavlakos2019expressive, kanazawa2019learning, kolotouros2019learning, choutas2020monocular, zanfir2020weakly, zhang2020learning, kocabas2020vibe, moon2022accurate, zhang2021pymaf, kocabas2021pare, kocabas2021spec, joo2021exemplar, sun2021monocular, moon2022accurate}\\
        &
        & \textbf{b) 3D keypoints}: \cite{kanazawa2018end, omran2018neural, pavlakos2018learning, xu2019denserac, kocabas2020vibe, choutas2020monocular, moon2020i2l, zhang2020learning, sengupta2020synthetic, li2021hybrik, lin2021end, lin2021mesh, moon2022accurate, zhang2021pymaf, kocabas2021pare, kocabas2021spec, joo2021exemplar, sun2021monocular}\\
        &
        & \textbf{c) Per-vertex}: \cite{pavlakos2018learning, kolotouros2019convolutional, rockwell2020full, zanfir2020weakly, moon2020i2l, zhang2020learning, sengupta2020synthetic, lin2021end, lin2021mesh, moon2022accurate}\\
        &
        & \textbf{d) Surface (edge/normal)}: Pose2Mesh~\cite{choi2020pose2mesh}, I2L-MeshNet~\cite{moon2020i2l}\\
        \hline
        
        \multirow{8}*{\raisebox{-0.2cm}[0pt]{\makecell[c]{Physical\\Plausibility}}}
        & \multirow{1}*{\raisebox{-0.3cm}[0pt]{\makecell[c]{Contact/\\Interpenetration}}}
        & SMPLify~\cite{bogo2016keep}, Zanfir~\etal~\cite{zanfir2018monocular}, PROX~\cite{hassan2019resolving}, SMPLify-X~\cite{pavlakos2019expressive}, MotioNet~\cite{shi2020motionet}, Jiang~\etal~\cite{jiang2020coherent}, Ugrinovic~\etal~\cite{ugrinovic2021body}, LEMO~\cite{zhang2021learning}, M\"{u}ller~\etal~\cite{muller2021self}, Rempe~\etal~\cite{rempe2020contact, rempe2021humor} \\
        \cline{2-3}
        & \multirow{6}*{\raisebox{-0.2cm}[0pt]{\makecell[c]{Pose Prior}}}
        & \textbf{a) Handcrafted prior}: Lee~etal~\cite{lee1985determination}, Akhter~\etal~\cite{akhter2015pose}, SMPlify~\cite{bogo2016keep}, SMPLify-X~\cite{pavlakos2019expressive}\\
        &
        & \textbf{b) GMM}: SMPLify~\cite{bogo2016keep}, MTC~\cite{xiang2019monocular}\\
        &
        & \textbf{c) MoE}: Sigal~\etal~\cite{sigal2007combined}, HoloPose~\cite{guler2019holopose}\\
        &
        & \textbf{d) GAN}: HMR~\cite{kanazawa2018end}, DenseRac~\cite{xu2019denserac}, Jiang~\etal~\cite{jiang2020coherent}, BMP~\cite{zhang2021body}\\
        &
        & \textbf{e) VAE}: SMPLify-X~\cite{pavlakos2019expressive}, HKMR~\cite{georgakis2020hierarchical}, GHUM~\cite{xu2020ghum}\\
        &
        & \textbf{f) Normalizing Flows}: Biggs~\etal~\cite{biggs20203d}, Zanfir~\etal~\cite{zanfir2020weakly}, GHUM~\cite{xu2020ghum}, Fan~\etal~\cite{fan2021revitalizing}, ProHMR~\cite{kolotouros2021probabilistic}\\
        \cline{2-3}
        & \multirow{1}*{\raisebox{-0.0cm}[0pt]{\makecell[c]{Motion Prior}}}
        & HMMR~\cite{kanazawa2019learning}, VIBE~\cite{kocabas2020vibe}, MEVA~\cite{luo20203d}, HuMoR~\cite{rempe2021humor}, LEMO~\cite{zhang2021learning}, SimPoE~\cite{yuan2021simpoe}, GLAMR~\cite{yuan2022glamr} \\
        \hline
    \end{tabular}
    \vspace{-2mm}
\end{table*}

\subsection{Whole Body Recovery with Hands and Face} \label{sec:whole_body_recovery}
To comprehensively understand human behavior, we need to further capture facial expressions and hand gestures along with body poses.
A straightforward way to get there is by performing individual reconstruction of the body, hands, and face from images and stitching them together.
However, such a strategy leads to unrealistic and unnatural results.
To overcome this, the community has introduced expressive human models~\cite{joo2018total,pavlakos2019expressive} for a unified reconstruction.

\subsubsection{Individual Reconstruction of Hands and Face}

We start with the individual methods of hand and face reconstruction. These methods can be directly combined with the body reconstruction methods to achieve a naive whole-body recovery.

\paragraph{Hands Reconstruction}.
There are also considerable efforts devoted to 3D hand pose prediction from monocular images~\cite{Baek_2019_CVPR,Boukhayma2019,honnotate2020,hasson_2019_cvpr,Iqbal_2018_ECCV,kulon2019rec,Mueller_2018_GANerated,Tekin_2019_CVPR}.
Based on the outputs, these methods can be grouped into two categories, \ie, methods for 3D joints prediction~\cite{Iqbal_2018_ECCV,Mueller_2018_GANerated,Tekin_2019_CVPR,brox_ICCV_2017}, methods producing statistical mesh models~\cite{Baek_2019_CVPR,Boukhayma2019,hasson_2019_cvpr,kulon2019rec,Zhang_2019_ICCV,Ge2019,Kulon_2020_CVPR,park2022handoccnet}.
\rv{Since the release of the two-hand dataset InterHand2.6M~\cite{moon2020interhand2}, there have been considerable efforts devoted to reconstructing interacting hands from monocular images. Similar to the body or hand mesh recovery methods, existing approaches to two-hand reconstruction have also explored different intermediate representations~\cite{wang2020rgb2hands,zhang2021interacting,li2022interacting}, refinement strategies~\cite{li2022interacting,wang2023memahand}, graph convolution networks~\cite{li2022interacting}, the implicit representation~\cite{lee2023im2hands}, the attention mechanism~\cite{li2022interacting,yu2023acr}, and strategies to handle in-the-wild inputs~\cite{moon2023bringing}.
We believe these advances in integrating hand reconstruction could also provide helpful insights and solutions for integrating human mesh recovery and whole-body mesh recovery.
}
For a thorough review of the recent advances in 3D hand pose and shape estimation, please refer to~\cite{chatzis2020comprehensive,huang2021survey}.

\paragraph{Face Reconstruction}.
To tackle the monocular 3D face reconstruction problem, existing solutions also follow the optimization-based~\cite{AldrianSmith2013,VetterBlanz1998,Thies2016} and regression-based strategies~\cite{Feng2018,Jackson2017,Sanyal2019_ringnet,Tewari2018}.
Recent state-of-the-art methods~\cite{Tewari2017,Tewari2018,Deng2019,LuanTran2019,DECA_2020,wang2022faceverse,zielonka2022towards} typically render face images with estimated lighting, albedo, and geometry of the face model using a differentiable renderer~\cite{Loper:ECCV:2014,ravi2020pytorch3d} and compare the synthetic images with the inputs.
Such an analysis-by-synthesis strategy facilitates the demand for in-the-wild images and helps to recover geometric details.
Moreover, recent progress~\cite {DECA_2020,Deng2019,Genova2018} also exploits face recognition~\cite{Cao2018_VGGFace2} to obtain more accurate facial reconstruction results.
For a complete overview of recent face reconstruction methods, please refer to~\cite{Egger_3DMM_survey}.

\subsubsection{Unified Reconstruction}
After unified 3D human models~\cite{joo2018total, pavlakos2019expressive, xu2020ghum} are developed to account for the limitations in expressiveness, whole human body recovery methods have been proposed accordingly to estimate body posture, facial expression together with hand gestures as a whole.

\paragraph{Optimization-based Paradigm}.
Similar to human body recovery, optimization-based methods~\cite{xiang2019monocular,pavlakos2019expressive,xu2020ghum,yi2023generating,zioulis2023kbody} for whole-body recovery detect reliable 2D cues using pre-trained detectors and fit the parametric model to these observations.
Xiang~\etal~\cite{xiang2019monocular} train a ConvNet to predict joint confidence maps and Part Orientation Fields (POF) for the body, hands, face, and feet. They iteratively optimize the objective function to fit the Adam model~\cite{joo2018total} to data terms.
To fit SMPL-X to a single RGB image, Pavlakos~\etal~\cite{pavlakos2019expressive} present SMPLify-X that follows SMPLify~\cite{bogo2016keep} by first detecting 2D features~\cite{cao2019openpose, simon2017hand} corresponding to the face, hands, and feet and optimizing the model parameters afterward. SMPLify-X makes several improvements, including a better-performing pose prior based on a variational auto-encoder (VAE), self-collision penalty terms, and an updated interpenetration term.
Xu~\etal~\cite{xu2020ghum} set anatomical joint angle limits and optimize GNUM parameters using a joint reprojection term and a semantic body-part alignment term.
Like body-only recovery, optimization-based methods tend to be slow and sensitive to initialization.

\paragraph{Regression-based Paradigm}.
Leveraging expressive human models and paired data, the community has also resorted to an end-to-end training fashion for whole-body reconstruction. Among existing solutions, the \emph{divide-and-conquer} strategy~\cite{choutas2020monocular, rong2020frankmocap, moon2022accurate, feng2021collaborative, zanfir2021neural, zhang2023pymafx, li2023hybrikx} is commonly used to break the reconstruction problem down into its parts where the estimation of the bodies, hands, faces is conducted separately with part-specific models. The final expressive 3D human mesh is obtained by forwarding the outputs of each branch to the body template layer.
For example,
ExPose~\cite{choutas2020monocular} directly predicts hands, face, and body parameters in the SMPL-X format and utilizes the body estimation to localize the face and hands regions and crop them from the high-resolution inputs for refinement. It learns part-specific knowledge from existing face- and hand-only datasets to improve performance.
Zhou~\etal~\cite{zhou2021monocular} is a real-time method that captures body, hands, and face with competitive accuracy by exploiting the inter-part relationship between body and hands. SMPL+H~\cite{romero2017embodied} and 3DMM~\cite{blanz1999morphable} are used to represent the body+hands and face.
Hand4Whole~\cite{moon2022accurate} obtains the joint-level features from feature maps, and regresses the 3D body/hand joint rotations from them.
PIXIE~\cite{feng2021collaborative} estimates the confidence of part-specific features and fuses the face-body and hand-body features weighted according to moderators. The fused features are fed to the independent regressors for robust regression.
In addition, fine facial details, \ie, geometry, albedo, and illumination, are predicted in \cite{zhou2021monocular, feng2021collaborative}.
\rv{
Sun~\etal~\cite{sun2022learning} predict hands, and face parameters based on detected whole-body 2D keypoints, making it feasible to take advantage of synthetic data during training.
To resolve conflicts and merge the results from all sub-networks,
PyMAF-X~\cite{zhang2023pymafx} proposes an adaptive integration with elbow-twist compensation.
HybrIK-X~\cite{li2023hybrikx} recalculates the rotations of the parents of the conflict joints.
OSX~\cite{lin2023one} proposes a transformer-based one-stage method to capture the connections of body parts.
SMPLer-X~\cite{cai2023smplerx} investigates scaling up the sizes of models and data for whole-body recovery.
RoboSMPLX~\cite{pang2023robosmplx} improves the localization and feature extraction of body parts for a more robust recovery of whole-body models.
SGNify~\cite{forte2023reconstructing} improves the 3D hand poses by leveraging linguistic priors as constraints for more natural whole-body mesh recovery from sign-language videos.
Despite the recent progress, recovering the whole-body model with plausible hand gestures remains challenging, especially in the cases of interacting hands, occlusions, and motion blur.
}

\section{Multi-person Recovery}
\label{sec:multi-person}
In order to recover 3D human mesh from crowded scenes, we categorize the mainstream methods into two classes based on the design strategy: 1) the top-down strategy and 2) the bottom-up strategy. 

The top-down design reduces the multi-person recovery task to the single-person setting. Cropped single-person images from off-the-shelf detectors~\cite{redmon2018yolov3, he2017mask} are the actual input to the network. In this way, we get to adopt all kinds of single-person regression methods mentioned above. However, truncations, person-person occlusions, and human-scene intersections are ubiquitous in multi-person scenes, which impede the network from perceiving holistic information about a target person. As pointed out in \cite{sun2021monocular}, when two people overlap each other badly, it lacks sufficient context for a network to distinguish the regression target from similar patches. 
Thus, 3DCrowdNet~\cite{choi20213dcrowdnet} takes advantage of robust 2D pose outputs to produce a pose-guided feature that disentangles the target person's feature from others. 3D joint coordinates and SMPL parameters are later derived from the 2D pose-guided feature.
Zanfir~\etal~\cite{zanfir2018monocular} fit a parametric human model for every person based on 2D and 3D observations provided by a multi-task deep neural network. They jointly perform multi-person optimization over all people in the scene, including collision and ground-plane constraints.
Zanfir~\etal~\cite{zanfir2018deep} identify and score different body joint connections, and assemble limbs into skeletons. The feature volume and its identified skeleton are mapped into a shape and pose parameter pair for each person. Note that it is still a multi-stage pipeline, and the last module operates in a top-down manner. 
Jiang~\etal~\cite{jiang2020coherent} explore a R-CNN-based architecture for detection and estimation for all people in the image. To encourage reconstruction in the depth order and avoid overlapping, they incorporate a depth ordering-aware loss and an interpenetration loss during training.
\rv{REMIPS~\cite{fieraru2021remips} creates a sequence of spatial feature tokens and person tokens based on the detected bounding boxes. The tokens are fed to a transformer encoder to make predictions.
In \cite{cha2022multi}, a relation-aware transformer takes every person's image feature and 3D mesh as input to refine the multi-person predictions.}

The top-down paradigm has been criticized for repeated feature extraction and limited receptive field within the bounding box. These drawbacks make it harder to speed up and perform robustly in occlusion and truncation cases.
Instead of designing a multi-stage pipeline, the bottom-up paradigm preserves a holistic view and provides simple one-stage solutions that are computationally efficient.
Single-shot methods~\cite{sun2021monocular, zhang2021body} exploit point-based representation to represent instances by a single point at their center. Using multiple heads, they simultaneously predict an instance localization heatmap and a mesh parameter map.
ROMP~\cite{sun2021monocular} constructs a repulsion field to push apart body centers that are too close. 
BMP~\cite{zhang2021body} improves the inter-instance ordinal depth loss and adopts a keypoint-aware augmentation strategy during training.
\rv{Crowd3D~\cite{wen2023crowd3d} proposes a framework to reconstruct the body model and global locations of hundreds of people from a single large-scene image. 
PSVT~\cite{qiu2023psvt} is an end-to-end multiperson 3D human pose and shape estimation framework with the proposed progressive video Transformer.
}

To deal with the human-human interactions, Zanfir~\etal~\cite{zanfir2018monocular} introduce a collision constraint to prevent the human models from overlapping. Parallelepipeds are fitted to each person at first. If the far-range check fails, the authors adaptively represent the limbs by a series of spheres and calculate the distances based on centers and radius as the volume occupancy loss.
\rv{REMIPS~\cite{fieraru2021remips} employs an interaction-contact loss based on the contact signature and the distance at the facet level.}
Jiang~\etal~\cite{jiang2020coherent} deploy an adapted Signed Distance Field (SDF) to the multi-person scene that takes positive inside each human and zero outside. Based on this, they compute an interpenetration loss for every vertex in every human model.
\rv{OCHMR~\cite{khirodkar2022occluded} uses a global centermap and a subject-specific local centermap to encode the spatial context for each person, which serves as a conditioned input to normalize intermediate features.
}
\rv{Besides interpenetration, depth order incorrection often occurs in rendering multiple persons. The authors also propose a depth ordering-aware loss based on the segmentation and depth maps.}
\section{Recovery from Monocular Videos}
\label{sec:video}
Human mesh recovery from monocular videos is a step forward in understanding human behavior. Image-based methods process each frame independently. The reconstruction results are prone to suffer from occlusions and motion jitters due to the lack of temporal constraints.
For this reason, a good design for videos should exploit the full potential in feature encoding to enhance consistency in spatial and temporal dimensions.

Temporal encoding functions are typically represented as convolutional and recurrent networks.
For example, Doersch~\etal~\cite{doersch2019sim2real} extract features from a combination of optical flow and 2D heatmaps via a single-frame ConvNet followed by an LSTM.
In typical methods~\cite{kanazawa2019learning, sun2019human, kocabas2020vibe, luo20203d, choi2021beyond, lee2021uncertainty, wei2022capturing,zhang2023two}, a pre-trained backbone like ResNet-50~\cite{he2016deep} is used to process raw images to generate static features.
After that, Kanazawa~\etal~\cite{kanazawa2019learning} adopt a 1D fully convolutional model as a temporal encoder. 
Follow-up works~\cite{choi2021beyond, lee2021uncertainty, kocabas2020vibe, luo20203d} employ bidirectional GRUs to incorporate the information from all frames and get temporally correlated features. Besides, TCMR~\cite{choi2021beyond} applies two more GRUs to forecast additional temporal features for the current target pose from the past and future frames. 
Lee~\etal~\cite{lee2021uncertainty} consider the uncertainty-aware embedding and include optical flow information.
\rv{Wei~\etal~\cite{wei2022capturing} extend the non-local operation~\cite{wang2018non} to recalibrate the range of attention in a motion sequence.}
Lately, there has been a trend to adopt the multi-head self-attention (MHA) module~\cite{vaswani2017attention} for long-term sequence dependency modeling~\cite{sun2019human, pavlakos2022human}.
Wan~\etal~\cite{wan2021encoder} modify the original MHA to perform spatial and temporal encoding simultaneously.
\rv{GLoT~\cite{shen2023global} proposes a transformer to decouple the long-term and short-term modeling of temporal motions.
}
HMR-ViT~\cite{Cho2023Video} takes into account both temporal and kinematic information by leveraging temporal-kinematic features in a vision transformer.
\rv{
To handle out-of-domain video inputs, Guan~\etal~\cite{guan2021bilevel} propose a bilevel online adaptation with temporal constraints, while Nam~\etal~\cite{Nam2023Cyclic} propose a cyclic test-time adaptation strategy.
}

On the other hand, different decoding strategies and optimization objectives have been proposed to reduce jitters and improve smoothness.
The decoder in \cite{kanazawa2019learning, kocabas2020vibe, luo20203d, choi2021beyond} iteratively refines the parameters for each frame based on HMR~\cite{kanazawa2018end}.
HMMR~\cite{kanazawa2019learning} includes a dynamics predictor to predict the change of pose parameters in a time step.
MEVA~\cite{luo20203d} learns a human motion subspace via variational autoencoder (VAE) to generate coarse but smooth motions. Finer motions are later retrieved as residuals.
TCMR~\cite{choi2021beyond} passes integrated features, features from past frames, and features from future frames to a shared regressor. All three outputs are supervised with the ground truth of the current frame.

Apart from the architecture, different supervision strategies have also been explored in existing solutions.
Tung~\etal~\cite{tung2017self} compute a motion re-projection error between the predicted 2D body flow and estimated 2D optical flow field in two consecutive frames.
Zanfir~\etal~\cite{zanfir2018monocular} design a velocity prior, assuming that the displacement of pose angles and translation in two adjacent frames should be constants.
Sun~\etal~\cite{sun2019human} first shuffle the frames and adopt an adversarial training strategy to recover the correct temporal order.
Arnab~\etal~\cite{arnab2019exploiting} adopt a temporal error on 3D joints, camera parameters, and 2D keypoint re-projection to penalize the changes between two consecutive frames.
MTC~\cite{xiang2019monocular} defines a texture consistency term based on the flow mapping and enforces a smoothness constraint for the depth of 3D joint locations.
Tripathi~\etal~\cite{tripathi2020posenet3d} use a sliding window to penalize 3D joints of the same frames before and after the window strides.
Wan~\etal~\cite{wan2021encoder} use a series of learnable linear regressors to decode joint rotations in a hierarchical order.
Some objective terms are predefined empirically or learned from large motion capture datasets~\cite{cmu_mocap, mahmood2019amass}. We treat them as ``motion priors", which are of great importance and will be discussed thoroughly in \textsection~\ref{sec:motion_prior}.

\rv{There has been a movement to predict in the world coordinate system by combining camera motion, multi-person tracking, and reconstruction into one system~\cite{yuan2022glamr,ye2023decoupling,goel20234d,sun2023trace}.}
GLAMR~\cite{yuan2022glamr} recovers human meshes in a consistent global coordinate system after extracting motions in the local coordinate system, infilling missing poses, predicting global trajectories, and jointly optimizing camera poses and global motions. It deals with monocular videos that are recorded with dynamic cameras.
\rv{
D\&D~\cite{li2022d} proposes an inertial force control module to improve the dynamic motions estimated from videos with moving cameras.
SLAHMR~\cite{ye2023decoupling} first initialize relative camera motion with SfM, and people tracks with PHALP. These are fed to a joint optimization system to solve the camera scale, the ground plane, and each person's trajectory in the world coordinate system.
TRACE~\cite{sun2023trace} predicts a motion offset map, a world motion map to reason about human trajectories in camera coordinates and world coordinates, respectively. A memory unit is used to predict the tracked identities.
4DHumans~\cite{goel20234d} takes HMR 2.0 as the backbone and adapts the PHALP tracker~\cite{rajasegaran2021tracking} with a vanilla transformer to track people as well as predict future pose parameters.
}
\section{\rv{Human-Scene Interactions}}
\rv{
Human-Scene interactions are ubiquitous. However, given monocular inputs, most works perceive 3D humans in isolation from the surroundings.
Considering 3D humans, scenes, and interacting objects as a whole and inferring the spatial arrangement and contacts help us understand interactive scenarios better. This section discusses several works that reason 3D humans together with scenes from monocular RGB images.
Pioneering attempts~\cite{weng2021holistic, zhang2020perceiving, xie2022chore, yi2022human} infer 3D humans and objects separately, which is prone to be visually unrealistic. To encourage plausibility, various objective functions are proposed over interactions, collisions, and contacts to optimize modules in the scene jointly, which we will introduce in Section~\ref{sec:plausibility} with more details.
}

\rv{
HolisticMesh~\cite{weng2021holistic} imposes human-scene losses at the joint optimization stage, including human-scene collision, human-object contact, and ground support. PHOSA~\cite{zhang2020perceiving} optimizes spatial arrangement using instance-level and part-level interaction losses, a scale loss, and an ordinal depth loss. 
But both of these two methods~\cite{weng2021holistic, zhang2020perceiving}  depend on pre-defined candidate contact vertices or pairs to constrain interaction, which limits the generalization to diverse scenes.
CHORE~\cite{xie2022chore} first learns to predict several 3D neural fields that are more robust than plain 2D evidence. The predicted neural fields serve as stronger 3D terms to provide constraints in the optimization process of SMPL and the object template.
MOVER~\cite{yi2022human} optimizes a plausible scene by refining the camera orientation, object layout, interactions, and ground plane based on expected contacts and 2D segmentations. 
There are also scene-aware approaches~\cite{luo2022embodied,shen2023learning} to recovering plausible human motions in a pre-scaned 3D scene.
}

\section{Physical Plausibility}
\label{sec:plausibility}
Existing methods can produce 3D shape and pose well-aligned to 2D joints but still suffer from visual artifacts, such as ground penetration, foot skating, and body leaning. Only supervising the human body is insufficient to get a consistent result.
To obtain a physically plausible reconstruction, realistic camera models, contact constraints, and shape/pose priors should be taken into account.

\subsection{Camera Model}
Due to the lack of camera information in uncalibrated images, it is virtually impossible to find the exact intrinsic and extrinsic parameters of a perspective camera from monocular images.
For simplicity, a weak-perspective camera model with a large constant focal length is widely used to approximate perspective cameras, in which only scale $s\in \mathbb{R}$ and camera translation $t\in \mathbb{R}^{2}$ along the $x, y$-axis are unknown and need to be predicted~\cite{kanazawa2018end, zanfir2020weakly, kolotouros2019learning, sengupta2020synthetic, zhang2021pymaf}. The scale parameter is further converted into camera translation along the $z$-axis.
However, this simplification does not completely tally with the real-world data.
Focal length impacts the field of view (fov), depth of field, and the sense of perspective.
Also, the effect of camera rotation, such as a significant pitch, can not be entirely counteracted by the rotations along the body kinematic chain. Besides, by merely processing cropped images, we fail to know the actual location in the original image, causing difficulty in a real-world application.

There are methods~\cite{kissos2020beyond, kocabas2021spec, yuan2022glamr} that recover human meshes based on original images and estimate in the world coordinate system.
\rv{Kissos~\etal~\cite{kissos2020beyond} and CLIFF~\cite{li2022cliff} approximate a realistic focal based on the width and height of the original image and convert the camera translation parameter to calculate the reprojection loss in the full image instead of the cropped one.}
SPEC~\cite{kocabas2021spec} computes the camera intrinsics and rotation by predicting the pitch, roll, and vertical field of view (vfov).
GLAMR~\cite{yuan2022glamr} adopts a dynamic camera in global coordinates and jointly optimizes the camera poses and global motions to match the video evidence.
For more robust pose estimation in the real world, Cho~\etal~\cite{cho2021camera} and Zolly~\cite{Wang2023Zolly} also take the perspective distortion issue into account.

\subsection{Contact Constraint}
\label{sec:contact}
The primary purpose of contact constraints is to encourage proper contacts and penalize erroneous interpenetration.

We start with the human-scene contact.
\cite{zanfir2018monocular, ugrinovic2021body} fit a ground plane to the selected 3D ankle positions of all people in a frame, and use the estimated normal vector and a reference point fixed in the plane to penalize the ankle joints away from the plane.
In~\cite{shi2020motionet, rempe2020contact, rempe2021humor,shen2023learning}, the human-scene contact status is predicted to improve the plausibility.
Specifically, Rempe~\etal~\cite{rempe2020contact} design a physics-based trajectory optimization that takes the predicted foot contacts from 2D poses as input and outputs 6D center-of-mass motion, feet positions, and contact forces.
\rv{
The physics-based models are also used to enable full-body contacts~\cite{gartner2022differentiable} or trajectory optimization~\cite{gartner2022trajectory}.
}
Shi~\etal~\cite{shi2020motionet} supervise the network to infer a binary label indicating whether the foot is in contact with the ground and encourage the velocity of foot positions to be zero in contact.
HuMoR~\cite{rempe2021humor} generates a contact probability for each joint. The contact probability output gives weight to an environment regularizer to ensure consistency in joint positions and joint heights among frames.
\rv{Xie~\etal~\cite{xie2021physics} exert contact forces on the feet at 4 different points and design a contact loss to penalize violation of Signorini conditions.}
Going beyond the flat ground,
\cite{hassan2019resolving, zhang2021learning} delve deep into the vertex representation and perform scene reconstruction as the first step. PROX~\cite{hassan2019resolving} penalizes the contact candidate vertices of the body far away from the nearest 3D scene mesh vertices. The contact term only considers body-scene proximity and thus fails to prevent the foot-ground skating problem. To address this issue, LEMO~\cite{zhang2021learning} decomposes the velocity of contacted vertices and regularizes the component along the scene normal to be non-negative, and the component tangential to the scene to be small to prevent sliding.
\rv{
Huang~\etal~\cite{huang2022neural} propose to train a motion distribution prior with a physics simulator and introduce an interaction constraint based on signed distance fields to
enforce ground contact modeling.
}
IPMAN~\cite{tripathi20233d} defines a stability loss based on the estimated Center of Pressure (CoP) and Center of Mass (CoM), and a ground contact loss based on the vertices' height.

Apart from the human-human interactions elaborated in Section~\ref{sec:multi-person}, some self-contacts exist between body parts.
To vividly model the hands touching the body and contact between other body parts,
M\"{u}ller~\etal~\cite{muller2021self} compute an approximated surface-to-surface distance to detect self-contact. They adapt SMPLify-X~\cite{pavlakos2019expressive} by adding self-contact-related objectives, and one of them encourages every vertex in the self-contact pairs to be in contact.
Similarly, Fieraru~\etal~\cite{fieraru2021learning} detect self-contact and design losses to enforce the constraint explicitly.
On the other hand, to avoid self-collision and penetration of several body parts,
Bogo~\etal~\cite{bogo2016keep} approximate the body parts using an ensemble of capsules and penalize the intersections between the incompatible capsules. Although the approximation is computationally efficient, it lacks details.
\cite{pavlakos2019expressive, hassan2019resolving} leverage Bounding Volume Hierarchies (BVH)~\cite{teschner2005collision} to detect a list of colliding body triangles for a more accurate collision penalizer.
M\"{u}ller~\etal~\cite{muller2021self} design a term to push the vertices inside the mesh to the surface.
ProxyCap~\cite{zhang2023proxycap} introduces a contact-aware neural motion descent module such that the network can be aware of foot-ground contact and the misalignment with 2D observations. 

\subsection{Pose Prior and Shape Prior}
\label{sec:pose_prior}
The inherent ambiguity in lifting 2D observations to 3D space gives rise to the need for prior knowledge. Priors favor plausible predictions and rule out impossible ones, helping to restrict the outputs to a feasible distribution. Besides, priors play an indispensable role when 3D labels are not available.
Existing shape and pose priors are set heuristically by handcrafted designs or learned by generative models.
Classic generative models like the Gaussian Mixture Model (GMM), and Mixture of Experts (MoE) are used to discover patterns and correlations in data.
Compared to traditional machine learning methods, deep generative models such as Generative Adversarial Network (GAN)~\cite{goodfellow2014generative},  Variational Autoencoder (VAE)~\cite{kingma2013auto}, Normalizing Flows~\cite{rezende2015variational} are better qualified for prior modeling, especially when large-scale training data is available.
\rv{Priors can be treated as loss terms and added to objective functions in the training or iterative optimization processes. The decoder or generator of a generative prior can also be integrated into a regression network as a human mesh regressor.}

\paragraph{Handcrafted Prior}.
Priors can be designed empirically to achieve a certain direct-viewing effect. For example, known limb lengths are adopted~\cite{lee1985determination} and pose-dependent joint angle limits are explored~\cite{akhter2015pose}.
A pose prior in \cite{bogo2016keep, pavlakos2019expressive} is represented as the sum of the exponentials to penalize the unnatural bending in elbows and knees for the exponential value would soar if the rotations violate natural constraints.
As for shape prior, Bogo~\etal~\cite{bogo2016keep} compute the shape prior quadratically with the squared singular values estimated via PCA. 
A simple $L_2$ shape prior is adopted in \cite{kolotouros2019learning, zanfir2020weakly, pavlakos2019expressive,feng2021collaborative}, assuming $\bm{\beta}$ should stay near the neutral zero vector.

\paragraph{Gaussian Mixture Model (GMM)}.
Bogo~\etal~\cite{bogo2016keep} study the multi-model nature of the pose by fitting a mixture of 8 Gaussian distributions to a collection of reasonable pose parameters.
Xiang~\etal~\cite{xiang2019monocular} compute a Gaussian distribution for the pose parameters as a whole.

\paragraph{K-Means}.
G{\"u}ler~\etal~\cite{guler2019holopose} obtain $K$ representative angle values for each body joint after applying K-Means. The prediction outputs are restricted within the convex hull of the rotation clusters.
\rv{Rong~\etal~\cite{rong2022chasing} build a prototypical memory using K-Means to store multiple sets of mean parameters for regression initialization.}

\paragraph{Generative Adversarial Network (GAN)}.
Researchers first resort to GAN~\cite{goodfellow2014generative} to obtain adversarial priors. The discriminator is forced to distinguish between candidates produced by the network and real data~\cite{kanazawa2018end, xu2019denserac, jiang2020coherent, zhang2021body}. For instance,
Kanazawa~\etal~\cite{kanazawa2018end} assign a discriminator for shape and pose independently, and further train an adversarial prior for each joint.
Similarly, DenseRac~\cite{xu2019denserac} trains the discriminator with millions of synthetic samples to learn an admissible manifold of IUV representation.
\rv{Davydov~\etal~\cite{davydov2022adversarial} define a generator and a discriminator with the same architecture as the decoder in VPoser~\cite{pavlakos2019expressive} and the discriminator in HMR~\cite{kanazawa2018end}, respectively. After training, the GAN-based pose prior can be used in the optimization process to optimize a latent vector $z$ in the latent space. It can also serve as a drop-in human mesh regressor.}

\paragraph{Variational Autoencoder (VAE)}.
In a VAE~\cite{kingma2013auto}, the encoder compresses the data $\mathbf{x}$ into a latent distribution $P(\mathbf{z}\mid\mathbf{x})$. A latent variable $\mathbf{z}$ is sampled from $P(\mathbf{z})$, typically $\mathcal{N}(0, I)$. The decoder reconstructs $\hat{\mathbf{x}}$ given the hidden vector $\mathbf{z}$.
Pavlakos~\etal~\cite{pavlakos2019expressive} propose a VAE-based pose prior, VPoser, to learn a regularized latent distribution of human poses.
To employ VPoser in the optimization, the pose parameters are encoded to a latent variable $\mathbf{z}$, and a quadratic penalty is applied to $\mathbf{z}$.
A similar strategy is used in Georgakis~\etal~\cite{georgakis2020hierarchical} to obtain plausible poses.
Besides, the body and facial deformation in the GHUM/GHUML models~\cite{xu2020ghum} is also based on the latent space in VAE.

\paragraph{Normalizing Flows}.
Normalizing flows~\cite{rezende2015variational} are powerful in distribution approximation and efficient in derivation calculation.
Zanfir~\etal~\cite{zanfir2020weakly} introduce normalizing flows to model 3D human pose. They cascade multiple Real-NVP steps~\cite{dinh2016density} to build a model that embodies a flow-based prior for weakly-supervised training. Inspired by this, Biggs~\etal~\cite{biggs20203d} also adopt the Real-NVP architecture. Fan~\etal~\cite{fan2021revitalizing} design a normalizing flow using fully-connected layers.
The GHUM/GHUML models~\cite{xu2020ghum} rely on normalizing flows to represent skeleton kinematics. The authors also train a kinematic prior for hands and body based on normalizing flows.
ProHMR~\cite{kolotouros2021probabilistic} acts as an image-based pose prior to the fitting process, predicting the distribution of plausible poses given an input image. This distribution is modeled using conditional normalizing flows.

\paragraph{Diffusion Model}.
The Diffusion model~\cite{ho2020denoising} is a generative model based on stochastic processes.
Recently, there have been several approaches~\cite{Foo2023Distribution,Cho2023Generative,Zhang2023EgoHMR} applying diffusion models in the task of human mesh recovery.
Thanks to the probabilistic nature of diffusion models, these approaches can produce multiple hypotheses to handle the ambiguity in cases of occlusions.

\subsection{Motion Prior}
\label{sec:motion_prior}
Motions can be predicted to some extent since they have some patterns in nature.
Simply penalizing the velocity or acceleration of each joint will degrade motion naturalness.
Instead, priors based on recurrent models~\cite{kocabas2020vibe} and autoencoders~\cite{kaufmann2020convolutional,rempe2021humor,zhang2021learning} have larger temporal receptive fields to learn motion patterns.
VIBE~\cite{kocabas2020vibe} contains a motion discriminator and MPoser. The motion discriminator consists of multiple GRU layers to identify plausibility. MPoser, an extension of VPoser~\cite{pavlakos2019expressive} to temporal sequences, is based on sequential VAE.
Inspired by VIBE, He~\etal~\cite{he2021challencap} generate marker-based motion maps as input to a discriminator to obtain an adversarial motion prior.
In HuMoR~\cite{rempe2021humor}, the probability distribution of possible state transitions is formulated by a conditional variational autoencoder (CVAE). This dynamic prior is later used for robust test-time optimization.
LEMO~\cite{zhang2021learning} smooths the motion in the latent space of a convolutional autoencoder to reduce the pose jitters.
GLAMR~\cite{yuan2022glamr} contains a CVAE-based generative motion infiller to infill missing poses.
SimPoE~\cite{yuan2021simpoe} resorts to reinforcement learning and introduces a simulation-based motion modeling approach.
\rv{HM-VAE~\cite{li2021task} contains skeleton-based convolution, pooling, and unpooling operations. With the learned HM-VAE, one can refine noisy motion sequences by first projecting into the latent space and then decoding back.
Xu~\etal~\cite{xu2021exploring} exploit sequence-based and segment-based frequencies to compress input motions adaptively. The pretrained motion prior can be embedded into VIBE~\cite{kocabas2020vibe} in a video-to-mesh regression task.}


\ifarxiv
\begin{table*}[t]
  \centering
  \caption{Comparison of datasets involved in network training and evaluation. Each dataset provides images with paired 3D (pseudo) ground truth.}
  \vspace{-3mm}
  \label{tab:datasets}%
  \resizebox{\textwidth}{!}{
\begin{tabular}{clrrrrccc}
\toprule
\textbf{Type} & \multicolumn{1}{c}{\textbf{Dataset}} & \multicolumn{1}{c}{\textbf{\makecell{\#\\Frames}}} & \multicolumn{1}{c}{\textbf{\makecell{\#\\Scenes}}} & \multicolumn{1}{c}{\textbf{\makecell{\#\\Subjects}}} & \multicolumn{1}{c}{\textbf{\makecell{\# Subjects\\Per Frame}}} & \textbf{\makecell{In-\\the-\\wild}} & \textbf{\makecell{Mesh\\Type}} & \textbf{\makecell{Mesh\\Annotation\\Source}} \\
\midrule
\multirow{7}[2]{*}{\makecell{Rendered\\Datasets}} & \cellcolor[rgb]{ .851,  .851,  .851}SURREAL~\cite{varol2017learning} & \cellcolor[rgb]{ .851,  .851,  .851}6.5M & \cellcolor[rgb]{ .851,  .851,  .851}2,607 & \cellcolor[rgb]{ .851,  .851,  .851}145 & \cellcolor[rgb]{ .851,  .851,  .851}1 & \cellcolor[rgb]{ .851,  .851,  .851}- & \cellcolor[rgb]{ .851,  .851,  .851}SMPL & \cellcolor[rgb]{ .851,  .851,  .851}\cite{varol2017learning} \\
      & GTA-Human~\cite{cai2021playing} & 1.4M  & -     & $>$ 600 & 1     & -     & SMPL  & \cite{cai2021playing} \\
      & \cellcolor[rgb]{ .851,  .851,  .851}AGORA~\cite{patel2021agora} & \cellcolor[rgb]{ .851,  .851,  .851}17K & \cellcolor[rgb]{ .851,  .851,  .851}$>$ 350 & \cellcolor[rgb]{ .851,  .851,  .851}4,240 & \cellcolor[rgb]{ .851,  .851,  .851}5 $\sim$ 15 & \cellcolor[rgb]{ .851,  .851,  .851}- & \cellcolor[rgb]{ .851,  .851,  .851}SMPL-X & \cellcolor[rgb]{ .851,  .851,  .851}\cite{patel2021agora} \\
      & THUman2.0~\cite{yu2021function4d} & -     & -     & $\sim$ 200 & 1     & -     & SMPL-X & \cite{yu2021function4d} \\
      & \cellcolor[rgb]{ .851,  .851,  .851}MultiHuman~\cite{zheng2021deepmulticap} & \cellcolor[rgb]{ .851,  .851,  .851}- & \cellcolor[rgb]{ .851,  .851,  .851}- & \cellcolor[rgb]{ .851,  .851,  .851}$\sim$ 50 & \cellcolor[rgb]{ .851,  .851,  .851}1 $\sim$ 3 & \cellcolor[rgb]{ .851,  .851,  .851}- & \cellcolor[rgb]{ .851,  .851,  .851}SMPL-X & \cellcolor[rgb]{ .851,  .851,  .851}\cite{zheng2021deepmulticap} \\
      & \rv{HSPACE~\cite{bazavan2021hspace}} & \rv{1M} & \rv{100} & \rv{100} & \rv{avg. 5} & \rv{-} & \rv{GHUM} & \rv{\cite{bazavan2021hspace}} \\
      & \cellcolor[rgb]{ .851,  .851,  .851}\rv{BEDLAM~\cite{black2023bedlam}} & \cellcolor[rgb]{ .851,  .851,  .851}\rv{380K} & \cellcolor[rgb]{ .851,  .851,  .851}\rv{103} & \cellcolor[rgb]{ .851,  .851,  .851}\rv{-} & \cellcolor[rgb]{ .851,  .851,  .851}\rv{1 $\sim$ 10} & \cellcolor[rgb]{ .851,  .851,  .851}\rv{-} & \cellcolor[rgb]{ .851,  .851,  .851}\rv{SMPL-X} & \cellcolor[rgb]{ .851,  .851,  .851}\rv{\cite{black2023bedlam}} \\
\midrule
\multirow{4}[2]{*}{\makecell{Marker/Sensor-\\based MoCap}} & HumanEva~\cite{sigal2010humaneva} & 80K   & 1     & 4     & 1     & -     & -     & - \\
      & \cellcolor[rgb]{ .851,  .851,  .851}Human3.6M~\cite{ionescu2014human3} & \cellcolor[rgb]{ .851,  .851,  .851}3.6M & \cellcolor[rgb]{ .851,  .851,  .851}1 & \cellcolor[rgb]{ .851,  .851,  .851}11 & \cellcolor[rgb]{ .851,  .851,  .851}1 & \cellcolor[rgb]{ .851,  .851,  .851}- & \cellcolor[rgb]{ .851,  .851,  .851}SMPL & \cellcolor[rgb]{ .851,  .851,  .851}\cite{varol2017learning, kanazawa2018end, moon2020i2l} \\
      & \rv{Total Capture~\cite{trumble2017total}} & \rv{$\sim$ 1.9M} & \rv{1} & \rv{5} & \rv{1} & \rv{-} & \rv{-} & \rv{-} \\
      & \cellcolor[rgb]{ .851,  .851,  .851}3DPW~\cite{von2018recovering} & \cellcolor[rgb]{ .851,  .851,  .851}$>$ 51K & \cellcolor[rgb]{ .851,  .851,  .851}60 & \cellcolor[rgb]{ .851,  .851,  .851}7 & \cellcolor[rgb]{ .851,  .851,  .851}1 $\sim$ 2 & \cellcolor[rgb]{ .851,  .851,  .851}\checkmark & \cellcolor[rgb]{ .851,  .851,  .851}SMPL & \cellcolor[rgb]{ .851,  .851,  .851}\cite{von2018recovering} \\
\midrule
\multirow{11}[2]{*}{\makecell{Markerless\\Multiview\\MoCap}} & CMU Panoptic~\cite{joo2015panoptic} & 1.5M  & 1     & 40    & 3 $\sim$ 8 & -     & -     & - \\
      & \cellcolor[rgb]{ .851,  .851,  .851}MPI-INF-3DHP~\cite{mehta2017monocular} & \cellcolor[rgb]{ .851,  .851,  .851}$>$ 1.3M & \cellcolor[rgb]{ .851,  .851,  .851}1 & \cellcolor[rgb]{ .851,  .851,  .851}8 & \cellcolor[rgb]{ .851,  .851,  .851}1 & \cellcolor[rgb]{ .851,  .851,  .851}\rv{-} & \cellcolor[rgb]{ .851,  .851,  .851}SMPL & \cellcolor[rgb]{ .851,  .851,  .851}\cite{kolotouros2019learning} \\
      & MuCo-3DHP~\cite{mehta2018single} & 200K  & 1     & 8     & 1 $\sim$ 4 & -     &       & - \\
      & \cellcolor[rgb]{ .851,  .851,  .851}MuPoTs-3D~\cite{mehta2018single} & \cellcolor[rgb]{ .851,  .851,  .851}$>$ 8K & \cellcolor[rgb]{ .851,  .851,  .851}20 & \cellcolor[rgb]{ .851,  .851,  .851}8 & \cellcolor[rgb]{ .851,  .851,  .851}3 & \cellcolor[rgb]{ .851,  .851,  .851}\checkmark & \cellcolor[rgb]{ .851,  .851,  .851}- & \cellcolor[rgb]{ .851,  .851,  .851}- \\
      & MannequinChallenge~\cite{li2019learning} & 24,428 & 567   & 742   & 5     & \checkmark & SMPL  & \cite{leroy2020smply} \\
      & \cellcolor[rgb]{ .851,  .851,  .851}3DOH50K~\cite{zhang2020object} & \cellcolor[rgb]{ .851,  .851,  .851}51,600 & \cellcolor[rgb]{ .851,  .851,  .851}1 & \cellcolor[rgb]{ .851,  .851,  .851}- & \cellcolor[rgb]{ .851,  .851,  .851}1 & \cellcolor[rgb]{ .851,  .851,  .851}- & \cellcolor[rgb]{ .851,  .851,  .851}SMPL & \cellcolor[rgb]{ .851,  .851,  .851}\cite{zhang2020object} \\
      & Mirrored-Human~\cite{fang2021reconstructing} & 1.8M  & $>$ 200 & $>$ 200 & $\geq$ 1 & \checkmark & SMPL  & \cite{fang2021reconstructing} \\
      & \cellcolor[rgb]{ .851,  .851,  .851}MTC~\cite{xiang2019monocular} & \cellcolor[rgb]{ .851,  .851,  .851}834K & \cellcolor[rgb]{ .851,  .851,  .851}1 & \cellcolor[rgb]{ .851,  .851,  .851}40 & \cellcolor[rgb]{ .851,  .851,  .851}1 & \cellcolor[rgb]{ .851,  .851,  .851}- & \cellcolor[rgb]{ .851,  .851,  .851}- & \cellcolor[rgb]{ .851,  .851,  .851}- \\
      & EHF~\cite{pavlakos2019expressive} & 100   & 1     & 1     & 1     & -     & SMPL-X & \cite{pavlakos2019expressive} \\
      & \cellcolor[rgb]{ .851,  .851,  .851}HUMBI~\cite{yu2020humbi} & \cellcolor[rgb]{ .851,  .851,  .851}17.3M & \cellcolor[rgb]{ .851,  .851,  .851}1 & \cellcolor[rgb]{ .851,  .851,  .851}772 & \cellcolor[rgb]{ .851,  .851,  .851}1 & \cellcolor[rgb]{ .851,  .851,  .851}- & \cellcolor[rgb]{ .851,  .851,  .851}SMPL & \cellcolor[rgb]{ .851,  .851,  .851}\cite{yu2020humbi} \\
      & ZJU-MoCap~\cite{peng2021neural} & -     & 1     & 9     & 1     & -     & SMPL-X & \cite{zju3dv} \\
\midrule
\multirow{10}[2]{*}{\makecell{Datasets\\with\\Pseudo-\\3D Labels}} & \cellcolor[rgb]{ .851,  .851,  .851}LSP~\cite{johnson2010clustered} & \cellcolor[rgb]{ .851,  .851,  .851}2,000 & \cellcolor[rgb]{ .851,  .851,  .851}- & \cellcolor[rgb]{ .851,  .851,  .851}- & \cellcolor[rgb]{ .851,  .851,  .851}1 & \cellcolor[rgb]{ .851,  .851,  .851}\checkmark & \cellcolor[rgb]{ .851,  .851,  .851}SMPL & \cellcolor[rgb]{ .851,  .851,  .851}~\cite{lassner2017unite, kolotouros2019learning, joo2021exemplar} \\
      & LSP-Extended~\cite{johnson2011learning} & 10,000 & -     & -     & 1     & \checkmark & SMPL  & \cite{kolotouros2019learning, joo2021exemplar} \\
      & \cellcolor[rgb]{ .851,  .851,  .851}MSCOCO~\cite{lin2014microsoft} & \cellcolor[rgb]{ .851,  .851,  .851}38K & \cellcolor[rgb]{ .851,  .851,  .851}- & \cellcolor[rgb]{ .851,  .851,  .851}- & \cellcolor[rgb]{ .851,  .851,  .851}$\geq$ 1 & \cellcolor[rgb]{ .851,  .851,  .851}\checkmark & \cellcolor[rgb]{ .851,  .851,  .851}SMPL & \cellcolor[rgb]{ .851,  .851,  .851}\cite{kolotouros2019learning, joo2021exemplar} \\
      & MPII~\cite{andriluka20142d} & 24,920 & 3,913 & $>$ 40k & $\geq$ 1 & \checkmark & SMPL  & \cite{lassner2017unite, kolotouros2019learning, joo2021exemplar} \\
      & \cellcolor[rgb]{ .851,  .851,  .851}UP-3D~\cite{lassner2017unite} & \cellcolor[rgb]{ .851,  .851,  .851}8,515 & \cellcolor[rgb]{ .851,  .851,  .851}- & \cellcolor[rgb]{ .851,  .851,  .851}- & \cellcolor[rgb]{ .851,  .851,  .851}1 & \cellcolor[rgb]{ .851,  .851,  .851}\checkmark & \cellcolor[rgb]{ .851,  .851,  .851}SMPL & \cellcolor[rgb]{ .851,  .851,  .851}\cite{kolotouros2019learning, joo2021exemplar} \\
      & PoseTrack~\cite{andriluka2018posetrack} & 66,374 & 550   & 550   & $>$ 1 & \checkmark & SMPL  & \cite{joo2021exemplar} \\
      & \cellcolor[rgb]{ .851,  .851,  .851}SSP-3D~\cite{sengupta2020synthetic} & \cellcolor[rgb]{ .851,  .851,  .851}311 & \cellcolor[rgb]{ .851,  .851,  .851}62 & \cellcolor[rgb]{ .851,  .851,  .851}62 & \cellcolor[rgb]{ .851,  .851,  .851}1 & \cellcolor[rgb]{ .851,  .851,  .851}\checkmark & \cellcolor[rgb]{ .851,  .851,  .851}SMPL & \cellcolor[rgb]{ .851,  .851,  .851}\cite{sengupta2020synthetic} \\
      & OCHuman~\cite{zhang2019pose2seg} & 4,731 & -     & 8110  & $>$ 1 & \checkmark & SMPL  & \cite{joo2021exemplar} \\
      & \cellcolor[rgb]{ .851,  .851,  .851}MTP~\cite{muller2021self} & \cellcolor[rgb]{ .851,  .851,  .851}3,731 & \cellcolor[rgb]{ .851,  .851,  .851}- & \cellcolor[rgb]{ .851,  .851,  .851}148 & \cellcolor[rgb]{ .851,  .851,  .851}1 & \cellcolor[rgb]{ .851,  .851,  .851}\checkmark & \cellcolor[rgb]{ .851,  .851,  .851}SMPL-X & \cellcolor[rgb]{ .851,  .851,  .851}\cite{muller2021self} \\
      & \rv{Ubody~\cite{lin2023one}} & \rv{$>$ 1,050K} & \rv{-} & \rv{-} & \rv{$\geq$ 1} & \rv{\checkmark} & \rv{SMPL-X} & \rv{\cite{lin2023one}} \\
\bottomrule
\end{tabular}%

  }
 \vspace{-3mm}
\end{table*}%

\begin{table*}[!htbp]
	\centering
	\caption{Evaluation of the body recovery methods on 3DPW~\cite{von2018recovering} and Human3.6M~\cite{ionescu2014human3} datasets. The comparison is not completely fair considering the factors of backbones, output types, the quality of pseudo labels, dataset selection, and training strategy. \textsuperscript{$\natural$} represents approaches dealing with the multi-person 3D mesh recovery task. $^\dagger$ denotes the approaches using training data from 3DPW. Please refer to \textsection~\ref{sec:benchmark} for a comprehensive discussion.}
	\vspace{-2mm}
	\resizebox{\textwidth}{!}{
\begin{tabular}{cl|r|c|c|ccc|cc}
\toprule
\multirow{2}[4]{*}{} & \multicolumn{1}{c|}{\multirow{2}[4]{*}{\textbf{Method}}} & \multicolumn{1}{c|}{\multirow{2}[4]{*}{\textbf{Publication}}} & \multirow{2}[4]{*}{\textbf{\makecell{Output\\Type}}} & \multirow{2}[4]{*}{\textbf{\makecell{Pseudo-GT\\(2D Datasets)}}} & \multicolumn{3}{c|}{\textbf{3DPW}} & \multicolumn{2}{c}{\textbf{Human3.6M}} \\
\cmidrule{6-10}      &       &       &       &       & \textbf{MPJPE} & \textbf{PA-MPJPE} & \textbf{PVE} & \textbf{MPJPE} & \textbf{PA-MPJPE} \\
\midrule
\multirow{43}[2]{*}{\rotatebox[origin=c]{90}{\textbf{Frame-based}}} & \cellcolor[rgb]{ .851,  .851,  .851}Pavlakos~\etal~\cite{pavlakos2018learning} & \cellcolor[rgb]{ .851,  .851,  .851}CVPR'18 & \cellcolor[rgb]{ .851,  .851,  .851}Parameters & \cellcolor[rgb]{ .851,  .851,  .851}\cite{lassner2017unite} & \cellcolor[rgb]{ .851,  .851,  .851}- & \cellcolor[rgb]{ .851,  .851,  .851}- & \cellcolor[rgb]{ .851,  .851,  .851}- & \cellcolor[rgb]{ .851,  .851,  .851}- & \cellcolor[rgb]{ .851,  .851,  .851}75.9 \\
      & HMR~\cite{kanazawa2018end} & CVPR'18 & Parameters & -     & 130.0 & 76.7  & -     & 88.0  & 56.8 \\
      & \cellcolor[rgb]{ .851,  .851,  .851}NBF~\cite{omran2018neural} & \cellcolor[rgb]{ .851,  .851,  .851}3DV'18 & \cellcolor[rgb]{ .851,  .851,  .851}Parameters & \cellcolor[rgb]{ .851,  .851,  .851}\cite{lassner2017unite} & \cellcolor[rgb]{ .851,  .851,  .851}- & \cellcolor[rgb]{ .851,  .851,  .851}- & \cellcolor[rgb]{ .851,  .851,  .851}- & \cellcolor[rgb]{ .851,  .851,  .851}- & \cellcolor[rgb]{ .851,  .851,  .851}59.9 \\
      & GraphCMR~\cite{kolotouros2019convolutional} & CVPR'19 & Vertices & \cite{lassner2017unite} & -     & 70.2  & -     & -     & 50.1 \\
      & \cellcolor[rgb]{ .851,  .851,  .851}HoloPose~\cite{guler2019holopose} & \cellcolor[rgb]{ .851,  .851,  .851}CVPR'19 & \cellcolor[rgb]{ .851,  .851,  .851}Parameters & \cellcolor[rgb]{ .851,  .851,  .851}- & \cellcolor[rgb]{ .851,  .851,  .851}- & \cellcolor[rgb]{ .851,  .851,  .851}- & \cellcolor[rgb]{ .851,  .851,  .851}- & \cellcolor[rgb]{ .851,  .851,  .851}60.3 & \cellcolor[rgb]{ .851,  .851,  .851}46.5 \\
      & DenseRac~\cite{xu2019denserac} & ICCV'19 & Parameters & -     & -     & -     & -     & 76.8  & 48.0 \\
      & \cellcolor[rgb]{ .851,  .851,  .851}SPIN~\cite{kolotouros2019learning} & \cellcolor[rgb]{ .851,  .851,  .851}ICCV'21 & \cellcolor[rgb]{ .851,  .851,  .851}Parameters & \cellcolor[rgb]{ .851,  .851,  .851}\cite{kolotouros2019learning} & \cellcolor[rgb]{ .851,  .851,  .851}96.9 & \cellcolor[rgb]{ .851,  .851,  .851}59.2 & \cellcolor[rgb]{ .851,  .851,  .851}135.1 & \cellcolor[rgb]{ .851,  .851,  .851}62.5 & \cellcolor[rgb]{ .851,  .851,  .851}41.1 \\
      & Jiang~\etal~\cite{jiang2020coherent}{\textsuperscript{$\natural$}} & CVPR'20 & Parameters & -     & -     & -     & -     & -     & 52.7 \\
      & \cellcolor[rgb]{ .851,  .851,  .851}Zhang~\etal~\cite{zhang2020object} & \cellcolor[rgb]{ .851,  .851,  .851}CVPR'20 & \cellcolor[rgb]{ .851,  .851,  .851}Position map & \cellcolor[rgb]{ .851,  .851,  .851}- & \cellcolor[rgb]{ .851,  .851,  .851}- & \cellcolor[rgb]{ .851,  .851,  .851}- & \cellcolor[rgb]{ .851,  .851,  .851}- & \cellcolor[rgb]{ .851,  .851,  .851}- & \cellcolor[rgb]{ .851,  .851,  .851}41.7 \\
      & DecoMR~\cite{zeng20203d} & CVPR'20 & Position map & \cite{lassner2017unite} & -     & 61.7  & -     & -     & 39.3 \\
      & \cellcolor[rgb]{ .851,  .851,  .851}Zanfir~\etal~\cite{zanfir2020weakly} & \cellcolor[rgb]{ .851,  .851,  .851}ECCV'20 & \cellcolor[rgb]{ .851,  .851,  .851}Parameters & \cellcolor[rgb]{ .851,  .851,  .851}- & \cellcolor[rgb]{ .851,  .851,  .851}90.0 & \cellcolor[rgb]{ .851,  .851,  .851}57.1 & \cellcolor[rgb]{ .851,  .851,  .851}- & \cellcolor[rgb]{ .851,  .851,  .851}- & \cellcolor[rgb]{ .851,  .851,  .851}- \\
      & LearnedGD~\cite{song2020human} & ECCV'20 & Parameters & -     & -     & 55.9  & -     & -     & 56.4 \\
      & \cellcolor[rgb]{ .851,  .851,  .851}Pose2Mesh~\cite{choi2020pose2mesh} & \cellcolor[rgb]{ .851,  .851,  .851}ECCV'20 & \cellcolor[rgb]{ .851,  .851,  .851}Vertices & \cellcolor[rgb]{ .851,  .851,  .851}\cite{kolotouros2019learning} & \cellcolor[rgb]{ .851,  .851,  .851}89.2 & \cellcolor[rgb]{ .851,  .851,  .851}58.9 & \cellcolor[rgb]{ .851,  .851,  .851}- & \cellcolor[rgb]{ .851,  .851,  .851}64.9 & \cellcolor[rgb]{ .851,  .851,  .851}47.0 \\
      & HKMR~\cite{georgakis2020hierarchical} & ECCV'20 & Parameters & -     & -     & -     & -     & 59.6  & 43.2 \\
      & \cellcolor[rgb]{ .851,  .851,  .851}I2L-MeshNet~\cite{moon2020i2l} & \cellcolor[rgb]{ .851,  .851,  .851}ECCV'20 & \cellcolor[rgb]{ .851,  .851,  .851}Vertices & \cellcolor[rgb]{ .851,  .851,  .851}\cite{moon2020i2l} & \cellcolor[rgb]{ .851,  .851,  .851}93.2 & \cellcolor[rgb]{ .851,  .851,  .851}57.7 & \cellcolor[rgb]{ .851,  .851,  .851}- & \cellcolor[rgb]{ .851,  .851,  .851}55.7 & \cellcolor[rgb]{ .851,  .851,  .851}41.1 \\
      & DaNet~\cite{zhang2019danet} & MM'19 & Parameters & \cite{lassner2017unite} & 85.5  & 54.8  & 110.8 & -     & 40.5 \\
      & \cellcolor[rgb]{ .851,  .851,  .851}Hand4Whole~\cite{moon2022accurate} & \cellcolor[rgb]{ .851,  .851,  .851}CVPRW'22 & \cellcolor[rgb]{ .851,  .851,  .851}Parameters & \cellcolor[rgb]{ .851,  .851,  .851}\cite{moon2022neuralannot} & \cellcolor[rgb]{ .851,  .851,  .851}86.6 & \cellcolor[rgb]{ .851,  .851,  .851}54.4 & \cellcolor[rgb]{ .851,  .851,  .851}- & \cellcolor[rgb]{ .851,  .851,  .851}71.0 & \cellcolor[rgb]{ .851,  .851,  .851}47.4 \\
      & HybrIK~\cite{li2021hybrik} & CVPR'21 & Parameters & -     & 80.0  & 48.8  & 94.5  & -     & - \\
      & \cellcolor[rgb]{ .851,  .851,  .851}METRO~\cite{lin2021end} & \cellcolor[rgb]{ .851,  .851,  .851}CVPR'21 & \cellcolor[rgb]{ .851,  .851,  .851}Vertices & \cellcolor[rgb]{ .851,  .851,  .851}\cite{lassner2017unite, kolotouros2019learning} & \cellcolor[rgb]{ .851,  .851,  .851}77.1 & \cellcolor[rgb]{ .851,  .851,  .851}47.9 & \cellcolor[rgb]{ .851,  .851,  .851}88.2 & \cellcolor[rgb]{ .851,  .851,  .851}54.0 & \cellcolor[rgb]{ .851,  .851,  .851}36.7 \\
      & Sengupta~\etal~\cite{sengupta2021probabilistic} & CVPR'21 & Probabilistic & \cite{lassner2017unite} & -     & 61.0  & -     & -     & - \\
      & \cellcolor[rgb]{ .851,  .851,  .851}BMP~\cite{zhang2021body}{\textsuperscript{$\natural$}} & \cellcolor[rgb]{ .851,  .851,  .851}CVPR'21 & \cellcolor[rgb]{ .851,  .851,  .851}Parameters & \cellcolor[rgb]{ .851,  .851,  .851}- & \cellcolor[rgb]{ .851,  .851,  .851}104.1 & \cellcolor[rgb]{ .851,  .851,  .851}63.8 & \cellcolor[rgb]{ .851,  .851,  .851}119.3 & \cellcolor[rgb]{ .851,  .851,  .851}- & \cellcolor[rgb]{ .851,  .851,  .851}51.3 \\
      & HUND~\cite{zanfir2021neural} & CVPR'21 & Parameters & -     & 81.4  & 57.5  & -     & 69.5  & 52.6 \\
      & \cellcolor[rgb]{ .851,  .851,  .851}EFT~\cite{joo2021exemplar} & \cellcolor[rgb]{ .851,  .851,  .851}3DV'21 & \cellcolor[rgb]{ .851,  .851,  .851}Parameters & \cellcolor[rgb]{ .851,  .851,  .851}\cite{joo2021exemplar} & \cellcolor[rgb]{ .851,  .851,  .851}- & \cellcolor[rgb]{ .851,  .851,  .851}54.2 & \cellcolor[rgb]{ .851,  .851,  .851}- & \cellcolor[rgb]{ .851,  .851,  .851}- & \cellcolor[rgb]{ .851,  .851,  .851}43.7 \\
      & ProHMR~\cite{kolotouros2021probabilistic} & ICCV'21 & Probabilistic & \cite{kolotouros2019learning} & -     & 59.8  & -     & -     & 41.2 \\
      & \cellcolor[rgb]{ .851,  .851,  .851}DSR~\cite{dwivedi2021learning} & \cellcolor[rgb]{ .851,  .851,  .851}ICCV'21 & \cellcolor[rgb]{ .851,  .851,  .851}Parameters & \cellcolor[rgb]{ .851,  .851,  .851}\cite{joo2021exemplar} & \cellcolor[rgb]{ .851,  .851,  .851}91.7 & \cellcolor[rgb]{ .851,  .851,  .851}54.1 & \cellcolor[rgb]{ .851,  .851,  .851}105.8 & \cellcolor[rgb]{ .851,  .851,  .851}60.9 & \cellcolor[rgb]{ .851,  .851,  .851}40.3 \\
      & ROMP~\cite{sun2021monocular}{\textsuperscript{$\natural$}} & ICCV'21 & Parameters & \cite{kolotouros2019learning} & 89.3  & 53.5  & 103.1 & -     & - \\
      & \cellcolor[rgb]{ .851,  .851,  .851}Graphormer~\cite{lin2021mesh} & \cellcolor[rgb]{ .851,  .851,  .851}ICCV'21 & \cellcolor[rgb]{ .851,  .851,  .851}Vertices & \cellcolor[rgb]{ .851,  .851,  .851}- & \cellcolor[rgb]{ .851,  .851,  .851}74.7 & \cellcolor[rgb]{ .851,  .851,  .851}45.6 & \cellcolor[rgb]{ .851,  .851,  .851}87.7 & \cellcolor[rgb]{ .851,  .851,  .851}51.2 & \cellcolor[rgb]{ .851,  .851,  .851}34.5 \\
      & THUNDR~\cite{zanfir2021thundr} & ICCV'21 & Parameters & -     & 74.8  & 51.5  & 88.0  & 55.0  & 39.8 \\
      & \cellcolor[rgb]{ .851,  .851,  .851}PyMAF~\cite{zhang2021pymaf} & \cellcolor[rgb]{ .851,  .851,  .851}ICCV'21 & \cellcolor[rgb]{ .851,  .851,  .851}Parameters & \cellcolor[rgb]{ .851,  .851,  .851}\cite{kolotouros2019learning} & \cellcolor[rgb]{ .851,  .851,  .851}92.8 & \cellcolor[rgb]{ .851,  .851,  .851}58.9 & \cellcolor[rgb]{ .851,  .851,  .851}110.1 & \cellcolor[rgb]{ .851,  .851,  .851}57.7 & \cellcolor[rgb]{ .851,  .851,  .851}40.5 \\
      & SPEC~\cite{kocabas2021spec} & ICCV'21 & Parameters & \cite{joo2021exemplar} & -     & 53.2  & -     & -     & - \\
      & \cellcolor[rgb]{ .851,  .851,  .851}PARE~\cite{kocabas2021pare} $^\dagger$ & \cellcolor[rgb]{ .851,  .851,  .851}ICCV'21 & \cellcolor[rgb]{ .851,  .851,  .851}Parameters & \cellcolor[rgb]{ .851,  .851,  .851}\cite{joo2021exemplar} & \cellcolor[rgb]{ .851,  .851,  .851}74.5 & \cellcolor[rgb]{ .851,  .851,  .851}46.5 & \cellcolor[rgb]{ .851,  .851,  .851}88.6 & \cellcolor[rgb]{ .851,  .851,  .851}- & \cellcolor[rgb]{ .851,  .851,  .851}- \\
      & BEV~\cite{sun2021putting}{\textsuperscript{$\natural$}} & CVPR'22 & Parameters & \cite{joo2021exemplar} & 78.5  & 46.9  & 92.3  & -     & - \\
      & \cellcolor[rgb]{ .851,  .851,  .851}CLIFF~\cite{li2022cliff} $^\dagger$ & \cellcolor[rgb]{ .851,  .851,  .851}ECCV'22 & \cellcolor[rgb]{ .851,  .851,  .851}Parameters & \cellcolor[rgb]{ .851,  .851,  .851}\cite{li2022cliff} & \cellcolor[rgb]{ .851,  .851,  .851}69.0 & \cellcolor[rgb]{ .851,  .851,  .851}43.0  & \cellcolor[rgb]{ .851,  .851,  .851}81.2 & \cellcolor[rgb]{ .851,  .851,  .851} 47.1 & \cellcolor[rgb]{ .851,  .851,  .851} 32.7 \\
      & FastMETRO~\cite{cho2022cross} $^\dagger$ & ECCV'22 & Vertices & -     & 73.5  & 44.6  & 84.1  & 52.2  & 33.7 \\
      & \cellcolor[rgb]{ .851,  .851,  .851}Cha~\cite{cha2022multi}{\textsuperscript{$\natural$}} & \cellcolor[rgb]{ .851,  .851,  .851}ECCV'22 & \cellcolor[rgb]{ .851,  .851,  .851}Parameters & \cellcolor[rgb]{ .851,  .851,  .851}- & \cellcolor[rgb]{ .851,  .851,  .851}66.0 & \cellcolor[rgb]{ .851,  .851,  .851}39.0 & \cellcolor[rgb]{ .851,  .851,  .851}76.3 & \cellcolor[rgb]{ .851,  .851,  .851} & \cellcolor[rgb]{ .851,  .851,  .851} \\
      & PyMAF-X~\cite{zhang2023pymafx} $^\dagger$ & TPAMI'23 & Parameters & \cite{joo2021exemplar} &  74.2 & 45.3  & 87.0  & -     & - \\
      & \cellcolor[rgb]{ .851,  .851,  .851}POTTER~\cite{zheng2023potter} $^\dagger$ & \cellcolor[rgb]{ .851,  .851,  .851}CVPR'23 & \cellcolor[rgb]{ .851,  .851,  .851}Parameters & \cellcolor[rgb]{ .851,  .851,  .851}- & \cellcolor[rgb]{ .851,  .851,  .851}75.0 & \cellcolor[rgb]{ .851,  .851,  .851} 44.8 & \cellcolor[rgb]{ .851,  .851,  .851} 87.4 & \cellcolor[rgb]{ .851,  .851,  .851}56.5 & \cellcolor[rgb]{ .851,  .851,  .851}35.1 \\
      & ProPose~\cite{fang2023learning} $^\dagger$ & CVPR'23 & Parameters & -     & 68.3  & 40.6  & 79.4  & 45.7  & 29.1 \\
      & \cellcolor[rgb]{ .851,  .851,  .851}NIKI~\cite{li2023niki} $^\dagger$ & \cellcolor[rgb]{ .851,  .851,  .851}CVPR'23 & \cellcolor[rgb]{ .851,  .851,  .851}Parameters & \cellcolor[rgb]{ .851,  .851,  .851}- & \cellcolor[rgb]{ .851,  .851,  .851}71.3 & \cellcolor[rgb]{ .851,  .851,  .851} 40.6 & \cellcolor[rgb]{ .851,  .851,  .851} 86.6 & \cellcolor[rgb]{ .851,  .851,  .851}- & \cellcolor[rgb]{ .851,  .851,  .851}- \\
      & PLIKS~\cite{shetty2023pliks} $^\dagger$ & CVPR'23 & Parameters & \cite{joo2021exemplar} & 60.5  & 38.5  &  73.3 & 47.0  & 34.5 \\
      & \cellcolor[rgb]{ .851,  .851,  .851}PointHMR~\cite{kim2023sampling} $^\dagger$ & \cellcolor[rgb]{ .851,  .851,  .851}CVPR'23 & \cellcolor[rgb]{ .851,  .851,  .851}Vertices & \cellcolor[rgb]{ .851,  .851,  .851}- & \cellcolor[rgb]{ .851,  .851,  .851}73.9 & \cellcolor[rgb]{ .851,  .851,  .851} 44.9 & \cellcolor[rgb]{ .851,  .851,  .851}85.5 & \cellcolor[rgb]{ .851,  .851,  .851}48.3 & \cellcolor[rgb]{ .851,  .851,  .851} 32.9 \\
      & ImpHMR~\cite{cho2023implicit} $^\dagger$ & CVPR'23 & Parameters & -     & 74.3  & 45.4  & 87.1  & -     & - \\
      & \cellcolor[rgb]{ .851,  .851,  .851}VirtualMarker~\cite{ma20233d} $^\dagger$ & \cellcolor[rgb]{ .851,  .851,  .851}CVPR'23 & \cellcolor[rgb]{ .851,  .851,  .851}Parameters & \cellcolor[rgb]{ .851,  .851,  .851}- & \cellcolor[rgb]{ .851,  .851,  .851}67.5 & \cellcolor[rgb]{ .851,  .851,  .851}41.3 & \cellcolor[rgb]{ .851,  .851,  .851}77.9 & \cellcolor[rgb]{ .851,  .851,  .851}47.3 & \cellcolor[rgb]{ .851,  .851,  .851}32.0 \\
\midrule
\multirow{12}[2]{*}{\rotatebox[origin=c]{90}{\textbf{Temporal}}} & HMMR~\cite{kanazawa2019learning} & CVPR'19 & Parameters & -     & 116.5 & 72.6  & 139.3 & -     & 56.9 \\
      & \cellcolor[rgb]{ .851,  .851,  .851}Arnab~\etal~\cite{arnab2019exploiting} & \cellcolor[rgb]{ .851,  .851,  .851}CVPR'19 & \cellcolor[rgb]{ .851,  .851,  .851}Parameters & \cellcolor[rgb]{ .851,  .851,  .851}- & \cellcolor[rgb]{ .851,  .851,  .851}- & \cellcolor[rgb]{ .851,  .851,  .851}72.2 & \cellcolor[rgb]{ .851,  .851,  .851}- & \cellcolor[rgb]{ .851,  .851,  .851}77.8 & \cellcolor[rgb]{ .851,  .851,  .851}54.3 \\
      & DSD-SATN~\cite{sun2019human} & ICCV'19 & Parameters & -     & -     & 69.5  & -     & 59.1  & 42.4 \\
      & \cellcolor[rgb]{ .851,  .851,  .851}Sim2Real~\cite{doersch2019sim2real} & \cellcolor[rgb]{ .851,  .851,  .851}NeurIPS'19 & \cellcolor[rgb]{ .851,  .851,  .851}Parameters & \cellcolor[rgb]{ .851,  .851,  .851}\cite{lassner2017unite} & \cellcolor[rgb]{ .851,  .851,  .851}- & \cellcolor[rgb]{ .851,  .851,  .851}74.7 & \cellcolor[rgb]{ .851,  .851,  .851}- & \cellcolor[rgb]{ .851,  .851,  .851}- & \cellcolor[rgb]{ .851,  .851,  .851}- \\
      & VIBE~\cite{kocabas2020vibe} $^\dagger$ & CVPR'20 & Parameters & -     & 82.9  & 51.9  & 99.1  &  65.6 & 41.4 \\
      & \cellcolor[rgb]{ .851,  .851,  .851}MEVA~\cite{luo20203d} $^\dagger$ & \cellcolor[rgb]{ .851,  .851,  .851}ACCV'20 & \cellcolor[rgb]{ .851,  .851,  .851}Parameters & \cellcolor[rgb]{ .851,  .851,  .851}- & \cellcolor[rgb]{ .851,  .851,  .851}86.9 & \cellcolor[rgb]{ .851,  .851,  .851}54.7 & \cellcolor[rgb]{ .851,  .851,  .851}- & \cellcolor[rgb]{ .851,  .851,  .851}- & \cellcolor[rgb]{ .851,  .851,  .851}- \\
      & TCMR~\cite{choi2021beyond} $^\dagger$ & CVPR'21 & Parameters & -     & 95.0  & 55.8  & 111.3 & 62.3  & 41.1 \\
      & \cellcolor[rgb]{ .851,  .851,  .851}Lee~\etal~\cite{lee2021uncertainty} $^\dagger$ & \cellcolor[rgb]{ .851,  .851,  .851}ICCV'21 & \cellcolor[rgb]{ .851,  .851,  .851}Parameters & \cellcolor[rgb]{ .851,  .851,  .851}- & \cellcolor[rgb]{ .851,  .851,  .851}92.8 & \cellcolor[rgb]{ .851,  .851,  .851}52.2 & \cellcolor[rgb]{ .851,  .851,  .851}106.1 & \cellcolor[rgb]{ .851,  .851,  .851}58.4 & \cellcolor[rgb]{ .851,  .851,  .851}38.4 \\
      & MAED~\cite{wan2021encoder} $^\dagger$ & ICCV'21 & Parameters & -     & 79.1  & 45.7  &  92.6 & 56.4  & 38.7 \\
      & \cellcolor[rgb]{ .851,  .851,  .851}MPS-Net~\cite{wei2022capturing} $^\dagger$ & \cellcolor[rgb]{ .851,  .851,  .851}CVPR'22 & \cellcolor[rgb]{ .851,  .851,  .851}Parameters & \cellcolor[rgb]{ .851,  .851,  .851}- & \cellcolor[rgb]{ .851,  .851,  .851}91.6 & \cellcolor[rgb]{ .851,  .851,  .851}54.0 & \cellcolor[rgb]{ .851,  .851,  .851}109.6 & \cellcolor[rgb]{ .851,  .851,  .851}69.4 & \cellcolor[rgb]{ .851,  .851,  .851}47.4 \\
      & GLoT~\cite{shen2023global} $^\dagger$ & CVPR'23 & Parameters & -     & 80.7  & 50.6  & 96.3  & 67.0  & 46.3 \\
      & \cellcolor[rgb]{ .851,  .851,  .851}PSVT~\cite{qiu2023psvt}{\textsuperscript{$\natural$}} $^\dagger$ & \cellcolor[rgb]{ .851,  .851,  .851}CVPR'23 & \cellcolor[rgb]{ .851,  .851,  .851}Parameters & \cellcolor[rgb]{ .851,  .851,  .851}- & \cellcolor[rgb]{ .851,  .851,  .851}73.1 & \cellcolor[rgb]{ .851,  .851,  .851}43.5 & \cellcolor[rgb]{ .851,  .851,  .851}84.0 & \cellcolor[rgb]{ .851,  .851,  .851}- & \cellcolor[rgb]{ .851,  .851,  .851}- \\
\bottomrule
\end{tabular}%
}
\label{tab:h36m_3dpw_sota}
\vspace{-2mm}
\end{table*}{}

\begin{table*}[th]
	\centering
	\caption{Evaluation of the whole-body recovery methods on the AGORA~\cite{patel2021agora} dataset. \textbf{FB}, \textbf{B}, \textbf{F}, \textbf{LH/RH} denote evaluation results on the full-body, body, face, left-hand/righ-hand, respectively. }
	\vspace{-3mm}
	\resizebox{\textwidth}{!}{
\begin{tabular}{l|c|rr|rr|rrrr|rrrr}
\toprule
      &       & \multicolumn{2}{c|}{\textbf{NMVE}} & \multicolumn{2}{c|}{\textbf{NMJE}} & \multicolumn{4}{c|}{\textbf{MVE}} & \multicolumn{4}{c}{\textbf{MPJPE}} \\
\cmidrule{3-14}\multicolumn{1}{c|}{\textbf{Method}} & \textbf{\makecell{Train on\\AGORA}} & \multicolumn{1}{c}{\textbf{FB}} & \multicolumn{1}{c|}{\textbf{B}} & \multicolumn{1}{c}{\textbf{FB}} & \multicolumn{1}{c|}{\textbf{B}} & \multicolumn{1}{c}{\textbf{FB}} & \multicolumn{1}{c}{\textbf{B}} & \multicolumn{1}{c}{\textbf{F}} & \multicolumn{1}{c|}{\textbf{LH/RH}} & \multicolumn{1}{c}{\textbf{FB}} & \multicolumn{1}{c}{\textbf{B}} & \multicolumn{1}{c}{\textbf{F}} & \multicolumn{1}{c}{\textbf{LH/RH9}} \\
\midrule
\rowcolor[rgb]{ .851,  .851,  .851} SMPLify-X~\cite{pavlakos2019expressive} & -     & 333.1 & 263.3 & 326.5 & 256.5 & 236.5 & 187   & 48.9  & 48.3/51.4 & 231.8 & 182.1 & 52.9  & 46.5/49.6 \\
ExPose~\cite{choutas2020monocular} & \xmark & 265.0 & 184.8 & 263.3 & 183.4 & 217.3 & 151.5 & 51.1  & 74.9/71.3 & 215.9 & 150.4 & 55.2  & 72.5/68.8 \\
\rowcolor[rgb]{ .851,  .851,  .851} FrankMocap~\cite{rong2020frankmocap} & \xmark &       & 207.8 &       & 204.0 &       & 168.3 &       & 54.7/55.7 &       & 165.2 &       & 52.3/53.1 \\
PIXIE~\cite{feng2021collaborative} & \xmark & 233.9 & 173.4 & 230.9 & 171.1 & 191.8 & 142.2 & 50.2  & 49.5/49.0 & 189.3 & 140.3 & 54.5  & 46.4/46.0 \\
\rowcolor[rgb]{ .851,  .851,  .851} Hand4Whole~\cite{moon2022accurate} & \cmark & 144.1 & 96.0  & 141.1 & 92.7  & 135.5 & 90.2  & 41.6  & 46.3/48.1 & 132.6 & 87.1  & 46.1  & 44.3/46.2 \\
PyMAF-X~\cite{zhang2023pymafx} & \cmark & 141.2 & 94.4  & 140.0 & 93.5  & 125.7 & 84.0  & 35.0  & 44.6/45.6 & 124.6 & 83.2  & 37.9  & 42.5/43.7 \\
\rowcolor[rgb]{ .851,  .851,  .851} OSX~\cite{lin2023one} & \cmark & 130.6 & 85.3  & 127.6 & 83.3  & 122.8 & 80.2  & 36.2  & 45.4/46.1 & 119.9 & 78.3  & 37.9  & 43.0/43.9 \\
HybrIK-X~\cite{li2023hybrikx} & \cmark & 120.5 & 73.7  & 115.7 & 72.3  & 112.1 & 68.5  & 37.0  & 46.7/47.0 & 107.6 & 67.2  & 38.5  & 41.2/41.4 \\
\bottomrule
\end{tabular}%
}
\label{tab:agora_sota}
\vspace{-3mm}
\end{table*}{}

\begin{table}[th]
	\centering
	\caption{Evaluation of the whole-body recovery methods on the EHF~\cite{pavlakos2019expressive} dataset.}
	\vspace{-3mm}
	\resizebox{0.48\textwidth}{!}{
\begin{tabular}{l|c|rrr|rr}
\toprule
      &       & \multicolumn{3}{c|}{\textbf{PA-V2V}} & \multicolumn{2}{c}{\textbf{PA-MPJPE}} \\
\midrule
\multicolumn{1}{c|}{\textbf{Methods}} & \textbf{\makecell{Body\\Model}} & \multicolumn{1}{c}{\textbf{All}} & \multicolumn{1}{c}{\textbf{Hands}} & \multicolumn{1}{c|}{\textbf{Face}} & \multicolumn{1}{c}{\textbf{Body}} & \multicolumn{1}{c}{\textbf{Hands}} \\
\midrule
\rowcolor[rgb]{ .851,  .851,  .851} MTC~\cite{xiang2019monocular} & Adam  & 67.2  & -     & -     & 107.8 & 16.7 \\
\midrule
SMPLify-X~\cite{pavlakos2019expressive} & SMPL-X & 65.3  & 12.3  & 6.3   & 87.6  & 12.9 \\
\rowcolor[rgb]{ .851,  .851,  .851} ExPose~\cite{choutas2020monocular} & SMPL-X & 54.5  & 12.8  & 5.8   & 62.8  & 13.1 \\
FrankMocap~\cite{rong2020frankmocap} & SMPL-X & 57.5  & 12.6  & -     & 62.3  & 12.9 \\
\rowcolor[rgb]{ .851,  .851,  .851} PIXIE~\cite{feng2021collaborative} & SMPL-X & 55.0  & 11.1  & 4.6   & 61.5  & 11.6 \\
Hand4Whole~\cite{moon2022accurate} & SMPL-X & 50.3  & 10.8  & 5.8   & 60.4  & 10.8 \\
\rowcolor[rgb]{ .851,  .851,  .851} PyMAF-X~\cite{zhang2023pymafx} & SMPL-X & 50.2  & 10.2  & 5.5   & 52.8  & 10.3 \\
\bottomrule
\end{tabular}%
	}
	\label{tab:ehf_sota}
	\vspace{-3mm}
\end{table}{}

\section{Datasets}
\label{sec:datasets}
In this section, we focus on the commonly-used datasets. First, we introduce the acquisition of human mesh annotations.
Then, we give brief descriptions of the commonly used datasets.

\subsection{The Acquisition of Mesh Annotations}
Obtaining samples paired with 3D mesh labels is not easy. 
The most precise image-label pairs are generated by \textbf{rendering} 3D body models~\cite{varol2017learning} or human scans~\cite{patel2021agora} to images.
The lack of realism remains a major issue in these synthetic images.
In order to collect real images and obtain corresponding 3D labels, marker/sensor-based~\cite{ionescu2014human3,von2018recovering} and marker-less~\cite{joo2015panoptic,huang2017towards} MoCap systems are deployed to capture body motions.
\textbf{Marker/sensor-based systems} attach reflective markers or Inertial Measurement Units (IMU) to the subjects' bodies and track them over time. These 3D sparse point sets are later processed by MoSh~\cite{loper2014mosh} to fit a body mesh.
\textbf{Marker-less systems} capture person images from multiple cameras, where 2D cues are further fitted to the body mesh by exploiting multi-view geometry.
The MoCap data is generally limited to constrained environments and lacks the diversity of subjects and actions.

To obtain mesh annotations for in-the-wild images, researchers fit the body model to image evidence to generate \textbf{pseudo-3D labels} in semi-automatic~\cite{bogo2016keep,lassner2017unite} or full-automatic~\cite{kolotouros2019learning, joo2021exemplar, moon2022neuralannot, li2022cliff} manners.
For instance, SPIN~\cite{kolotouros2019learning} combines the fitting and regression process in a loop. The regressed outputs serve as better initialization for optimization.
\rv{EFT~\cite{joo2021exemplar} finetunes a pretrained SPIN network to 2D joint coordinates for each sample. But as pointed out in \cite{moon2022neuralannot}, this may lead to overfitting, especially when the input image is partially invisible.
To overcome this, NeuralAnnot~\cite{moon2022neuralannot} is trained on a mixture of 3D datasets and the target 2D in-the-wild dataset. It is optimized for entire samples.
CLIFF~\cite{li2022cliff} trains an annotator with the information from the original frames instead of the cropped ones. Thus, the CLIFF annotator produces more accurate labels, especially the global rotations.
}
Even though pseudo-labels for in-the-wild datasets are not as accurate as MoCap data, they still remarkably improve the generalization of regression-based methods thanks to their scale and diversity.

\subsection{Datasets}
Datasets involved in training and evaluation can be categorized into four groups based on data and label acquisition strategies, \ie, rendered datasets, marker/sensor-based MoCap datasets, marker-less MoCap datasets, and datasets with pseudo-3D labels.
Table~\ref{tab:datasets} summarizes some key information about these datasets.

\subsubsection{Rendered Datasets}
\paragraph{SURREAL}. \emph{Synthetic hUmans foR REAL tasks}~\cite{varol2017learning} is a large-scale synthetic human body dataset. Bodies are created with the SMPL body model and driven by 3D MoCap motions. Textures are rendered with random attributes on the background images.
The dataset contains ground truth depth maps, optical flow, surface normals, human part segmentations, and 2D/3D joint locations.

\noindent\paragraph{GTA-Human}~\cite{cai2021playing} is a large-scale 3D human dataset with diverse subjects, actions, and scenarios. The dataset is generated with the GTA-V game engine. There are 20K video sequences with SMPL annotations in this dataset.

\noindent\paragraph{AGORA}. \emph{Avatars in Geography Optimized for Regression Analysis} dataset~\cite{patel2021agora} is a recently released synthetic dataset with high realism and accurate SMPL/SMPL-X models fitted to 3D scans. Over 4,000 photorealistic textured human scans, including some children's scans, are positioned in panoramic scenes.
AGORA has become a popular benchmark for SMPL and SMPL-X estimation from monocular images.

\noindent\paragraph{THUman2.0}~\cite{yu2021function4d} contains 500 high-quality human scans with different clothing and poses captured by a 128 DSLR camera dome system. The dataset provides the 3D scan model with the corresponding texture map and fitted SMPL-X model for each scan. The person images can be generated from any viewpoint using the rendering strategy mentioned in PIFu~\cite{saito2019pifu} and PaMIR~\cite{zheng2021pamir}.

\noindent\paragraph{MultiHuman}~\cite{zheng2021deepmulticap} consists of 453 high-quality 3D human scans with raw scan meshes, texture maps, and the fitted SMPL-X models. Each scan contains 1-3 persons under occluded or interactive scenes. Images can be synthesized in the same way as THUman2.0.

\noindent\paragraph{\rv{HSPACE}}\rv{~\cite{bazavan2021hspace} relies on a corpus of 100 human scans. After fitting the scans with GHUM mesh~\cite{xu2020ghum}, the authors augment them with 16 different shape parameters. Human meshes are placed in 100 synthetic environments and are animated with over 100 motion snippets.
}

\noindent\paragraph{\rv{BEDLAM}}\rv{~\cite{black2023bedlam} is a synthetic dataset aiming to increase the scale and realism by expanding the diversity of body poses, shapes, skin tones, hair, and clothing. Moreover, the clothing is more realistic clothing as they are simulated on the moving bodies using commercial clothing physics simulation. 
}

\noindent\paragraph{\rv{SynBody}}\rv{~\cite{Yang2023SynBody} is a large-scale synthetic dataset comprising 1.2M images with corresponding 3D annotations. It covers 1,187 actions in various viewpoints, 10,000 body models, and 26,960 video clips with 2.7M SMPL/SMPLX annotations.
}

\subsubsection{Marker/Sensor-based MoCap}
\paragraph{HumanEva}~\cite{sigal2010humaneva} includes HumanEva-I and HumanEva-II. The two datasets are captured in a multi-camera MoCap system. Reflective markers are attached to subjects to record 4 subjects performing 6 actions in HumanEva-I and 2 subjects performing 1 action in HumanEva-II. 
Both datasets contain synchronized video from multiple camera views and associated 3D pose ground truth.

\noindent\paragraph{Human3.6M}~\cite{ionescu2014human3} is a benchmark dataset for 3D pose estimation. It consists of 3.6 million video frames captured against indoor backgrounds from 4 viewpoints. 5 female and 6 male subjects perform 15 actions, with reflective markers attached to their body.
The extended SMPL model annotations are generated by~\cite{kanazawa2018end, varol2017learning} after applying MoSh~\cite{loper2014mosh} to sparse marker data.
Alternatively, Moon~\etal~\cite{moon2020i2l} apply SMPLify-X~\cite{pavlakos2019expressive} to the ground truth 3D joints to get the label.

\noindent\paragraph{\rv{Total Capture}}\rv{~\cite{trumble2017total} has fully synchronized video, IMU, and Vicon labeling for about 1.7M frames. There were 4 male and 1 female subjects participated, each performing five actions, repeated 3 times.}

\noindent\paragraph{3DPW}. \emph{3D Poses in the Wild Dataset}~\cite{von2018recovering} is captured in challenging outdoor scenes. The dataset includes over 51,000 frames for 7 actors in 18 clothing styles. A hand-held smartphone camera records 1 or 2 IMU-equipped actors performing rich activities. This dataset provides accurate mesh ground truth annotations by fitting the SMPL model to the raw ground-truth markers using a similar method to~\cite{loper2014mosh}.

\noindent\paragraph{AMASS}. \emph{Archive of Motion Capture as Surface Shapes}~\cite{mahmood2019amass} is a large and varied human motion dataset that spans over 300 subjects and contains more than 40 hours of motion data for over 110K motions. It unifies 15 marker-based MoCap datasets, including CMU MoCap~\cite{cmu_mocap} and PosePrior~\cite{akhter2015pose}. The SMPL model is used to represent motions via the proposed method MoSh++. Given credit for its sufficient richness, AMASS is widely adopted to learn human motion prior and assess the rationality of predicted poses or sequences of motions.

\subsubsection{Marker-less Multi-view MoCap}
\noindent\paragraph{CMU Panoptic}~\cite{joo2015panoptic} is a large-scale multi-person dataset captured by 480 synchronized cameras in the Panoptic Studio. For each session, 3 to 8 participants are asked to play various games together to get involved in social interactions. 1.5M frames with ground truth 3D skeletons from 65 sequences are currently available.

\noindent\paragraph{MPI-INF-3DHP}~\cite{mehta2017monocular} is a single-person 3D pose dataset collected in a multi-camera green screen studio. The system is equipped with 14 cameras and records 8 subjects in total. Each subject features 2 sets of clothing and performs 8 activities.
Ground truth 3D pose annotations are available, but some noise exists. The authors further propose MuCo-3DHP~\cite{mehta2018single} as data augmentation. It is built on the person masks in MPI-INF-3DHP. 1 to 4 subjects are pasted to real-world background images, resulting in 200K images that cover a range of inter-person overlap and activity scenarios.

\noindent\paragraph{MuPoTs-3D}. \emph{Multiperson Pose Test Set in 3D}~\cite{mehta2018single} is a multi-person dataset for evaluation. It consists of more than 8000 frames covering 5 indoor and 15 outdoor settings. The ground-truth 3D poses are captured in a multi-view marker-less motion capture system.

\noindent\paragraph{MannequinChallenge}~\cite{li2019learning} contains videos in which multiple people freeze in the pose, and the camera moves around to film the static scenes.
The dataset originally provides estimated camera poses and dense depth. Leroy~\etal~\cite{leroy2020smply} further extend the annotations with 3D keypoints locations and visibility information using a SMPL-based approach.

\noindent\paragraph{3DOH50K}. \emph{3D Occlusion Human 50K} dataset~\cite{zhang2020object} is captured in indoor scenes with 6 viewpoints. It contains more than 51,600 images, most of which are human activities in occlusion scenarios. The authors adapt SMPLify-X in a multi-view strategy to get the SMPL mesh ground truth.

\noindent\paragraph{Mirrored-Human}~\cite{fang2021reconstructing} consists of videos from the Internet, in which we can see a person and the person's image in a mirror. The mirror reflection provides an additional view to resolve the depth ambiguity. The dataset provides 2D keypoints and pseudo-ground truth SMPL annotations generated by an optimization-based framework.

\noindent\paragraph{MTC}. \emph{Monocular Total Capture} dataset~\cite{xiang2019monocular} is captured by the Panoptic Studio~\cite{joo2015panoptic} with 31 HD cameras in a multi-view setup. The dataset contains about 834K body images and 111K hand images, representing a wide range of motions in the body and hand of multiple subjects.

\noindent\paragraph{EHF}. \emph{Expressive Hands and Faces} dataset~\cite{pavlakos2019expressive} contains 100 samples for evaluation. Following \cite{romero2017embodied}, the SMPL-X model is aligned to original 4D scans. With special attention paid to hand poses and facial expressions, mesh annotations of the selected samples are of good alignment quality.

\noindent\paragraph{HUMBI}~\cite{yu2020humbi} is a large multi-view dataset for human body expressions with natural clothing. 107 synchronized HD cameras are employed to capture 772 distinctive subjects. The subjects at the capture stage are asked to perform a series of gaze, face, hand, and body expression tasks. Each frame contains up to 4 representations: multi-view images, 3D keypoints, 3D mesh, and appearance maps. Basel Face Model~\cite{paysan20093d}, MANO~\cite{romero2017embodied}, and SMPL~\cite{loper2015smpl} are adopted for face, hands, and body reconstruction, respectively.

\noindent\paragraph{ZJU-MoCap}~\cite{peng2021neural} consists of 9 dynamic human sequences captured by 21 synchronized cameras in a multi-view setup. The sequences have a length between 60 to 300 frames, in which actors do complex movements like twirling and kicking. The SMPL-X annotations are also available after iteratively optimizing the human model to align with the multi-view observations.

\rv{
\subsubsection{Datasets with Human-Scene Interactions}
There are several datasets for investigating the task of human/hand mesh recovery with human-object interactions.
PROX~\cite{hassan2019resolving} uses a single Kinect camera to capture 20 subjects interacting with the indoor scenes. The dataset provides 12 indoor scene meshes and 100K RGB-D frames with pseudo SMPL-X labels.
BEHAVE~\cite{bhatnagar2022behave} captures dynamic human-object interactions using 4 Kinects in natural environments. The dataset contains multi-view RGB-D sequences and corresponding human models, objects, and contact annotations. It has 10.7k frames for training and 4.5k frames for testing.
GRAB~\cite{taheri2020grab} uses a marker-based capture system to capture 10 subjects interacting with 51 everyday objects. The SMPL-X model is fitted to Mocap markers to present body pose, shape, facial expression, and hand gestures. However, the dataset does not have corresponding RGB(-D) frames.
RICH~\cite{huang2022capturing} contains multiview outdoor/indoor high-resolution video sequences, ground-truth 3D human bodies, 3D body scans, and high-resolution 3D scene scans.
SLOPER4D~\cite{dai2023sloper4d} is a scene-aware dataset collected in urban environments, consisting of 15 sequences, more than 100K LiDAR frames, 300k video frames, and 500K IMU-based motion frames. 
}

\subsubsection{Datasets with Pseudo 3D Labels}
2D pose datasets are known for their richness and diversity in subjects, poses, and scenes, but lack 3D pose or mesh annotations. Researchers have explored algorithms to generate pseudo-ground truth in an automatic or semi-automatic manner. 
\textbf{LSP}~\cite{johnson2010clustered}, \textbf{LSP-Extended}~\cite{johnson2011learning}, \textbf{MSCOCO}~\cite{lin2014microsoft}, \textbf{MPII}~\cite{andriluka20142d}, \textbf{PoseTrack}~\cite{andriluka2018posetrack}, \textbf{OCHuman}~\cite{zhang2019pose2seg}
are in-the-wild 2D human pose estimation datasets. Their labels are fitted in an optimization process~\cite{lassner2017unite} or with the help of regression networks~\cite{kolotouros2019learning,joo2021exemplar}.

\noindent\paragraph{SSP-3D}~\cite{sengupta2020synthetic} is collected from the Sports-1M video dataset~\cite{karpathy2014large}. SSP-3D comprises 311 in-the-wild images of 62 tightly-clothed sportspersons with a diverse range of body shapes and corresponding pseudo-ground truth SMPL shape and pose labels.

\noindent\paragraph{MTP}. \emph{Mimic The Pose} dataset~\cite{muller2021self} contains 3,731 images corresponding to 1,653 SMPL-X meshes. 3D meshes exhibit self-contact, and images are collected after asking participants to mimic the poses and contacts. Since the presented pose, shape, and gender are not aligned perfectly, the authors further adapt SMPLify-X~\cite{pavlakos2019expressive} to refine the original meshes.

\noindent\paragraph{\rv{UBody.}} \rv{\emph{Upper-Body} dataset~\cite{lin2023one} mainly focuses on representing upper bodies. It contains a series of close-up shots of humans with rich hand gestures and facial expressions in 15 real-life scenarios. The dataset has 2D annotations and high-quality 3D pseudo-GT SMPL-X fits.}

\noindent\paragraph{\rv{Motion-X}}~\cite{lin2023motionx} \rv{is a large-scale motion dataset comprising 15.6M 3D whole-body pose annotations in the form of SMPL-X. It consists of 81.1K motion sequences of in-the-wild scenes and provides corresponding semantic labels and pose descriptions.}

\section{Evaluation}
\label{sec:evaluation}
In this section, we discuss the evaluation metrics and the benchmark results from multiple perspectives.

\subsection{Metrics}
\paragraph{MPJPE}. \emph{Mean Per Joint Position Error} measures the average Euclidean distance between predicted 3D joints and ground truth after root matching. It is defined in the local space. Recently, SPEC~\cite{kocabas2021spec} introduces W-MPJPE that computes 3D joints error in the world coordinates. The authors believe it can better reflect performance in real-world applications.

\noindent\paragraph{PA-MPJPE}. \emph{Procrustes-aligned MPJPE} denotes MPJPE after rigid alignment of the predicted pose and ground truth. Procrustes Analysis removes the effects of translation, rotation, and scale. Thus, PA-MPJPE concerns the reconstructed 3D mesh/pose itself. It is also referred to as the reconstruction error.

\noindent\paragraph{PVE/V2V}. \emph{Mean Per-vertex Error} or \emph{Vertex-to-Vertex} is defined as the average point-to-point Euclidean distance between predicted mesh vertices and ground truth mesh vertices. Similar to MPJPEC, W-PVE, a variant of PVE, is proposed~\cite{kocabas2021spec} to calculate in the world space.

\noindent\paragraph{MPJAE}. \emph{Mean Per Joint Angle Error} represents the orientation deviation between predicted 3D joints and ground truth, which is measured in $SO(3)$ using the geodesic distance.

\noindent\paragraph{PA-MPJAE}. \emph{Procrustes-aligned MPJAE} is calculated according to MPJAE after executing Procrustes Analysis to align predicted poses with ground truth.

\subsection{Benchmark Leaderboards}
\label{sec:benchmark}
The quantitative comparison of 3D body mesh recovery on Human3.6M~\cite{ionescu2014human3} and 3DPW~\cite{von2018recovering} are illustrated in Table~\ref{tab:h36m_3dpw_sota}. With the researchers' persistent efforts, the performance has been improving each year. However, the deployment and evaluation standards for comparison are not fully consistent. Different combinations of backbones, output types, pseudo labels, datasets, training strategies, and evaluation protocols would lead to a fluctuation in values.
SPIN~\cite{kolotouros2019learning} establishes an evaluation protocol that is widely adopted by the follow-ups in the table.
In 3DPW, most approaches follow \emph{Protocol 2} and use the test set for evaluation without any fine-tuning on the training set. But the strategy is different in \cite{zanfir2020weakly, sengupta2021probabilistic, lin2021end, lin2021mesh, zanfir2021neural, kocabas2021spec, luo20203d, lee2021uncertainty} in which 3DPW train set is used during training.
\cite{zanfir2021neural, zanfir2021thundr} use the GHUM model~\cite{xu2020ghum} to represent the pose and shape, while others adopt the SMPL model~\cite{loper2015smpl}.
In general, ResNet-50~\cite{he2016deep} serves as a generic convolutional backbone to extract features from images, except that \cite{lin2021end, lin2021mesh} use HRNet~\cite{wang2020deep} and multi-stage pipelines~\cite{pavlakos2018learning, omran2018neural, zhang2019danet, zhang2020object, doersch2019sim2real} have multiple convolutional modules.
For the methods that yield non-parametric outputs, metrics will degrade in general after the outputs are converted to parameters with an additional parameter regression module~\cite{kolotouros2019convolutional, moon2020i2l}.

There are much fewer algorithms that deal with full-body mesh recovery with face and hands, compared to body-only mesh recovery.
Table~\ref{tab:agora_sota} and Table~\ref{tab:ehf_sota} show the performances of the full-body recovery task on the AGORA dataset~\cite{patel2021agora} and the EHF dataset~\cite{pavlakos2019expressive}, respectively.
Results on body-only, face, and hands are also included in these tables for comprehensive evaluations.
Since AGORA does not provide the ground-truth labels for its test set, the performances are calculated after uploading the results to the official evaluation platform~\cite{patel2021agora}.
Comparing the results of \textbf{FB} and \textbf{B} in Table~\ref{tab:agora_sota} and Table~\ref{tab:ehf_sota}, we can observe that this task is very challenging as the reconstruction error becomes much higher when taking face and hands into consideration.

\fi
\section{Conclusion and Future Directions}
\label{sec:conclusion}
In this survey, we provide a thorough overview of 3D human mesh recovery methods in the past decade. The categorization is based on design paradigm, reconstruction granularity, and application scenarios. We also give special considerations for physical plausibility, including camera models, contact constraints, and human priors. In the experiment section, we introduce relevant datasets, evaluation metrics and provide performance comparison.  
Next, we highlight a few promising future directions, hoping to promote advances in this field.

\paragraph{Under Heavy Occlusions}.
In real-world scenarios, occlusions are ubiquitous. People often appear partially or heavily occluded due to self-overlapping, close-range interaction with other people, or occlusion of scene objects.
Even though the occlusion has been extensively studied for years~\cite{zhang2019danet, zhang2020object, rockwell2020full, kocabas2021pare}, robustness and stability are still need to be improved.
Besides, the visual evidence may be insufficient to identify a 3D reconstruction uniquely, recover several plausible reconstructions~\cite{biggs20203d} or a pose distribution~\cite{kolotouros2021probabilistic, sengupta2021hierarchical} for one input is worthwhile.

\paragraph{Stable Reconstruction from Videos}. Motion jitters, \ie, irregular movement and variation across frames, remain a severe issue in existing regression-based temporal-based methods~\cite{kocabas2020vibe,luo20203d,choi2021beyond}. 
The visual performance is largely influenced by motion jitters. Jitters are slight when much of the body is observable, while severe jitters occur in those frames with heavy occlusion or in a complex context. To improve temporal smoothness, we need to deal with long-term motion jitters. There is a trend to perform pose refinement after primary estimation using low-pass filters or learning-based refinement networks~\cite{zeng2021smoothnet}. 

\paragraph{Reconstruction with Scene Constraints}. Standard methods perform 3D human pose estimation without explicitly considering the scene. This may lead to inter-penetration with the 3D scene.
Most methods ignore the scene constraint during estimation. In the methods that aim to reconstruct physically consistent results, scenes are typically assumed as flat floors~\cite{rempe2020contact, rempe2021humor, xie2021physics} for simplicity. \cite{hassan2019resolving, zhang2020place, zhang2020generating, liu20204d} are among the first to go beyond flat floors and resolve human pose and shape from static 3D scenes. Further work may take scene mesh into consideration to better capture the motion of humans interacting with a real static 3D scene.

\paragraph{Beyond Fully Supervised Learning}. Building 3D human mesh datasets is time-consuming and of high cost. A MoCap system needs to be set up beforehand. After capturing, the cleaning and annotation process of raw 3D data is highly demanding. Besides, 3D datasets lack diversity in human motion and background, but 2D datasets are far more substantial. In light of this, it is promising to make use of the abundant unlabeled data to train a network in an unsupervised fashion.
Recent unsupervised 3D pose estimation~\cite{chen2019unsupervised, rhodin2018unsupervised} has achieved exciting performance. Compared to this, unsupervised~\cite{tripathi2020posenet3d, yu2021skeleton2mesh} or self-supervised~\cite{xu20203d, rockwell2020full, kundu2020appearance} human mesh recovery is much more difficult due to richer reconstruction information.

\paragraph{Grouped Person Reconstruction}.
In public scenes, people often walk, talk, or work together in groups as family members, teammates, \etc.
An interesting future direction is reconstructing a group of people over space and time, which reveals the relationships and activities in the target group.
Moreover, when considering person matching across different cameras or long-range temporal sequences, the relationship of individuals within a group provides a more stable context that can be exploited to handle occlusions or detection failures.
This task can also be combined with person tracking~\cite{rajasegaran2021tracking} and re-identification~\cite{zheng2009associating,lisanti2017group} for more robust reconstruction in crowded scenarios.

\rv{
\paragraph{Whole-body Human Mesh Recovery}. 
There is a trend to utilize a unified framework to regress the body, hands, and face parameters of expressive human models~\cite{pavlakos2019expressive,xu2020ghum}. 
Compared with body-only mesh recovery, there are much fewer methods to deal with whole-body mesh recovery~\cite{choutas2020monocular,rong2020frankmocap,feng2021collaborative,zhang2023pymafx,li2023hybrikx}.
One major challenge is that the whole-body datasets are rather scarce for training. Separate body/hand/face-only datasets are typically used to compensate for the incompleteness of whole-body data. This brings challenges to the consistent recovery of body poses and hand gestures.
Moreover, the occlusions, motion blur, depth ambiguity, and interaction of the hand regions also impose great challenges to the monocular whole-body mesh recovery with plausible hand poses.
}

\paragraph{Detailed Shape Reconstruction with Clothing}.
Parametric models like SMPL and SMPL-X can only represent minimally clothed humans.
The research community needs to exploit other representations with more flexibility to go beyond the representation power of parametric models.
In existing work, meshes~\cite{xu2018monoperfcap,ma2020cape,zhu2019detailed}, point clouds~\cite{ma2021scale,ma2021power,lin2022learning,zhang2023closet}, and implicit fields~\cite{saito2019pifu,li2020robust,alldieck2022photorealistic,chen2021snarf} have been used to model the detailed deformation of clothing.
Though these methods can produce reasonable results, their reconstructed surfaces tend to be over-smoothed and not robust to novel poses.
These issues can be alleviated by incorporating different types of representations~\cite{bhatnagar2020combining,zheng2021pamir,shao2022dbfield,Feng2022scarf,moon20223d} to leverage the modeling power of different representations.

\ifCLASSOPTIONcompsoc
  \section*{Acknowledgments}
\else
  \section*{Acknowledgment}
\fi
This work was supported in part by the National Key R$\&$D Program of China under Grant 2022ZD0160900, the National Natural Science Foundation of China under Grants 62076119, 61921006, and 62125107, in part by the Fundamental Research Funds for the Central Universities under Grant 020214380091, and in part by the Collaborative Innovation Center of Novel Software Technology and Industrialization.


\ifCLASSOPTIONcaptionsoff
  \newpage
\fi

\bibliographystyle{IEEEtran}
\bibliography{IEEEabrv,bibtex}

\begin{thebibliography}{100}
\providecommand{\url}[1]{#1}
\csname url@samestyle\endcsname
\providecommand{\newblock}{\relax}
\providecommand{\bibinfo}[2]{#2}
\providecommand{\BIBentrySTDinterwordspacing}{\spaceskip=0pt\relax}
\providecommand{\BIBentryALTinterwordstretchfactor}{4}
\providecommand{\BIBentryALTinterwordspacing}{\spaceskip=\fontdimen2\font plus
\BIBentryALTinterwordstretchfactor\fontdimen3\font minus
  \fontdimen4\font\relax}
\providecommand{\BIBforeignlanguage}[2]{{%
\expandafter\ifx\csname l@#1\endcsname\relax
\typeout{** WARNING: IEEEtran.bst: No hyphenation pattern has been}%
\typeout{** loaded for the language `#1'. Using the pattern for}%
\typeout{** the default language instead.}%
\else
\language=\csname l@#1\endcsname
\fi
#2}}
\providecommand{\BIBdecl}{\relax}
\BIBdecl

\bibitem{cao2019openpose}
Z.~Cao, G.~Hidalgo, T.~Simon, S.-E. Wei, and Y.~Sheikh, ``{OpenPose}: Realtime
  multi-person {2D} pose estimation using part affinity fields,'' \emph{TPAMI},
  vol.~43, no.~1, pp. 172--186, 2019.

\bibitem{fang2017rmpe}
H.-S. Fang, S.~Xie, Y.-W. Tai, and C.~Lu, ``{RMPE}: Regional multi-person pose
  estimation,'' in \emph{ICCV}, 2017, pp. 2334--2343.

\bibitem{kreiss2021openpifpaf}
S.~Kreiss, L.~Bertoni, and A.~Alahi, ``{OpenPifPaf}: Composite fields for
  semantic keypoint detection and spatio-temporal association,'' \emph{TITS},
  vol.~23, no.~8, pp. 13\,498--13\,511, 2021.

\bibitem{chen2018semantic}
Q.~Chen, T.~Ge, Y.~Xu, Z.~Zhang, X.~Yang, and K.~Gai, ``Semantic human
  matting,'' in \emph{ACM MM}, 2018, pp. 618--626.

\bibitem{zhao2018understanding}
J.~Zhao, J.~Li, Y.~Cheng, T.~Sim, S.~Yan, and J.~Feng, ``Understanding humans
  in crowded scenes: Deep nested adversarial learning and a new benchmark for
  multi-human parsing,'' in \emph{ACM MM}, 2018, pp. 792--800.

\bibitem{grauman2003inferring}
K.~Grauman, G.~Shakhnarovich, and T.~Darrell, ``Inferring {3D} structure with a
  statistical image-based shape model,'' in \emph{ICCV}, 2003, pp. 641--648.

\bibitem{agarwal2005recovering}
A.~Agarwal and B.~Triggs, ``Recovering {3D} human pose from monocular images,''
  \emph{TPAMI}, vol.~28, no.~1, pp. 44--58, 2005.

\bibitem{martinez_2017_3dbaseline}
J.~Martinez, R.~Hossain, J.~Romero, and J.~J. Little, ``A simple yet effective
  baseline for {3D} human pose estimation,'' in \emph{ICCV}, 2017, pp.
  2659--2668.

\bibitem{pavlakos2017coarse}
G.~Pavlakos, X.~Zhou, K.~G. Derpanis, and K.~Daniilidis, ``Coarse-to-fine
  volumetric prediction for single-image {3D} human pose,'' in \emph{CVPR},
  2017, pp. 1263--1272.

\bibitem{sun2018integral}
X.~Sun, B.~Xiao, F.~Wei, S.~Liang, and Y.~Wei, ``Integral human pose
  regression,'' in \emph{ECCV}, 2018, pp. 536--553.

\bibitem{mehta2020xnect}
D.~Mehta, O.~Sotnychenko, F.~Mueller, W.~Xu, M.~Elgharib, P.~Fua, H.-P. Seidel,
  H.~Rhodin, G.~Pons-Moll, and C.~Theobalt, ``{XNect}: Real-time multi-person
  {3D} motion capture with a single {RGB} camera,'' \emph{TOG}, vol.~39, no.~4,
  pp. 82--1, 2020.

\bibitem{Weinzaepfel2020_dope}
P.~Weinzaepfel, R.~Br{\'e}gier, H.~Combaluzier, V.~Leroy, and G.~Rogez,
  ``{DOPE}: Distillation of part experts for whole-body {3D} pose estimation in
  the wild,'' in \emph{ECCV}, 2020, pp. 380--397.

\bibitem{bogo2016keep}
F.~Bogo, A.~Kanazawa, C.~Lassner, P.~Gehler, J.~Romero, and M.~J. Black, ``Keep
  it {SMPL}: Automatic estimation of {3D} human pose and shape from a single
  image,'' in \emph{ECCV}.\hskip 1em plus 0.5em minus 0.4em\relax Springer,
  2016, pp. 561--578.

\bibitem{huang2017towards}
Y.~Huang, F.~Bogo, C.~Lassner, A.~Kanazawa, P.~V. Gehler, J.~Romero, I.~Akhter,
  and M.~J. Black, ``Towards accurate marker-less human shape and pose
  estimation over time,'' in \emph{3DV}.\hskip 1em plus 0.5em minus 0.4em\relax
  IEEE, 2017, pp. 421--430.

\bibitem{zanfir2018monocular}
A.~Zanfir, E.~Marinoiu, and C.~Sminchisescu, ``Monocular {3D} pose and shape
  estimation of multiple people in natural scenes - the importance of multiple
  scene constraints,'' in \emph{CVPR}, 2018, pp. 2148--2157.

\bibitem{kanazawa2018end}
A.~Kanazawa, M.~J. Black, D.~W. Jacobs, and J.~Malik, ``End-to-end recovery of
  human shape and pose,'' in \emph{CVPR}, 2018, pp. 7122--7131.

\bibitem{pavlakos2018learning}
G.~Pavlakos, L.~Zhu, X.~Zhou, and K.~Daniilidis, ``Learning to estimate {3D}
  human pose and shape from a single color image,'' in \emph{CVPR}, 2018, pp.
  459--468.

\bibitem{omran2018neural}
M.~Omran, C.~Lassner, G.~Pons-Moll, P.~Gehler, and B.~Schiele, ``{Neural Body
  Fitting}: Unifying deep learning and model-based human pose and shape
  estimation,'' in \emph{3DV}.\hskip 1em plus 0.5em minus 0.4em\relax IEEE,
  2018, pp. 484--494.

\bibitem{zhang2021pymaf}
H.~Zhang, Y.~Tian, X.~Zhou, W.~Ouyang, Y.~Liu, L.~Wang, and Z.~Sun, ``{PyMAF}:
  {3D} human pose and shape regression with pyramidal mesh alignment feedback
  loop,'' in \emph{ICCV}, 2021.

\bibitem{kocabas2021pare}
M.~Kocabas, C.-H.~P. Huang, O.~Hilliges, and M.~J. Black, ``{PARE}: Part
  attention regressor for {3D} human body estimation,'' in \emph{ICCV}, 2021,
  pp. 11\,127--11\,137.

\bibitem{joo2021exemplar}
H.~Joo, N.~Neverova, and A.~Vedaldi, ``Exemplar fine-tuning for {3D} human pose
  fitting towards in-the-wild {3D} human pose estimation,'' in \emph{3DV},
  2021, pp. 42--52.

\bibitem{pavlakos2019expressive}
G.~Pavlakos, V.~Choutas, N.~Ghorbani, T.~Bolkart, A.~A. Osman, D.~Tzionas, and
  M.~J. Black, ``Expressive body capture: {3D} hands, face, and body from a
  single image,'' in \emph{CVPR}, 2019, pp. 10\,975--10\,985.

\bibitem{choutas2020monocular}
V.~Choutas, G.~Pavlakos, T.~Bolkart, D.~Tzionas, and M.~J. Black, ``Monocular
  expressive body regression through body-driven attention,'' in
  \emph{ECCV}.\hskip 1em plus 0.5em minus 0.4em\relax Springer, 2020, pp.
  20--40.

\bibitem{feng2021collaborative}
Y.~Feng, V.~Choutas, T.~Bolkart, D.~Tzionas, and M.~J. Black, ``Collaborative
  regression of expressive bodies using moderation,'' in \emph{3DV}, 2021.

\bibitem{moon2022accurate}
G.~Moon, H.~Choi, and K.~M. Lee, ``Accurate {3D} hand pose estimation for
  whole-body {3D} human mesh estimation,'' in \emph{CVPRW}, 2022, pp.
  2308--2317.

\bibitem{zhang2021lightweight}
Y.~Zhang, Z.~Li, L.~An, M.~Li, T.~Yu, and Y.~Liu, ``Lightweight multi-person
  total motion capture using sparse multi-view cameras,'' in \emph{ICCV}, 2021,
  pp. 5560--5569.

\bibitem{loper2015smpl}
M.~Loper, N.~Mahmood, J.~Romero, G.~Pons-Moll, and M.~J. Black, ``{SMPL}: A
  skinned multi-person linear model,'' \emph{TOG}, vol.~34, no.~6, pp. 1--16,
  2015.

\bibitem{yu2018doublefusion}
T.~Yu, Z.~Zheng, K.~Guo, J.~Zhao, Q.~Dai, H.~Li, G.~Pons-Moll, and Y.~Liu,
  ``{DoubleFusion}: Real-time capture of human performances with inner body
  shapes from a single depth sensor,'' in \emph{CVPR}, 2018, pp. 7287--7296.

\bibitem{zheng2021pamir}
Z.~Zheng, T.~Yu, Y.~Liu, and Q.~Dai, ``{PaMIR}: Parametric model-conditioned
  implicit representation for image-based human reconstruction,'' \emph{TPAMI},
  2021.

\bibitem{zheng2021deepmulticap}
Y.~Zheng, R.~Shao, Y.~Zhang, T.~Yu, Z.~Zheng, Q.~Dai, and Y.~Liu,
  ``{DeepMultiCap}: Performance capture of multiple characters using sparse
  multiview cameras,'' in \emph{ICCV}, 2021, pp. 6239--6249.

\bibitem{li2021image}
K.~Li, H.~Wen, Q.~Feng, Y.~Zhang, X.~Li, J.~Huang, C.~Yuan, Y.-K. Lai, and
  Y.~Liu, ``Image-guided human reconstruction via multi-scale graph
  transformation networks,'' \emph{TIP}, vol.~30, pp. 5239--5251, 2021.

\bibitem{feng2022fof}
Q.~Feng, Y.~Liu, Y.-K. Lai, J.~Yang, and K.~Li, ``{FOF}: Learning fourier
  occupancy field for monocular real-time human reconstruction,'' in
  \emph{NeurIPS}, 2022.

\bibitem{xiu2022icon}
Y.~Xiu, J.~Yang, D.~Tzionas, and M.~J. Black, ``{ICON}: implicit clothed humans
  obtained from normals,'' in \emph{CVPR}, 2022, pp. 13\,286--13\,296.

\bibitem{xiu2023econ}
Y.~Xiu, J.~Yang, X.~Cao, D.~Tzionas, and M.~J. Black, ``{ECON}: Explicit
  clothed humans optimized via normal integration,'' in \emph{CVPR}, 2023, pp.
  512--523.

\bibitem{peng2021neural}
S.~Peng, Y.~Zhang, Y.~Xu, Q.~Wang, Q.~Shuai, H.~Bao, and X.~Zhou, ``{Neural
  Body}: Implicit neural representations with structured latent codes for novel
  view synthesis of dynamic humans,'' in \emph{CVPR}, 2021, pp. 9054--9063.

\bibitem{hu2022hvtr}
T.~Hu, T.~Yu, Z.~Zheng, H.~Zhang, Y.~Liu, and M.~Zwicker, ``{HVTR}: Hybrid
  volumetric-textural rendering for human avatars,'' in \emph{3DV}.\hskip 1em
  plus 0.5em minus 0.4em\relax IEEE, 2022, pp. 197--208.

\bibitem{huang2020arch}
Z.~Huang, Y.~Xu, C.~Lassner, H.~Li, and T.~Tung, ``{ARCH}: Animatable
  reconstruction of clothed humans,'' in \emph{CVPR}, 2020, pp. 3093--3102.

\bibitem{ma2021power}
Q.~Ma, J.~Yang, S.~Tang, and M.~J. Black, ``The power of points for modeling
  humans in clothing,'' in \emph{ICCV}, 2021, pp. 10\,974--10\,984.

\bibitem{zheng2022structured}
Z.~Zheng, H.~Huang, T.~Yu, H.~Zhang, Y.~Guo, and Y.~Liu, ``Structured local
  radiance fields for human avatar modeling,'' in \emph{CVPR}, 2022, pp.
  15\,893--15\,903.

\bibitem{zheng2023avatarrex}
Z.~Zheng, X.~Zhao, H.~Zhang, B.~Liu, and Y.~Liu, ``{AvatarReX}: Real-time
  expressive full-body avatars,'' \emph{ACM TOG}, vol.~42, no.~4, 2023.

\bibitem{anguelov2005scape}
D.~Anguelov, P.~Srinivasan, D.~Koller, S.~Thrun, J.~Rodgers, and J.~Davis,
  ``{SCAPE:} shape completion and animation of people,'' \emph{TOG}, vol.~24,
  pp. 408--416, 2005.

\bibitem{lassner2017unite}
C.~Lassner, J.~Romero, M.~Kiefel, F.~Bogo, M.~J. Black, and P.~V. Gehler,
  ``Unite the people: Closing the loop between {3D} and {2D} human
  representations,'' in \emph{CVPR}, 2017, pp. 6050--6059.

\bibitem{tung2017self}
H.-Y.~F. Tung, H.-W. Tung, E.~Yumer, and K.~Fragkiadaki, ``Self-supervised
  learning of motion capture,'' \emph{NeurIPS}, pp. 5236--5246, 2017.

\bibitem{kolotouros2019learning}
N.~Kolotouros, G.~Pavlakos, M.~J. Black, and K.~Daniilidis, ``Learning to
  reconstruct {3D} human pose and shape via model-fitting in the loop,'' in
  \emph{ICCV}, 2019, pp. 2252--2261.

\bibitem{chen2021towards}
L.~Chen, S.~Peng, and X.~Zhou, ``Towards efficient and photorealistic {3D}
  human reconstruction: A brief survey,'' \emph{Visual Informatics}, vol.~5,
  no.~4, pp. 11--19, 2021.

\bibitem{tewari2020state}
A.~Tewari, O.~Fried, J.~Thies, V.~Sitzmann, S.~Lombardi, K.~Sunkavalli,
  R.~Martin-Brualla, T.~Simon, J.~Saragih, M.~Nie{\ss}ner \emph{et~al.},
  ``State of the art on neural rendering,'' in \emph{CGF}, vol.~39.\hskip 1em
  plus 0.5em minus 0.4em\relax Wiley Online Library, 2020, pp. 701--727.

\bibitem{chen2020monocular}
Y.~Chen, Y.~Tian, and M.~He, ``Monocular human pose estimation: A survey of
  deep learning-based methods,'' \emph{CVIU}, vol. 192, p. 102897, 2020.

\bibitem{zheng2020deep}
C.~Zheng, W.~Wu, T.~Yang, S.~Zhu, C.~Chen, R.~Liu, J.~Shen, N.~Kehtarnavaz, and
  M.~Shah, ``Deep learning-based human pose estimation: A survey,'' \emph{arXiv
  preprint arXiv:2012.13392}, 2020.

\bibitem{liu2021recent}
W.~Liu, Q.~Bao, Y.~Sun, and T.~Mei, ``Recent advances of monocular {2D} and
  {3D} human pose estimation: A deep learning perspective,'' \emph{ACM
  Computing Surveys}, vol.~55, no.~4, pp. 1--41, 2022.

\bibitem{lee1985determination}
H.-J. Lee and Z.~Chen, ``Determination of {3D} human body postures from a
  single view,'' \emph{Computer Vision, Graphics, and Image Processing},
  vol.~30, no.~2, pp. 148--168, 1985.

\bibitem{nevatia1977description}
R.~Nevatia and T.~O. Binford, ``Description and recognition of curved
  objects,'' \emph{Artificial intelligence}, vol.~8, no.~1, pp. 77--98, 1977.

\bibitem{ju1996cardboard}
S.~X. Ju, M.~J. Black, and Y.~Yacoob, ``Cardboard people: A parameterized model
  of articulated image motion,'' in \emph{FG}.\hskip 1em plus 0.5em minus
  0.4em\relax IEEE, 1996, pp. 38--44.

\bibitem{blanz1999morphable}
V.~Blanz and T.~Vetter, ``A morphable model for the synthesis of {3D} faces,''
  in \emph{SIGGRAPH}, 1999, pp. 187--194.

\bibitem{marr1978representation}
D.~Marr and H.~K. Nishihara, ``Representation and recognition of the spatial
  organization of three-dimensional shapes,'' \emph{Proceedings of the Royal
  Society of London. Series B. Biological Sciences}, vol. 200, no. 1140, pp.
  269--294, 1978.

\bibitem{rohr1994towards}
K.~Rohr, ``Towards model-based recognition of human movements in image
  sequences,'' \emph{CVGIP: Image understanding}, vol.~59, no.~1, pp. 94--115,
  1994.

\bibitem{wachter1999tracking}
S.~Wachter and H.-H. Nagel, ``Tracking of persons in monocular image
  sequences,'' \emph{CVIU}, vol.~74, no.~3, pp. 174--192, 1999.

\bibitem{sidenbladh2000stochastic}
H.~Sidenbladh, M.~J. Black, and D.~J. Fleet, ``Stochastic tracking of {3D}
  human figures using {2D} image motion,'' in \emph{ECCV}.\hskip 1em plus 0.5em
  minus 0.4em\relax Springer, 2000, pp. 702--718.

\bibitem{sigal2010humaneva}
L.~Sigal, A.~O. Balan, and M.~J. Black, ``{HumanEva}: Synchronized video and
  motion capture dataset and baseline algorithm for evaluation of articulated
  human motion,'' \emph{IJCV}, vol.~87, no. 1-2, p.~4, 2010.

\bibitem{wang2020monocular}
M.~Wang, F.~Qiu, W.~Liu, C.~Qian, X.~Zhou, and L.~Ma, ``Monocular human pose
  and shape reconstruction using part differentiable rendering,'' in
  \emph{CGF}, vol.~39.\hskip 1em plus 0.5em minus 0.4em\relax Wiley Online
  Library, 2020, pp. 351--362.

\bibitem{pentland1991recovery}
A.~Pentland and B.~Horowitz, ``Recovery of nonrigid motion and structure,''
  \emph{TPAMI}, vol.~13, no.~07, pp. 730--742, 1991.

\bibitem{metaxas1993shape}
D.~Metaxas and D.~Terzopoulos, ``Shape and nonrigid motion estimation through
  physics-based synthesis,'' \emph{TPAMI}, vol.~15, no.~6, pp. 580--591, 1993.

\bibitem{gavrila1996vision}
D.~M. Gavrila, \emph{Vision-based {3-D} tracking of humans in action}.\hskip
  1em plus 0.5em minus 0.4em\relax University of Maryland, College Park, 1996.

\bibitem{sminchisescu2003estimating}
C.~Sminchisescu and B.~Triggs, ``Estimating articulated human motion with
  covariance scaled sampling,'' \emph{International Journal of Robotics
  Research}, vol.~22, no.~6, pp. 371--391, 2003.

\bibitem{plankers2001tracking}
R.~Pl{\"a}nkers and P.~Fua, ``Tracking and modeling people in video
  sequences,'' \emph{CVIU}, vol.~81, no.~3, pp. 285--302, 2001.

\bibitem{kakadiaris2000model}
L.~Kakadiaris and D.~Metaxas, ``Model-based estimation of {3D} human motion,''
  \emph{TPAMI}, vol.~22, no.~12, pp. 1453--1459, 2000.

\bibitem{pons-moll_rosenhahn_2011}
G.~Pons-Moll and B.~Rosenhahn, ``Model-based pose estimation,'' \emph{Visual
  Analysis of Humans}, pp. 139--170, 2011.

\bibitem{allen2003space}
B.~Allen, B.~Curless, and Z.~Popovi{\'c}, ``The space of human body shapes:
  reconstruction and parameterization from range scans,'' \emph{TOG}, vol.~22,
  no.~3, pp. 587--594, 2003.

\bibitem{hasler2009statistical}
N.~Hasler, C.~Stoll, M.~Sunkel, B.~Rosenhahn, and H.-P. Seidel, ``A statistical
  model of human pose and body shape,'' in \emph{CGF}, vol.~28.\hskip 1em plus
  0.5em minus 0.4em\relax Wiley Online Library, 2009, pp. 337--346.

\bibitem{chen2013tensor}
Y.~Chen, Z.~Liu, and Z.~Zhang, ``Tensor-based human body modeling,'' in
  \emph{CVPR}, 2013, pp. 105--112.

\bibitem{freifeld2012lie}
O.~Freifeld and M.~J. Black, ``Lie bodies: A manifold representation of {3D}
  human shape,'' in \emph{ECCV}.\hskip 1em plus 0.5em minus 0.4em\relax
  Springer, 2012, pp. 1--14.

\bibitem{hirshberg2012coregistration}
D.~A. Hirshberg, M.~Loper, E.~Rachlin, and M.~J. Black, ``Coregistration:
  Simultaneous alignment and modeling of articulated {3D} shape,'' in
  \emph{ECCV}.\hskip 1em plus 0.5em minus 0.4em\relax Springer, 2012, pp.
  242--255.

\bibitem{pons2015dyna}
G.~Pons-Moll, J.~Romero, N.~Mahmood, and M.~J. Black, ``{Dyna}: A model of
  dynamic human shape in motion,'' \emph{TOG}, vol.~34, no.~4, pp. 1--14, 2015.

\bibitem{allen2006learning}
B.~Allen, B.~Curless, Z.~Popovi{\'c}, and A.~Hertzmann, ``Learning a correlated
  model of identity and pose-dependent body shape variation for real-time
  synthesis,'' in \emph{SCA}.\hskip 1em plus 0.5em minus 0.4em\relax ACM, 2006,
  pp. 147--156.

\bibitem{hasler2010learning}
N.~Hasler, T.~Thorm{\"a}hlen, B.~Rosenhahn, and H.-P. Seidel, ``Learning
  skeletons for shape and pose,'' in \emph{I3D}, 2010, pp. 23--30.

\bibitem{wang2020blsm}
H.~Wang, R.~A. G{\"u}ler, I.~Kokkinos, G.~Papandreou, and S.~Zafeiriou,
  ``{BLSM}: A bone-level skinned model of the human mesh,'' in
  \emph{ECCV}.\hskip 1em plus 0.5em minus 0.4em\relax Springer, 2020, pp.
  1--17.

\bibitem{CAESAR}
K.~M. Robinette, S.~Blackwell, H.~Daanen, M.~Boehmer, S.~Fleming, T.~Brill,
  D.~Hoeferlin, and D.~Burnsides, ``{Civilian American and European Surface
  Anthropometry Resource (CAESAR)} final report,'' {US Air Force Research
  Laboratory}, Tech. Rep. AFRL-HE-WP-TR-2002-0169, 2002.

\bibitem{zuffi2015stitched}
S.~Zuffi and M.~J. Black, ``The stitched puppet: A graphical model of {3D}
  human shape and pose,'' in \emph{CVPR}, 2015, pp. 3537--3546.

\bibitem{mohr2003building}
A.~Mohr and M.~Gleicher, ``Building efficient, accurate character skins from
  examples,'' \emph{TOG}, vol.~22, no.~3, pp. 562--568, 2003.

\bibitem{li2017learning}
T.~Li, T.~Bolkart, M.~J. Black, H.~Li, and J.~Romero, ``Learning a model of
  facial shape and expression from {2D} scans,'' \emph{TOG}, vol.~36, no.~6,
  pp. 194--1, 2017.

\bibitem{romero2017embodied}
J.~Romero, D.~Tzionas, and M.~J. Black, ``Embodied hands: Modeling and
  capturing hands and bodies together,'' \emph{TOG}, vol.~36, no.~6, pp. 1--17,
  2017.

\bibitem{hesse2018learning}
N.~Hesse, S.~Pujades, J.~Romero, M.~J. Black, C.~Bodensteiner, M.~Arens, U.~G.
  Hofmann, U.~Tacke, M.~Hadders-Algra, R.~Weinberger \emph{et~al.}, ``Learning
  an infant body model from {RGB-D} data for accurate full body motion
  analysis,'' in \emph{MICCAI}.\hskip 1em plus 0.5em minus 0.4em\relax
  Springer, 2018, pp. 792--800.

\bibitem{santesteban2020softsmpl}
I.~Santesteban, E.~Garces, M.~A. Otaduy, and D.~Casas, ``{SoftSMPL}:
  Data-driven modeling of nonlinear soft-tissue dynamics for parametric
  humans,'' in \emph{CGF}, vol.~39.\hskip 1em plus 0.5em minus 0.4em\relax
  Wiley Online Library, 2020, pp. 65--75.

\bibitem{osman2020star}
A.~A. Osman, T.~Bolkart, and M.~J. Black, ``{STAR}: Sparse trained articulated
  human body regressor,'' in \emph{ECCV}.\hskip 1em plus 0.5em minus
  0.4em\relax Springer, 2020, pp. 598--613.

\bibitem{deng2020nasa}
B.~Deng, J.~P. Lewis, T.~Jeruzalski, G.~Pons-Moll, G.~Hinton, M.~Norouzi, and
  A.~Tagliasacchi, ``{NASA}: Neural articulated shape approximation,'' in
  \emph{ECCV}.\hskip 1em plus 0.5em minus 0.4em\relax Springer, 2020, pp.
  612--628.

\bibitem{mihajlovic2021leap}
M.~Mihajlovic, Y.~Zhang, M.~J. Black, and S.~Tang, ``{LEAP}: Learning
  articulated occupancy of people,'' in \emph{CVPR}, 2021, pp.
  10\,461--10\,471.

\bibitem{chen2021snarf}
X.~Chen, Y.~Zheng, M.~J. Black, O.~Hilliges, and A.~Geiger, ``{SNARF}:
  Differentiable forward skinning for animating non-rigid neural implicit
  shapes,'' in \emph{ICCV}, 2021, pp. 11\,594--11\,604.

\bibitem{mihajlovic2022coap}
M.~Mihajlovic, S.~Saito, A.~Bansal, M.~Zollhoefer, and S.~Tang, ``{COAP}:
  Compositional articulated occupancy of people,'' in \emph{CVPR}, 2022, pp.
  13\,201--13\,210.

\bibitem{sun2023learning}
X.~Sun, Q.~Feng, X.~Li, J.~Zhang, Y.-K. Lai, J.~Yang, and K.~Li, ``Learning
  semantic-aware disentangled representation for flexible {3D} human body
  editing,'' in \emph{CVPR}, 2023.

\bibitem{joo2018total}
H.~Joo, T.~Simon, and Y.~Sheikh, ``{Total Capture}: A {3D} deformation model
  for tracking faces, hands, and bodies,'' in \emph{CVPR}, 2018, pp.
  8320--8329.

\bibitem{cao2013facewarehouse}
C.~Cao, Y.~Weng, S.~Zhou, Y.~Tong, and K.~Zhou, ``{FaceWarehouse}: a {3D}
  facial expression database for visual computing,'' \emph{TVCG}, vol.~20,
  no.~3, pp. 413--425, 2013.

\bibitem{osman2022supr}
A.~A. Osman, T.~Bolkart, D.~Tzionas, and M.~J. Black, ``{SUPR}: A sparse
  unified part-based human representation,'' in \emph{ECCV}.\hskip 1em plus
  0.5em minus 0.4em\relax Springer, 2022, pp. 568--585.

\bibitem{xu2020ghum}
H.~Xu, E.~G. Bazavan, A.~Zanfir, W.~T. Freeman, R.~Sukthankar, and
  C.~Sminchisescu, ``{GHUM \& GHUML}: Generative {3D} human shape and
  articulated pose models,'' in \emph{CVPR}, 2020, pp. 6184--6193.

\bibitem{balan2007detailed}
A.~O. Balan, L.~Sigal, M.~J. Black, J.~E. Davis, and H.~W. Haussecker,
  ``Detailed human shape and pose from images,'' in \emph{CVPR}.\hskip 1em plus
  0.5em minus 0.4em\relax IEEE, 2007, pp. 1--8.

\bibitem{loper2014mosh}
M.~Loper, N.~Mahmood, and M.~J. Black, ``{MoSh}: Motion and shape capture from
  sparse markers,'' \emph{TOG}, vol.~33, no.~6, pp. 1--13, 2014.

\bibitem{ionescu2014human3}
C.~Ionescu, D.~Papava, V.~Olaru, and C.~Sminchisescu, ``{Human3.6M}: Large
  scale datasets and predictive methods for {3D} human sensing in natural
  environments,'' \emph{TPAMI}, vol.~36, no.~7, pp. 1325--1339, 2014.

\bibitem{von2018recovering}
T.~von Marcard, R.~Henschel, M.~J. Black, B.~Rosenhahn, and G.~Pons-Moll,
  ``Recovering accurate {3D} human pose in the wild using {IMUs} and a moving
  camera,'' in \emph{ECCV}, 2018, pp. 601--617.

\bibitem{varol2017learning}
G.~Varol, J.~Romero, X.~Martin, N.~Mahmood, M.~J. Black, I.~Laptev, and
  C.~Schmid, ``Learning from synthetic humans,'' in \emph{CVPR}, 2017, pp.
  109--117.

\bibitem{mahmood2019amass}
N.~Mahmood, N.~Ghorbani, N.~F. Troje, G.~Pons-Moll, and M.~J. Black, ``{AMASS}:
  Archive of motion capture as surface shapes,'' in \emph{ICCV}, 2019, pp.
  5442--5451.

\bibitem{kocabas2020vibe}
M.~Kocabas, N.~Athanasiou, and M.~J. Black, ``{VIBE}: Video inference for human
  body pose and shape estimation,'' in \emph{CVPR}, 2020, pp. 5253--5263.

\bibitem{bualan2008naked}
A.~O. B{\u{a}}lan and M.~J. Black, ``The naked truth: Estimating body shape
  under clothing,'' in \emph{ECCV}.\hskip 1em plus 0.5em minus 0.4em\relax
  Springer, 2008, pp. 15--29.

\bibitem{guan2009estimating}
P.~Guan, A.~Weiss, A.~O. Balan, and M.~J. Black, ``Estimating human shape and
  pose from a single image,'' in \emph{ICCV}.\hskip 1em plus 0.5em minus
  0.4em\relax IEEE, 2009, pp. 1381--1388.

\bibitem{hasler2010multilinear}
N.~Hasler, H.~Ackermann, B.~Rosenhahn, T.~Thorm{\"a}hlen, and H.-P. Seidel,
  ``Multilinear pose and body shape estimation of dressed subjects from image
  sets,'' in \emph{CVPR}.\hskip 1em plus 0.5em minus 0.4em\relax IEEE, 2010,
  pp. 1823--1830.

\bibitem{zhou2010parametric}
S.~Zhou, H.~Fu, L.~Liu, D.~Cohen-Or, and X.~Han, ``Parametric reshaping of
  human bodies in images,'' \emph{TOG}, vol.~29, no.~4, pp. 1--10, 2010.

\bibitem{cmu_mocap}
``Carnegie mellon university - cmu graphics lab - motion capture library,''
  \url{http://mocap.cs.cmu.edu/}, 2010.

\bibitem{xiang2019monocular}
D.~Xiang, H.~Joo, and Y.~Sheikh, ``Monocular total capture: Posing face, body,
  and hands in the wild,'' in \emph{CVPR}, 2019, pp. 10\,965--10\,974.

\bibitem{guler2019holopose}
R.~A. G{\"u}ler and I.~Kokkinos, ``{HoloPose}: Holistic {3D} human
  reconstruction in-the-wild,'' in \emph{CVPR}, 2019, pp. 10\,884--10\,894.

\bibitem{guler2018densepose}
R.~A. G{\"u}ler, N.~Neverova, and I.~Kokkinos, ``{DensePose}: Dense human pose
  estimation in the wild,'' in \emph{CVPR}, 2018, pp. 7297--7306.

\bibitem{song2020human}
J.~Song, X.~Chen, and O.~Hilliges, ``Human body model fitting by learned
  gradient descent,'' in \emph{ECCV}.\hskip 1em plus 0.5em minus 0.4em\relax
  Springer, 2020, pp. 744--760.

\bibitem{iqbal2021kama}
U.~Iqbal, K.~Xie, Y.~Guo, J.~Kautz, and P.~Molchanov, ``{KAMA}: {3D} keypoint
  aware body mesh articulation,'' in \emph{3DV}.\hskip 1em plus 0.5em minus
  0.4em\relax IEEE, 2021, pp. 689--699.

\bibitem{yu2021skeleton2mesh}
Z.~Yu, J.~Wang, J.~Xu, B.~Ni, C.~Zhao, M.~Wang, and W.~Zhang,
  ``{Skeleton2Mesh}: Kinematics prior injected unsupervised human mesh
  recovery,'' in \emph{ICCV}, 2021, pp. 8619--8629.

\bibitem{li2021hybrik}
J.~Li, C.~Xu, Z.~Chen, S.~Bian, L.~Yang, and C.~Lu, ``{HybrIK}: A hybrid
  analytical-neural inverse kinematics solution for {3D} human pose and shape
  estimation,'' in \emph{CVPR}, 2021, pp. 3383--3393.

\bibitem{li2023niki}
J.~Li, S.~Bian, Q.~Liu, J.~Tang, F.~Wang, and C.~Lu, ``{NIKI}: Neural inverse
  kinematics with invertible neural networks for {3D} human pose and shape
  estimation,'' in \emph{CVPR}, 2023, pp. 12\,933--12\,942.

\bibitem{shetty2023pliks}
K.~Shetty, A.~Birkhold, S.~Jaganathan, N.~Strobel, M.~Kowarschik, A.~Maier, and
  B.~Egger, ``{PLIKS}: A pseudo-linear inverse kinematic solver for {3D} human
  body estimation,'' in \emph{CVPR}, 2023, pp. 574--584.

\bibitem{zhang2019danet}
H.~Zhang, J.~Cao, G.~Lu, W.~Ouyang, and Z.~Sun, ``{DaNet}:
  Decompose-and-aggregate network for {3D} human shape and pose estimation,''
  in \emph{ACM MM}, 2019, pp. 935--944.

\bibitem{rockwell2020full}
C.~Rockwell and D.~F. Fouhey, ``Full-body awareness from partial
  observations,'' in \emph{ECCV}.\hskip 1em plus 0.5em minus 0.4em\relax
  Springer, 2020, pp. 522--539.

\bibitem{sengupta2020synthetic}
A.~Sengupta, I.~Budvytis, and R.~Cipolla, ``Synthetic training for accurate
  {3D} human pose and shape estimation in the wild,'' in \emph{BMVC}, September
  2020.

\bibitem{xu20203d}
X.~Xu, H.~Chen, F.~Moreno-Noguer, L.~A. Jeni, and F.~De~la Torre, ``{3D} human
  shape and pose from a single low-resolution image with self-supervised
  learning,'' in \emph{ECCV}.\hskip 1em plus 0.5em minus 0.4em\relax Springer,
  2020, pp. 284--300.

\bibitem{zanfir2020weakly}
A.~Zanfir, E.~G. Bazavan, H.~Xu, W.~T. Freeman, R.~Sukthankar, and
  C.~Sminchisescu, ``Weakly supervised {3D} human pose and shape reconstruction
  with normalizing flows,'' in \emph{ECCV}.\hskip 1em plus 0.5em minus
  0.4em\relax Springer, 2020, pp. 465--481.

\bibitem{kocabas2021spec}
M.~Kocabas, C.-H.~P. Huang, J.~Tesch, L.~Muller, O.~Hilliges, and M.~J. Black,
  ``{SPEC}: Seeing people in the wild with an estimated camera,'' in
  \emph{ICCV}, 2021, pp. 11\,035--11\,045.

\bibitem{li2022cliff}
Z.~Li, J.~Liu, Z.~Zhang, S.~Xu, and Y.~Yan, ``{CLIFF}: Carrying location
  information in full frames into human pose and shape estimation,'' in
  \emph{ECCV}.\hskip 1em plus 0.5em minus 0.4em\relax Springer, 2022, pp.
  590--606.

\bibitem{kanazawa2019learning}
A.~Kanazawa, J.~Y. Zhang, P.~Felsen, and J.~Malik, ``Learning {3D} human
  dynamics from video,'' in \emph{CVPR}, 2019, pp. 5614--5623.

\bibitem{xu2019denserac}
Y.~Xu, S.-C. Zhu, and T.~Tung, ``{DenseRaC}: Joint {3D} pose and shape
  estimation by dense render-and-compare,'' in \emph{ICCV}, 2019, pp.
  7760--7770.

\bibitem{sun2019human}
Y.~Sun, Y.~Ye, W.~Liu, W.~Gao, Y.~Fu, and T.~Mei, ``Human mesh recovery from
  monocular images via a skeleton-disentangled representation,'' in
  \emph{ICCV}, 2019, pp. 5349--5358.

\bibitem{georgakis2020hierarchical}
G.~Georgakis, R.~Li, S.~Karanam, T.~Chen, J.~Ko{\v{s}}eck{\'a}, and Z.~Wu,
  ``Hierarchical kinematic human mesh recovery,'' in \emph{ECCV}.\hskip 1em
  plus 0.5em minus 0.4em\relax Springer, 2020, pp. 768--784.

\bibitem{zhou2019continuity}
Y.~Zhou, C.~Barnes, J.~Lu, J.~Yang, and H.~Li, ``On the continuity of rotation
  representations in neural networks,'' in \emph{CVPR}, 2019, pp. 5745--5753.

\bibitem{zhang2020learning}
H.~Zhang, J.~Cao, G.~Lu, W.~Ouyang, and Z.~Sun, ``Learning {3D} human shape and
  pose from dense body parts,'' \emph{TPAMI}, 2020.

\bibitem{luo20203d}
Z.~Luo, S.~A. Golestaneh, and K.~M. Kitani, ``{3D} human motion estimation via
  motion compression and refinement,'' in \emph{ACCV}, 2020.

\bibitem{zhou2021monocular}
Y.~Zhou, M.~Habermann, I.~Habibie, A.~Tewari, C.~Theobalt, and F.~Xu,
  ``Monocular real-time full body capture with inter-part correlations,'' in
  \emph{CVPR}, 2021, pp. 4811--4822.

\bibitem{choi2021beyond}
H.~Choi, G.~Moon, J.~Y. Chang, and K.~M. Lee, ``Beyond static features for
  temporally consistent {3D} human pose and shape from a video,'' in
  \emph{CVPR}, 2021, pp. 1964--1973.

\bibitem{Wang2023SGRE}
D.~Wang and S.~Zhang, ``{3D} human mesh recovery with sequentially global
  rotation estimation,'' in \emph{ICCV}, 2023, pp. 14\,953--14\,962.

\bibitem{varol2018bodynet}
G.~Varol, D.~Ceylan, B.~Russell, J.~Yang, E.~Yumer, I.~Laptev, and C.~Schmid,
  ``{BodyNet}: Volumetric inference of {3D} human body shapes,'' in
  \emph{ECCV}, 2018, pp. 20--36.

\bibitem{Zheng_2019_ICCV}
Z.~Zheng, T.~Yu, Y.~Wei, Q.~Dai, and Y.~Liu, ``{DeepHuman}: {3D} human
  reconstruction from a single image,'' in \emph{ICCV}, 2019, pp. 7738--7748.

\bibitem{kolotouros2019convolutional}
N.~Kolotouros, G.~Pavlakos, and K.~Daniilidis, ``Convolutional mesh regression
  for single-image human shape reconstruction,'' in \emph{CVPR}, 2019, pp.
  4501--4510.

\bibitem{moon2020i2l}
G.~Moon and K.~M. Lee, ``{I2L-MeshNet}: Image-to-lixel prediction network for
  accurate {3D} human pose and mesh estimation from a single {RGB} image,'' in
  \emph{ECCV}.\hskip 1em plus 0.5em minus 0.4em\relax Springer, 2020, pp.
  752--768.

\bibitem{lin2021mesh}
K.~Lin, L.~Wang, and Z.~Liu, ``{Mesh Graphormer},'' in \emph{ICCV}, 2021.

\bibitem{lin2021end}
------, ``End-to-end human pose and mesh reconstruction with transformers,'' in
  \emph{CVPR}, 2021, pp. 1954--1963.

\bibitem{luan2021pc}
T.~Luan, Y.~Wang, J.~Zhang, Z.~Wang, Z.~Zhou, and Y.~Qiao, ``{PC-HMR}: Pose
  calibration for {3D} human mesh recovery from {2D} images/videos,'' in
  \emph{AAAI}, vol.~35, no.~3, 2021, pp. 2269--2276.

\bibitem{zanfir2021thundr}
M.~Zanfir, A.~Zanfir, E.~G. Bazavan, W.~T. Freeman, R.~Sukthankar, and
  C.~Sminchisescu, ``{THUNDR}: Transformer-based {3D} human reconstruction with
  markers,'' in \emph{CVPR}, 2021.

\bibitem{yao2019densebody}
P.~Yao, Z.~Fang, F.~Wu, Y.~Feng, and J.~Li, ``{DenseBody}: Directly regressing
  dense {3D} human pose and shape from a single color image,'' \emph{arXiv
  preprint arXiv:1903.10153}, 2019.

\bibitem{zeng20203d}
W.~Zeng, W.~Ouyang, P.~Luo, W.~Liu, and X.~Wang, ``{3D} human mesh regression
  with dense correspondence,'' in \emph{CVPR}, 2020, pp. 7054--7063.

\bibitem{zhang2020object}
T.~Zhang, B.~Huang, and Y.~Wang, ``Object-occluded human shape and pose
  estimation from a single color image,'' in \emph{CVPR}, 2020, pp. 7376--7385.

\bibitem{biggs20203d}
B.~Biggs, D.~Novotny, S.~Ehrhardt, H.~Joo, B.~Graham, and A.~Vedaldi, ``{3D}
  multi-bodies: Fitting sets of plausible {3D} human models to ambiguous image
  data,'' \emph{NeurIPS}, vol.~33, 2020.

\bibitem{sengupta2021probabilistic}
A.~Sengupta, I.~Budvytis, and R.~Cipolla, ``Probabilistic {3D} human shape and
  pose estimation from multiple unconstrained images in the wild,'' in
  \emph{CVPR}, 2021, pp. 16\,094--16\,104.

\bibitem{kolotouros2021probabilistic}
N.~Kolotouros, G.~Pavlakos, D.~Jayaraman, and K.~Daniilidis, ``Probabilistic
  modeling for human mesh recovery,'' in \emph{ICCV}, 2021, pp.
  11\,605--11\,614.

\bibitem{sengupta2021hierarchical}
A.~Sengupta, I.~Budvytis, and R.~Cipolla, ``Hierarchical kinematic probability
  distributions for {3D} human shape and pose estimation from images in the
  wild,'' in \emph{ICCV}, 2021, pp. 11\,219--11\,229.

\bibitem{fang2023learning}
Q.~Fang, K.~Chen, Y.~Fan, Q.~Shuai, J.~Li, and W.~Zhang, ``Learning analytical
  posterior probability for human mesh recovery,'' in \emph{CVPR}, 2023, pp.
  8781--8791.

\bibitem{sengupta2023humaniflow}
A.~Sengupta, I.~Budvytis, and R.~Cipolla, ``{HuManiFlow}: Ancestor-conditioned
  normalising flows on {SO} (3) manifolds for human pose and shape distribution
  estimation,'' in \emph{CVPR}, 2023, pp. 4779--4789.

\bibitem{rueegg2020chained}
N.~Rueegg, C.~Lassner, M.~Black, and K.~Schindler, ``Chained representation
  cycling: Learning to estimate {3D} human pose and shape by cycling between
  representations,'' in \emph{AAAI}, vol.~34, 2020, pp. 5561--5569.

\bibitem{zanfir2021neural}
A.~Zanfir, E.~G. Bazavan, M.~Zanfir, W.~T. Freeman, R.~Sukthankar, and
  C.~Sminchisescu, ``Neural descent for visual {3D} human pose and shape,'' in
  \emph{CVPR}, 2021, pp. 14\,484--14\,493.

\bibitem{doersch2019sim2real}
C.~Doersch and A.~Zisserman, ``Sim2real transfer learning for {3D} human pose
  estimation: motion to the rescue,'' \emph{NeurIPS}, vol.~32, pp.
  12\,949--12\,961, 2019.

\bibitem{choi2020pose2mesh}
H.~Choi, G.~Moon, and K.~M. Lee, ``{Pose2Mesh}: Graph convolutional network for
  {3D} human pose and mesh recovery from a {2D} human pose,'' in \emph{ECCV},
  2020.

\bibitem{li2022deep}
Z.~Li, B.~Xu, H.~Huang, C.~Lu, and Y.~Guo, ``Deep two-stream video inference
  for human body pose and shape estimation,'' in \emph{WACV}, 2022, pp.
  430--439.

\bibitem{gong2022self}
X.~Gong, M.~Zheng, B.~Planche, S.~Karanam, T.~Chen, D.~Doermann, and Z.~Wu,
  ``Self-supervised human mesh recovery with cross-representation alignment,''
  in \emph{ECCV}.\hskip 1em plus 0.5em minus 0.4em\relax Springer, 2022, pp.
  212--230.

\bibitem{choi20213dcrowdnet}
H.~Choi, G.~Moon, J.~Park, and K.~M. Lee, ``Learning to estimate robust {3D}
  human mesh from in-the-wild crowded scenes,'' in \emph{CVPR}, 2022.

\bibitem{pavlakos2022human}
G.~Pavlakos, J.~Malik, and A.~Kanazawa, ``Human mesh recovery from multiple
  shots,'' in \emph{CVPR}, 2022, pp. 1485--1495.

\bibitem{rong2019delving}
Y.~Rong, Z.~Liu, C.~Li, K.~Cao, and C.~C. Loy, ``Delving deep into hybrid
  annotations for {3D} human recovery in the wild,'' in \emph{ICCV}, 2019, pp.
  5340--5348.

\bibitem{dwivedi2021learning}
S.~K. Dwivedi, N.~Athanasiou, M.~Kocabas, and M.~J. Black, ``Learning to
  regress bodies from images using differentiable semantic rendering,'' in
  \emph{ICCV}, 2021, pp. 11\,250--11\,259.

\bibitem{he2016deep}
K.~He, X.~Zhang, S.~Ren, and J.~Sun, ``Deep residual learning for image
  recognition,'' in \emph{CVPR}, 2016, pp. 770--778.

\bibitem{wang2020deep}
J.~Wang, K.~Sun, T.~Cheng, B.~Jiang, C.~Deng, Y.~Zhao, D.~Liu, Y.~Mu, M.~Tan,
  X.~Wang \emph{et~al.}, ``Deep high-resolution representation learning for
  visual recognition,'' \emph{TPAMI}, 2020.

\bibitem{cho2023implicit}
H.~Cho, Y.~Cho, J.~Ahn, and J.~Kim, ``Implicit {3D} human mesh recovery using
  consistency with pose and shape from unseen-view,'' in \emph{CVPR}, 2023, pp.
  21\,148--21\,158.

\bibitem{goel20234d}
S.~Goel, G.~Pavlakos, J.~Rajasegaran, A.~Kanazawa, and J.~Malik, ``Humans in
  {4D}: Reconstructing and tracking humans with transformers,'' in \emph{ICCV},
  2023.

\bibitem{dosovitskiy2020image}
A.~Dosovitskiy, L.~Beyer, A.~Kolesnikov, D.~Weissenborn, X.~Zhai,
  T.~Unterthiner, M.~Dehghani, M.~Minderer, G.~Heigold, S.~Gelly \emph{et~al.},
  ``An image is worth 16x16 words: Transformers for image recognition at
  scale,'' in \emph{ICLR}, 2021.

\bibitem{cho2022cross}
J.~Cho, K.~Youwang, and T.-H. Oh, ``Cross-attention of disentangled modalities
  for {3D} human mesh recovery with transformers,'' in \emph{ECCV}.\hskip 1em
  plus 0.5em minus 0.4em\relax Springer, 2022, pp. 342--359.

\bibitem{zheng2023potter}
C.~Zheng, X.~Liu, G.-J. Qi, and C.~Chen, ``{POTTER}: Pooling attention
  transformer for efficient human mesh recovery,'' in \emph{CVPR}, 2023, pp.
  1611--1620.

\bibitem{Dou2023TORE}
Z.~Dou, Q.~Wu, C.~Lin, Z.~Cao, Q.~Wu, W.~Wan, T.~Komura, and W.~Wang, ``{TORE}:
  Token reduction for efficient human mesh recovery with transformer,'' in
  \emph{ICCV}, 2023, pp. 15\,143--15\,155.

\bibitem{kim2023sampling}
J.~Kim, M.-G. Gwon, H.~Park, H.~Kwon, G.-M. Um, and W.~Kim, ``Sampling is
  matter: Point-guided {3D} human mesh reconstruction,'' in \emph{CVPR}, 2023,
  pp. 12\,880--12\,889.

\bibitem{yoshiyasu2023deformable}
Y.~Yoshiyasu, ``Deformable mesh transformer for {3D} human mesh recovery,'' in
  \emph{CVPR}, 2023, pp. 17\,006--17\,015.

\bibitem{Li2023JOTR}
J.~Li, Z.~Yang, X.~Wang, J.~Ma, C.~Zhou, and Y.~Yang, ``{JOTR}: {3D} joint
  contrastive learning with transformers for occluded human mesh recovery,'' in
  \emph{ICCV}, 2023, pp. 9110--9121.

\bibitem{choutas2022accurate}
V.~Choutas, L.~M{\"u}ller, C.-H.~P. Huang, S.~Tang, D.~Tzionas, and M.~J.
  Black, ``Accurate {3D} body shape regression using metric and semantic
  attributes,'' in \emph{CVPR}, 2022, pp. 2718--2728.

\bibitem{ma20233d}
X.~Ma, J.~Su, C.~Wang, W.~Zhu, and Y.~Wang, ``{3D} human mesh estimation from
  virtual markers,'' in \emph{CVPR}, 2023, pp. 534--543.

\bibitem{fan2021revitalizing}
T.~Fan, K.~V. Alwala, D.~Xiang, W.~Xu, T.~Murphey, and M.~Mukadam,
  ``Revitalizing optimization for {3D} human pose and shape estimation: A
  sparse constrained formulation,'' in \emph{ICCV}, 2021.

\bibitem{zhang2023pymafx}
H.~Zhang, Y.~Tian, Y.~Zhang, M.~Li, L.~An, Z.~Sun, and Y.~Liu, ``{PyMAF-X}:
  Towards well-aligned full-body model regression from monocular images,''
  \emph{TPAMI}, 2023.

\bibitem{wang2022best}
Z.~Wang, J.~Yang, and C.~Fowlkes, ``The best of both worlds: combining
  model-based and nonparametric approaches for {3D} human body estimation,'' in
  \emph{CVPRW}, 2022, pp. 2318--2327.

\bibitem{jiang2020coherent}
W.~Jiang, N.~Kolotouros, G.~Pavlakos, X.~Zhou, and K.~Daniilidis, ``Coherent
  reconstruction of multiple humans from a single image,'' in \emph{CVPR},
  2020, pp. 5579--5588.

\bibitem{ugrinovic2021body}
N.~Ugrinovic, A.~Ruiz, A.~Agudo, A.~Sanfeliu, and F.~Moreno-Noguer, ``Body size
  and depth disambiguation in multi-person reconstruction from single images,''
  in \emph{3DV}.\hskip 1em plus 0.5em minus 0.4em\relax IEEE, 2021, pp. 53--63.

\bibitem{fieraru2021remips}
M.~Fieraru, M.~Zanfir, T.~Szente, E.~Bazavan, V.~Olaru, and C.~Sminchisescu,
  ``{REMIPS}: Physically consistent {3D} reconstruction of multiple interacting
  people under weak supervision,'' \emph{NeurIPS}, vol.~34, pp.
  19\,385--19\,397, 2021.

\bibitem{cha2022multi}
J.~Cha, M.~Saqlain, G.~Kim, M.~Shin, and S.~Baek, ``Multi-person {3D} pose and
  shape estimation via inverse kinematics and refinement,'' in
  \emph{ECCV}.\hskip 1em plus 0.5em minus 0.4em\relax Springer, 2022, pp.
  660--677.

\bibitem{khirodkar2022occluded}
R.~Khirodkar, S.~Tripathi, and K.~Kitani, ``Occluded human mesh recovery,'' in
  \emph{CVPR}, 2022, pp. 1715--1725.

\bibitem{zanfir2018deep}
A.~Zanfir, E.~Marinoiu, M.~Zanfir, A.-I. Popa, and C.~Sminchisescu, ``Deep
  network for the integrated {3D} sensing of multiple people in natural
  images,'' in \emph{NeurIPS}, vol.~31, 2018, pp. 8410--8419.

\bibitem{sun2021monocular}
Y.~Sun, Q.~Bao, W.~Liu, Y.~Fu, M.~J. Black, and T.~Mei, ``Monocular, one-stage,
  regression of multiple {3D} people,'' in \emph{ICCV}, 2021, pp.
  11\,179--11\,188.

\bibitem{sun2021putting}
Y.~Sun, W.~Liu, Q.~Bao, Y.~Fu, T.~Mei, and M.~J. Black, ``Putting people in
  their place: Monocular regression of {3D} people in depth,'' in \emph{CVPR},
  2021.

\bibitem{qiu2023psvt}
Z.~Qiu, Q.~Yang, J.~Wang, H.~Feng, J.~Han, E.~Ding, C.~Xu, D.~Fu, and J.~Wang,
  ``{PSVT}: End-to-end multi-person {3D} pose and shape estimation with
  progressive video transformers,'' in \emph{CVPR}, 2023, pp. 21\,254--21\,263.

\bibitem{arnab2019exploiting}
A.~Arnab, C.~Doersch, and A.~Zisserman, ``Exploiting temporal context for {3D}
  human pose estimation in the wild,'' in \emph{CVPR}, 2019, pp. 3395--3404.

\bibitem{lee2021uncertainty}
G.-H. Lee and S.-W. Lee, ``Uncertainty-aware human mesh recovery from video by
  learning part-based {3D} dynamics,'' in \emph{ICCV}, 2021, pp.
  12\,375--12\,384.

\bibitem{wan2021encoder}
Z.~Wan, Z.~Li, M.~Tian, J.~Liu, S.~Yi, and H.~Li, ``Encoder-decoder with
  multi-level attention for {3D} human shape and pose estimation,'' in
  \emph{ICCV}, 2021, pp. 13\,033--13\,042.

\bibitem{yuan2021simpoe}
Y.~Yuan, S.-E. Wei, T.~Simon, K.~Kitani, and J.~Saragih, ``{SimPoE}: Simulated
  character control for {3D} human pose estimation,'' in \emph{CVPR}, 2021, pp.
  7159--7169.

\bibitem{wei2022capturing}
W.-L. Wei, J.-C. Lin, T.-L. Liu, and H.-Y.~M. Liao, ``Capturing humans in
  motion: temporal-attentive {3D} human pose and shape estimation from
  monocular video,'' in \emph{CVPR}, 2022, pp. 13\,211--13\,220.

\bibitem{rajasegaran2021tracking}
J.~Rajasegaran, G.~Pavlakos, A.~Kanazawa, and J.~Malik, ``Tracking people with
  {3D} representations,'' in \emph{NeurIPS}, 2021.

\bibitem{yuan2022glamr}
Y.~Yuan, U.~Iqbal, P.~Molchanov, K.~Kitani, and J.~Kautz, ``{GLAMR}: Global
  occlusion-aware human mesh recovery with dynamic cameras,'' in \emph{CVPR},
  2022, pp. 11\,038--11\,049.

\bibitem{sigal2007combined}
L.~Sigal, A.~Balan, and M.~Black, ``Combined discriminative and generative
  articulated pose and non-rigid shape estimation,'' \emph{NeurIPS}, vol.~20,
  pp. 1337--1344, 2007.

\bibitem{hassan2019resolving}
M.~Hassan, V.~Choutas, D.~Tzionas, and M.~J. Black, ``Resolving {3D} human pose
  ambiguities with {3D} scene constraints,'' in \emph{ICCV}, 2019, pp.
  2282--2292.

\bibitem{shi2020motionet}
M.~Shi, K.~Aberman, A.~Aristidou, T.~Komura, D.~Lischinski, D.~Cohen-Or, and
  B.~Chen, ``{MotioNet}: {3D} human motion reconstruction from monocular video
  with skeleton consistency,'' \emph{TOG}, vol.~40, no.~1, pp. 1--15, 2020.

\bibitem{zhang2021learning}
S.~Zhang, Y.~Zhang, F.~Bogo, M.~Pollefeys, and S.~Tang, ``Learning motion
  priors for {2D} human body capture in {3D} scenes,'' in \emph{ICCV}, 2021,
  pp. 11\,343--11\,353.

\bibitem{muller2021self}
L.~M{\"{u}}ller, A.~A. Osman, S.~Tang, C.-H.~P. Huang, and M.~J. Black, ``On
  self-contact and human pose,'' in \emph{CVPR}, 2021, pp. 9990--9999.

\bibitem{rempe2020contact}
D.~Rempe, L.~J. Guibas, A.~Hertzmann, B.~Russell, R.~Villegas, and J.~Yang,
  ``Contact and human dynamics from monocular video,'' in \emph{ECCV}.\hskip
  1em plus 0.5em minus 0.4em\relax Springer, 2020, pp. 71--87.

\bibitem{rempe2021humor}
D.~Rempe, T.~Birdal, A.~Hertzmann, J.~Yang, S.~Sridhar, and L.~J. Guibas,
  ``{HuMoR}: {3D} human motion model for robust pose estimation,'' in
  \emph{ICCV}, 2021.

\bibitem{akhter2015pose}
I.~Akhter and M.~J. Black, ``Pose-conditioned joint angle limits for {3D} human
  pose reconstruction,'' in \emph{CVPR}, 2015, pp. 1446--1455.

\bibitem{zhang2021body}
J.~Zhang, D.~Yu, J.~H. Liew, X.~Nie, and J.~Feng, ``Body meshes as points,'' in
  \emph{CVPR}, 2021, pp. 546--556.

\bibitem{Baek_2019_CVPR}
S.~Baek, K.~I. Kim, and T.-K. Kim, ``Pushing the envelope for {RGB}-based dense
  {3D} hand pose estimation via neural rendering,'' in \emph{CVPR}, 2019, pp.
  1067--1076.

\bibitem{Boukhayma2019}
A.~Boukhayma, R.~d. Bem, and P.~H. Torr, ``{3D} hand shape and pose from images
  in the wild,'' in \emph{CVPR}, 2019, pp. 10\,843--10\,852.

\bibitem{honnotate2020}
S.~Hampali, M.~Rad, M.~Oberweger, and V.~Lepetit, ``{HOnnotate}: A method for
  {3D} annotation of hand and object poses,'' in \emph{CVPR}, 2020, pp.
  3193--3203.

\bibitem{hasson_2019_cvpr}
Y.~Hasson, G.~Varol, D.~Tzionas, I.~Kalevatykh, M.~J. Black, I.~Laptev, and
  C.~Schmid, ``Learning joint reconstruction of hands and manipulated
  objects,'' in \emph{CVPR}, 2019, pp. 11\,807--11\,816.

\bibitem{Iqbal_2018_ECCV}
U.~Iqbal, P.~Molchanov, T.~Breuel, J.~Gall, and J.~Kautz, ``Hand pose
  estimation via latent {2.5D} heatmap regression,'' in \emph{ECCV}, 2018, pp.
  125--143.

\bibitem{kulon2019rec}
D.~Kulon, H.~Wang, R.~A. G{\"{u}}ler, M.~M. Bronstein, and S.~Zafeiriou,
  ``Single image {3D} hand reconstruction with mesh convolutions,'' in
  \emph{BMVC}, 2019.

\bibitem{Mueller_2018_GANerated}
F.~Mueller, F.~Bernard, O.~Sotnychenko, D.~Mehta, S.~Sridhar, D.~Casas, and
  C.~Theobalt, ``{GANerated} hands for real-time {3D} hand tracking from
  monocular {RGB},'' in \emph{CVPR}, 2018, pp. 49--59.

\bibitem{Tekin_2019_CVPR}
B.~Tekin, F.~Bogo, and M.~Pollefeys, ``{H+O}: Unified egocentric recognition of
  {3D} hand-object poses and interactions,'' in \emph{CVPR}, 2019, pp.
  4511--4520.

\bibitem{brox_ICCV_2017}
C.~Zimmermann and T.~Brox, ``Learning to estimate {3D} hand pose from single
  {RGB} images,'' in \emph{ICCV}, 2017, pp. 4913--4921.

\bibitem{Zhang_2019_ICCV}
X.~Zhang, Q.~Li, H.~Mo, W.~Zhang, and W.~Zheng, ``End-to-end hand mesh recovery
  from a monocular {RGB} image,'' in \emph{ICCV}, 2019, pp. 2354--2364.

\bibitem{Ge2019}
L.~Ge, Z.~Ren, Y.~Li, Z.~Xue, Y.~Wang, J.~Cai, and J.~Yuan, ``{3D} hand shape
  and pose estimation from a single {RGB} image,'' in \emph{CVPR}, 2019, pp.
  10\,833--10\,842.

\bibitem{Kulon_2020_CVPR}
D.~Kulon, R.~A. G{\"u}ler, I.~Kokkinos, M.~M. Bronstein, and S.~Zafeiriou,
  ``Weakly-supervised mesh-convolutional hand reconstruction in the wild,'' in
  \emph{CVPR}, 2020, pp. 4989--4999.

\bibitem{park2022handoccnet}
J.~Park, Y.~Oh, G.~Moon, H.~Choi, and K.~M. Lee, ``{HandOccNet}:
  Occlusion-robust {3D} hand mesh estimation network,'' in \emph{CVPR}, 2022,
  pp. 1496--1505.

\bibitem{moon2020interhand2}
G.~Moon, S.-I. Yu, H.~Wen, T.~Shiratori, and K.~M. Lee, ``{InterHand2.6M}: A
  dataset and baseline for {3D} interacting hand pose estimation from a single
  {RGB} image,'' in \emph{ECCV}.\hskip 1em plus 0.5em minus 0.4em\relax
  Springer, 2020, pp. 548--564.

\bibitem{wang2020rgb2hands}
J.~Wang, F.~Mueller, F.~Bernard, S.~Sorli, O.~Sotnychenko, N.~Qian, M.~A.
  Otaduy, D.~Casas, and C.~Theobalt, ``{RGB2Hands}: real-time tracking of {3D}
  hand interactions from monocular {RGB} video,'' \emph{ACM TOG}, vol.~39,
  no.~6, pp. 1--16, 2020.

\bibitem{zhang2021interacting}
B.~Zhang, Y.~Wang, X.~Deng, Y.~Zhang, P.~Tan, C.~Ma, and H.~Wang, ``Interacting
  two-hand {3D} pose and shape reconstruction from single color image,'' in
  \emph{ICCV}, 2021, pp. 11\,354--11\,363.

\bibitem{li2022interacting}
M.~Li, L.~An, H.~Zhang, L.~Wu, F.~Chen, T.~Yu, and Y.~Liu, ``Interacting
  attention graph for single image two-hand reconstruction,'' in \emph{CVPR},
  2022.

\bibitem{wang2023memahand}
C.~Wang, F.~Zhu, and S.~Wen, ``{MeMaHand}: Exploiting mesh-mano interaction for
  single image two-hand reconstruction,'' in \emph{CVPR}, 2023, pp. 564--573.

\bibitem{lee2023im2hands}
J.~Lee, M.~Sung, H.~Choi, and T.-K. Kim, ``{Im2Hands}: Learning attentive
  implicit representation of interacting two-hand shapes,'' in \emph{CVPR},
  2023, pp. 21\,169--21\,178.

\bibitem{yu2023acr}
Z.~Yu, S.~Huang, C.~Fang, T.~P. Breckon, and J.~Wang, ``{ACR}: Attention
  collaboration-based regressor for arbitrary two-hand reconstruction,'' in
  \emph{CVPR}, 2023, pp. 12\,955--12\,964.

\bibitem{moon2023bringing}
G.~Moon, ``Bringing inputs to shared domains for {3D} interacting hands
  recovery in the wild,'' in \emph{CVPR}, 2023, pp. 17\,028--17\,037.

\bibitem{chatzis2020comprehensive}
T.~Chatzis, A.~Stergioulas, D.~Konstantinidis, K.~Dimitropoulos, and P.~Daras,
  ``A comprehensive study on deep learning-based {3D} hand pose estimation
  methods,'' \emph{Applied Sciences}, vol.~10, no.~19, p. 6850, 2020.

\bibitem{huang2021survey}
L.~Huang, B.~Zhang, Z.~Guo, Y.~Xiao, Z.~Cao, and J.~Yuan, ``Survey on depth and
  {RGB} image-based {3D} hand shape and pose estimation,'' \emph{Virtual
  Reality \& Intelligent Hardware}, vol.~3, no.~3, pp. 207--234, 2021.

\bibitem{AldrianSmith2013}
O.~Aldrian and W.~A. Smith, ``Inverse rendering of faces with a {3D} morphable
  model,'' \emph{TPAMI}, vol.~35, no.~5, pp. 1080--1093, 2013.

\bibitem{VetterBlanz1998}
T.~Vetter and V.~Blanz, ``Estimating coloured {3D} face models from single
  images: An example based approach,'' in \emph{ECCV}, 1998, pp. 499--513.

\bibitem{Thies2016}
J.~Thies, M.~Zollh{\"o}fer, M.~Stamminger, C.~Theobalt, and M.~Nie{\ss}ner,
  ``{Face2Face}: Real-time face capture and reenactment of {RGB} videos,'' in
  \emph{CVPR}, 2016, pp. 2387--2395.

\bibitem{Feng2018}
Y.~Feng, F.~Wu, X.~Shao, Y.~Wang, and X.~Zhou, ``Joint {3D} face reconstruction
  and dense alignment with position map regression network,'' in \emph{ECCV},
  2018, pp. 557--574.

\bibitem{Jackson2017}
A.~S. Jackson, A.~Bulat, V.~Argyriou, and G.~Tzimiropoulos, ``Large pose {3D}
  face reconstruction from a single image via direct volumetric {CNN}
  regression,'' in \emph{ICCV}, 2017, pp. 1031--1039.

\bibitem{Sanyal2019_ringnet}
S.~Sanyal, T.~Bolkart, H.~Feng, and M.~J. Black, ``Learning to regress {3D}
  face shape and expression from an image without {3D} supervision,'' in
  \emph{CVPR}, 2019, pp. 7763--7772.

\bibitem{Tewari2018}
A.~Tewari, M.~Zollh{\"o}fer, P.~Garrido, F.~Bernard, H.~Kim, P.~P{\'e}rez, and
  C.~Theobalt, ``Self-supervised multi-level face model learning for monocular
  reconstruction at over 250 {Hz},'' in \emph{CVPR}, 2018, pp. 2549--2559.

\bibitem{Tewari2017}
A.~Tewari, M.~Zollh{\"o}fer, H.~Kim, P.~Garrido, F.~Bernard, P.~Perez, and
  C.~Theobalt, ``{MoFA:} model-based deep convolutional face autoencoder for
  unsupervised monocular reconstruction,'' in \emph{ICCV}, 2017, pp.
  3735--3744.

\bibitem{Deng2019}
Y.~Deng, J.~Yang, S.~Xu, D.~Chen, Y.~Jia, and X.~Tong, ``Accurate {3D} face
  reconstruction with weakly-supervised learning: From single image to image
  set,'' in \emph{CVPRW}, 2019, pp. 285--295.

\bibitem{LuanTran2019}
L.~Tran, F.~Liu, and X.~Liu, ``Towards high-fidelity nonlinear {3D} face
  morphable model,'' in \emph{CVPR}, 2019, pp. 1126--1135.

\bibitem{DECA_2020}
Y.~Feng, H.~Feng, M.~J. Black, and T.~Bolkart, ``Learning an animatable
  detailed {3D} face model from in-the-wild images,'' \emph{TOG}, vol.~40,
  no.~4, pp. 88:1--88:13, 2021.

\bibitem{wang2022faceverse}
L.~Wang, Z.~Chen, T.~Yu, C.~Ma, L.~Li, and Y.~Liu, ``{FaceVerse}: a
  fine-grained and detail-controllable {3D} face morphable model from a hybrid
  dataset,'' in \emph{CVPR}, 2022.

\bibitem{zielonka2022towards}
W.~Zielonka, T.~Bolkart, and J.~Thies, ``Towards metrical reconstruction of
  human faces,'' in \emph{ECCV}.\hskip 1em plus 0.5em minus 0.4em\relax
  Springer, 2022, pp. 250--269.

\bibitem{Loper:ECCV:2014}
M.~M. Loper and M.~J. Black, ``{OpenDR}: An approximate differentiable
  renderer,'' in \emph{ECCV}, 2014, pp. 154--169.

\bibitem{ravi2020pytorch3d}
N.~Ravi, J.~Reizenstein, D.~Novotny, T.~Gordon, W.-Y. Lo, J.~Johnson, and
  G.~Gkioxari, ``Accelerating {3D} deep learning with {PyTorch3D},''
  \emph{arXiv preprint arXiv:2007.08501}, 2020.

\bibitem{Genova2018}
K.~Genova, F.~Cole, A.~Maschinot, A.~Sarna, D.~Vlasic, and W.~T. Freeman,
  ``Unsupervised training for {3D} morphable model regression,'' in
  \emph{CVPR}, 2018, pp. 8377--8386.

\bibitem{Cao2018_VGGFace2}
Q.~Cao, L.~Shen, W.~Xie, O.~M. Parkhi, and A.~Zisserman, ``{VGGFace2}: A
  dataset for recognising faces across pose and age,'' in \emph{FG}, 2018, pp.
  67--74.

\bibitem{Egger_3DMM_survey}
B.~Egger, W.~A.~P. Smith, A.~Tewari, S.~Wuhrer, M.~Zollhoefer, T.~Beeler,
  F.~Bernard, T.~Bolkart, A.~Kortylewski, S.~Romdhani, C.~Theobalt, V.~Blanz,
  and T.~Vetter, ``{3D} morphable face models - past, present and future,''
  \emph{TOG}, vol.~39, no.~5, pp. 157:1--157:38, 2020.

\bibitem{yi2023generating}
H.~Yi, H.~Liang, Y.~Liu, Q.~Cao, Y.~Wen, T.~Bolkart, D.~Tao, and M.~J. Black,
  ``Generating holistic {3D} human motion from speech,'' in \emph{CVPR}, 2023,
  pp. 469--480.

\bibitem{zioulis2023kbody}
N.~Zioulis and J.~F. O'Brien, ``{KBody}: Towards general, robust, and aligned
  monocular whole-body estimation,'' in \emph{CVPRW}, 2023, pp. 6214--6224.

\bibitem{simon2017hand}
T.~Simon, H.~Joo, I.~Matthews, and Y.~Sheikh, ``Hand keypoint detection in
  single images using multiview bootstrapping,'' in \emph{CVPR}, 2017, pp.
  4645--4653.

\bibitem{rong2020frankmocap}
Y.~Rong, T.~Shiratori, and H.~Joo, ``{FrankMocap}: A monocular {3D} whole-body
  pose estimation system via regression and integration,'' in \emph{ICCVW},
  2021.

\bibitem{li2023hybrikx}
J.~Li, S.~Bian, C.~Xu, Z.~Chen, L.~Yang, and C.~Lu, ``{HybrIK-X}: Hybrid
  analytical-neural inverse kinematics for whole-body mesh recovery,''
  \emph{arXiv preprint arXiv:2304.05690}, 2023.

\bibitem{sun2022learning}
Y.~Sun, T.~Huang, Q.~Bao, W.~Liu, G.~Wenpeng, and Y.~Fu, ``Learning monocular
  mesh recovery of multiple body parts via synthesis,'' in \emph{ICASSP}, 2022.

\bibitem{lin2023one}
J.~Lin, A.~Zeng, H.~Wang, L.~Zhang, and Y.~Li, ``One-stage {3D} whole-body mesh
  recovery with component aware transformer,'' in \emph{CVPR}, 2023, pp.
  21\,159--21\,168.

\bibitem{cai2023smplerx}
Z.~Cai, W.~Yin, A.~Zeng, C.~Wei, Q.~Sun, Y.~Wang, H.~E. Pang, H.~Mei, M.~Zhang,
  L.~Zhang, C.~C. Loy, L.~Yang, and Z.~Liu, ``{SMPLer-X}: Scaling up expressive
  human pose and shape estimation,'' in \emph{NeurIPS Datasets and Benchmarks},
  2023.

\bibitem{pang2023robosmplx}
H.~E. Pang, Z.~Cai, L.~Yang, T.~Qingyi, W.~Zhonghua, T.~Zhang, and Z.~Liu,
  ``Towards robust and expressive whole-body human pose and shape estimation,''
  \emph{NeurIPS}, 2023.

\bibitem{forte2023reconstructing}
M.-P. Forte, P.~Kulits, C.-H.~P. Huang, V.~Choutas, D.~Tzionas, K.~J.
  Kuchenbecker, and M.~J. Black, ``Reconstructing signing avatars from video
  using linguistic priors,'' in \emph{CVPR}, 2023, pp. 12\,791--12\,801.

\bibitem{redmon2018yolov3}
J.~Redmon and A.~Farhadi, ``{YOLOv3}: An incremental improvement,'' \emph{arXiv
  preprint arXiv:1804.02767}, 2018.

\bibitem{he2017mask}
K.~He, G.~Gkioxari, P.~Doll{\'a}r, and R.~Girshick, ``{Mask R-CNN},'' in
  \emph{ICCV}, 2017, pp. 2961--2969.

\bibitem{wen2023crowd3d}
H.~Wen, J.~Huang, H.~Cui, H.~Lin, Y.-K. Lai, L.~Fang, and K.~Li, ``{Crowd3D}:
  Towards hundreds of people reconstruction from a single image,'' in
  \emph{CVPR}, 2023, pp. 8937--8946.

\bibitem{zhang2023two}
B.~Zhang, K.~Ma, S.~Wu, and Z.~Yuan, ``Two-stage co-segmentation network based
  on discriminative representation for recovering human mesh from videos,'' in
  \emph{CVPR}, 2023, pp. 5662--5670.

\bibitem{wang2018non}
X.~Wang, R.~Girshick, A.~Gupta, and K.~He, ``Non-local neural networks,'' in
  \emph{CVPR}, 2018, pp. 7794--7803.

\bibitem{vaswani2017attention}
A.~Vaswani, N.~Shazeer, N.~Parmar, J.~Uszkoreit, L.~Jones, A.~N. Gomez,
  {\L}.~Kaiser, and I.~Polosukhin, ``Attention is all you need,'' in
  \emph{NeurIPS}, 2017, pp. 5998--6008.

\bibitem{shen2023global}
X.~Shen, Z.~Yang, X.~Wang, J.~Ma, C.~Zhou, and Y.~Yang, ``Global-to-local
  modeling for video-based {3D} human pose and shape estimation,'' in
  \emph{CVPR}, 2023, pp. 8887--8896.

\bibitem{Cho2023Video}
H.~Cho, J.~Ahn, Y.~Cho, and J.~Kim, ``Video inference for human mesh recovery
  with vision transformer,'' in \emph{FG}, 2023, pp. 1--6.

\bibitem{guan2021bilevel}
S.~Guan, J.~Xu, Y.~Wang, B.~Ni, and X.~Yang, ``Bilevel online adaptation for
  out-of-domain human mesh reconstruction,'' in \emph{CVPR}, 2021, pp.
  10\,472--10\,481.

\bibitem{Nam2023Cyclic}
H.~Nam, D.~S. Jung, Y.~Oh, and K.~M. Lee, ``Cyclic test-time adaptation on
  monocular video for {3D} human mesh reconstruction,'' in \emph{ICCV}, 2023,
  pp. 14\,829--14\,839.

\bibitem{tripathi2020posenet3d}
S.~Tripathi, S.~Ranade, A.~Tyagi, and A.~Agrawal, ``{PoseNet3D}: Learning
  temporally consistent {3D} human pose via knowledge distillation,'' in
  \emph{3DV}.\hskip 1em plus 0.5em minus 0.4em\relax IEEE, 2020, pp. 311--321.

\bibitem{ye2023decoupling}
V.~Ye, G.~Pavlakos, J.~Malik, and A.~Kanazawa, ``Decoupling human and camera
  motion from videos in the wild,'' in \emph{CVPR}, 2023, pp. 21\,222--21\,232.

\bibitem{sun2023trace}
Y.~Sun, Q.~Bao, W.~Liu, T.~Mei, and M.~J. Black, ``{TRACE}: {5D} temporal
  regression of avatars with dynamic cameras in {3D} environments,'' in
  \emph{CVPR}, 2023, pp. 8856--8866.

\bibitem{li2022d}
J.~Li, S.~Bian, C.~Xu, G.~Liu, G.~Yu, and C.~Lu, ``{D \&D}: Learning human
  dynamics from dynamic camera,'' in \emph{ECCV}.\hskip 1em plus 0.5em minus
  0.4em\relax Springer, 2022, pp. 479--496.

\bibitem{weng2021holistic}
Z.~Weng and S.~Yeung, ``Holistic {3D} human and scene mesh estimation from
  single view images,'' in \emph{CVPR}, 2021, pp. 334--343.

\bibitem{zhang2020perceiving}
J.~Y. Zhang, S.~Pepose, H.~Joo, D.~Ramanan, J.~Malik, and A.~Kanazawa,
  ``Perceiving {3D} human-object spatial arrangements from a single image in
  the wild,'' in \emph{ECCV}.\hskip 1em plus 0.5em minus 0.4em\relax Springer,
  2020, pp. 34--51.

\bibitem{xie2022chore}
X.~Xie, B.~L. Bhatnagar, and G.~Pons-Moll, ``{CHORE}: Contact, human and object
  reconstruction from a single {RGB} image,'' in \emph{ECCV}.\hskip 1em plus
  0.5em minus 0.4em\relax Springer, 2022, pp. 125--145.

\bibitem{yi2022human}
H.~Yi, C.-H.~P. Huang, D.~Tzionas, M.~Kocabas, M.~Hassan, S.~Tang, J.~Thies,
  and M.~J. Black, ``Human-aware object placement for visual environment
  reconstruction,'' in \emph{CVPR}, 2022, pp. 3959--3970.

\bibitem{luo2022embodied}
Z.~Luo, S.~Iwase, Y.~Yuan, and K.~M. Kitani, ``Embodied scene-aware human pose
  estimation,'' in \emph{NeurIPS}, 2022.

\bibitem{shen2023learning}
Z.~Shen, Z.~Cen, S.~Peng, Q.~Shuai, H.~Bao, and X.~Zhou, ``Learning human mesh
  recovery in {3D} scenes,'' in \emph{CVPR}, 2023, pp. 17\,038--17\,047.

\bibitem{kissos2020beyond}
I.~Kissos, L.~Fritz, M.~Goldman, O.~Meir, E.~Oks, and M.~Kliger, ``Beyond weak
  perspective for monocular {3D} human pose estimation,'' in \emph{ECCV}.\hskip
  1em plus 0.5em minus 0.4em\relax Springer, 2020, pp. 541--554.

\bibitem{cho2021camera}
H.~Cho, Y.~Cho, J.~Yu, and J.~Kim, ``Camera distortion-aware {3D} human pose
  estimation in video with optimization-based meta-learning,'' in \emph{ICCV},
  2021, pp. 11\,169--11\,178.

\bibitem{Wang2023Zolly}
W.~Wang, Y.~Ge, H.~Mei, Z.~Cai, Q.~Sun, Y.~Wang, C.~Shen, L.~Yang, and
  T.~Komura, ``{Zolly}: Zoom focal length correctly for perspective-distorted
  human mesh reconstruction,'' in \emph{ICCV}, 2023, pp. 3925--3935.

\bibitem{gartner2022differentiable}
E.~G{\"a}rtner, M.~Andriluka, E.~Coumans, and C.~Sminchisescu, ``Differentiable
  dynamics for articulated {3D} human motion reconstruction,'' in \emph{CVPR},
  2022, pp. 13\,190--13\,200.

\bibitem{gartner2022trajectory}
E.~G{\"a}rtner, M.~Andriluka, H.~Xu, and C.~Sminchisescu, ``Trajectory
  optimization for physics-based reconstruction of {3D} human pose from
  monocular video,'' in \emph{CVPR}, 2022, pp. 13\,106--13\,115.

\bibitem{xie2021physics}
K.~Xie, T.~Wang, U.~Iqbal, Y.~Guo, S.~Fidler, and F.~Shkurti, ``Physics-based
  human motion estimation and synthesis from videos,'' in \emph{ICCV}, 2021,
  pp. 11\,532--11\,541.

\bibitem{huang2022neural}
B.~Huang, L.~Pan, Y.~Yang, J.~Ju, and Y.~Wang, ``{Neural MoCon}: Neural motion
  control for physically plausible human motion capture,'' in \emph{CVPR},
  2022, pp. 6417--6426.

\bibitem{tripathi20233d}
S.~Tripathi, L.~M{\"u}ller, C.-H.~P. Huang, O.~Taheri, M.~J. Black, and
  D.~Tzionas, ``{3D} human pose estimation via intuitive physics,'' in
  \emph{CVPR}, 2023, pp. 4713--4725.

\bibitem{fieraru2021learning}
M.~Fieraru, M.~Zanfir, E.~Oneata, A.-I. Popa, V.~Olaru, and C.~Sminchisescu,
  ``Learning complex {3D} human self-contact,'' in \emph{AAAI}, 2021.

\bibitem{teschner2005collision}
M.~Teschner, S.~Kimmerle, B.~Heidelberger, G.~Zachmann, L.~Raghupathi,
  A.~Fuhrmann, M.-P. Cani, F.~Faure, N.~Magnenat-Thalmann, W.~Strasser
  \emph{et~al.}, ``Collision detection for deformable objects,'' in \emph{CGF},
  vol.~24, no.~1.\hskip 1em plus 0.5em minus 0.4em\relax Wiley Online Library,
  2005, pp. 61--81.

\bibitem{zhang2023proxycap}
Y.~Zhang, H.~Zhang, L.~Hu, J.~Zhang, H.~Yi, S.~Zhang, and Y.~Liu, ``{ProxyCap}:
  Real-time monocular full-body capture in world space via human-centric
  proxy-to-motion learning,'' \emph{arXiv preprint arXiv:2307.01200}, 2023.

\bibitem{goodfellow2014generative}
I.~Goodfellow, J.~Pouget-Abadie, M.~Mirza, B.~Xu, D.~Warde-Farley, S.~Ozair,
  A.~Courville, and Y.~Bengio, ``Generative adversarial nets,'' \emph{NeurIPS},
  vol.~27, 2014.

\bibitem{kingma2013auto}
D.~P. Kingma and M.~Welling, ``Auto-encoding variational bayes,'' in
  \emph{ICLR}, 2014.

\bibitem{rezende2015variational}
D.~Rezende and S.~Mohamed, ``Variational inference with normalizing flows,'' in
  \emph{International conference on machine learning}.\hskip 1em plus 0.5em
  minus 0.4em\relax PMLR, 2015, pp. 1530--1538.

\bibitem{rong2022chasing}
Y.~Rong, Z.~Liu, and C.~C. Loy, ``Chasing the tail in monocular {3D} human
  reconstruction with prototype memory,'' \emph{TIP}, vol.~31, pp. 2907--2919,
  2022.

\bibitem{davydov2022adversarial}
A.~Davydov, A.~Remizova, V.~Constantin, S.~Honari, M.~Salzmann, and P.~Fua,
  ``Adversarial parametric pose prior,'' in \emph{CVPR}, 2022, pp.
  10\,997--11\,005.

\bibitem{dinh2016density}
L.~Dinh, J.~Sohl-Dickstein, and S.~Bengio, ``Density estimation using real
  {NVP},'' in \emph{ICLR}, 2017.

\bibitem{ho2020denoising}
J.~Ho, A.~Jain, and P.~Abbeel, ``Denoising diffusion probabilistic models,''
  \emph{NeurIPS}, vol.~33, pp. 6840--6851, 2020.

\bibitem{Foo2023Distribution}
L.~G. Foo, J.~Gong, H.~Rahmani, and J.~Liu, ``Distribution-aligned diffusion
  for human mesh recovery,'' in \emph{ICCV}, 2023, pp. 9221--9232.

\bibitem{Cho2023Generative}
H.~Cho and J.~Kim, ``Generative approach for probabilistic human mesh recovery
  using diffusion models,'' in \emph{ICCV Workshops}, 2023, pp. 4183--4188.

\bibitem{Zhang2023EgoHMR}
S.~Zhang, Q.~Ma, Y.~Zhang, S.~Aliakbarian, D.~Cosker, and S.~Tang,
  ``Probabilistic human mesh recovery in 3d scenes from egocentric views,'' in
  \emph{ICCV}, 2023, pp. 7989--8000.

\bibitem{kaufmann2020convolutional}
M.~Kaufmann, E.~Aksan, J.~Song, F.~Pece, R.~Ziegler, and O.~Hilliges,
  ``Convolutional autoencoders for human motion infilling,'' in
  \emph{3DV}.\hskip 1em plus 0.5em minus 0.4em\relax IEEE, 2020, pp. 918--927.

\bibitem{he2021challencap}
Y.~He, A.~Pang, X.~Chen, H.~Liang, M.~Wu, Y.~Ma, and L.~Xu, ``{ChallenCap}:
  Monocular {3D} capture of challenging human performances using multi-modal
  references,'' in \emph{CVPR}, 2021, pp. 11\,400--11\,411.

\bibitem{li2021task}
J.~Li, R.~Villegas, D.~Ceylan, J.~Yang, Z.~Kuang, H.~Li, and Y.~Zhao,
  ``Task-generic hierarchical human motion prior using {VAEs},'' in
  \emph{3DV}.\hskip 1em plus 0.5em minus 0.4em\relax IEEE, 2021, pp. 771--781.

\bibitem{xu2021exploring}
J.~Xu, M.~Wang, J.~Gong, W.~Liu, C.~Qian, Y.~Xie, and L.~Ma, ``Exploring
  versatile prior for human motion via motion frequency guidance,'' in
  \emph{3DV}.\hskip 1em plus 0.5em minus 0.4em\relax IEEE, 2021, pp. 606--616.

\bibitem{cai2021playing}
Z.~Cai, M.~Zhang, J.~Ren, C.~Wei, D.~Ren, J.~Li, Z.~Lin, H.~Zhao, S.~Yi,
  L.~Yang \emph{et~al.}, ``Playing for {3D} human recovery,'' \emph{arXiv
  preprint arXiv:2110.07588}, 2021.

\bibitem{patel2021agora}
P.~Patel, C.-H.~P. Huang, J.~Tesch, D.~T. Hoffmann, S.~Tripathi, and M.~J.
  Black, ``{AGORA}: Avatars in geography optimized for regression analysis,''
  in \emph{CVPR}, 2021, pp. 13\,468--13\,478.

\bibitem{yu2021function4d}
T.~Yu, Z.~Zheng, K.~Guo, P.~Liu, Q.~Dai, and Y.~Liu, ``{Function4D}: Real-time
  human volumetric capture from very sparse consumer {RGBD} sensors,'' in
  \emph{CVPR}, 2021, pp. 5746--5756.

\bibitem{bazavan2021hspace}
E.~G. Bazavan, A.~Zanfir, M.~Zanfir, W.~T. Freeman, R.~Sukthankar, and
  C.~Sminchisescu, ``{HSPACE}: Synthetic parametric humans animated in complex
  environments,'' \emph{arXiv preprint arXiv:2112.12867}, 2021.

\bibitem{black2023bedlam}
M.~J. Black, P.~Patel, J.~Tesch, and J.~Yang, ``{BEDLAM}: A synthetic dataset
  of bodies exhibiting detailed lifelike animated motion,'' in \emph{CVPR},
  2023, pp. 8726--8737.

\bibitem{trumble2017total}
M.~Trumble, A.~Gilbert, C.~Malleson, A.~Hilton, and J.~Collomosse, ``{Total
  Capture}: {3D} human pose estimation fusing video and inertial sensors,'' in
  \emph{BMVC}, 2017, pp. 1--13.

\bibitem{joo2015panoptic}
H.~Joo, H.~Liu, L.~Tan, L.~Gui, B.~Nabbe, I.~Matthews, T.~Kanade, S.~Nobuhara,
  and Y.~Sheikh, ``Panoptic studio: A massively multiview system for social
  motion capture,'' in \emph{ICCV}, 2015, pp. 3334--3342.

\bibitem{mehta2017monocular}
D.~Mehta, H.~Rhodin, D.~Casas, P.~Fua, O.~Sotnychenko, W.~Xu, and C.~Theobalt,
  ``Monocular {3D} human pose estimation in the wild using improved {CNN}
  supervision,'' in \emph{3DV}.\hskip 1em plus 0.5em minus 0.4em\relax IEEE,
  2017, pp. 506--516.

\bibitem{mehta2018single}
D.~Mehta, O.~Sotnychenko, F.~Mueller, W.~Xu, S.~Sridhar, G.~Pons-Moll, and
  C.~Theobalt, ``Single-shot multi-person {3D} pose estimation from monocular
  {RGB},'' in \emph{3DV}.\hskip 1em plus 0.5em minus 0.4em\relax IEEE, 2018,
  pp. 120--130.

\bibitem{li2019learning}
Z.~Li, T.~Dekel, F.~Cole, R.~Tucker, N.~Snavely, C.~Liu, and W.~T. Freeman,
  ``Learning the depths of moving people by watching frozen people,'' in
  \emph{CVPR}, 2019, pp. 4521--4530.

\bibitem{leroy2020smply}
V.~Leroy, P.~Weinzaepfel, R.~Br{\'e}gier, H.~Combaluzier, and G.~Rogez,
  ``{SMPLy} benchmarking {3D} human pose estimation in the wild,'' in
  \emph{3DV}.\hskip 1em plus 0.5em minus 0.4em\relax IEEE, 2020, pp. 301--310.

\bibitem{fang2021reconstructing}
Q.~Fang, Q.~Shuai, J.~Dong, H.~Bao, and X.~Zhou, ``Reconstructing {3D} human
  pose by watching humans in the mirror,'' in \emph{CVPR}, 2021, pp.
  12\,814--12\,823.

\bibitem{yu2020humbi}
Z.~Yu, J.~S. Yoon, I.~K. Lee, P.~Venkatesh, J.~Park, J.~Yu, and H.~S. Park,
  ``{HUMBI}: A large multiview dataset of human body expressions,'' in
  \emph{CVPR}, 2020, pp. 2990--3000.

\bibitem{zju3dv}
\BIBentryALTinterwordspacing
zju3dv, ``{EasyMoCap} - make human motion capture easier.'' GitHub, 2021.
  [Online]. Available: \url{https://github.com/zju3dv/EasyMocap}
\BIBentrySTDinterwordspacing

\bibitem{johnson2010clustered}
S.~Johnson and M.~Everingham, ``Clustered pose and nonlinear appearance models
  for human pose estimation,'' in \emph{BMVC}, 2010, pp. 12.1--12.11.

\bibitem{johnson2011learning}
------, ``Learning effective human pose estimation from inaccurate
  annotation,'' in \emph{CVPR}.\hskip 1em plus 0.5em minus 0.4em\relax IEEE,
  2011, pp. 1465--1472.

\bibitem{lin2014microsoft}
T.-Y. Lin, M.~Maire, S.~Belongie, J.~Hays, P.~Perona, D.~Ramanan,
  P.~Doll{\'a}r, and C.~L. Zitnick, ``Microsoft {COCO}: Common objects in
  context,'' in \emph{ECCV}.\hskip 1em plus 0.5em minus 0.4em\relax Springer,
  2014, pp. 740--755.

\bibitem{andriluka20142d}
M.~Andriluka, L.~Pishchulin, P.~Gehler, and B.~Schiele, ``{2D} human pose
  estimation: New benchmark and state of the art analysis,'' in \emph{CVPR},
  2014, pp. 3686--3693.

\bibitem{andriluka2018posetrack}
M.~Andriluka, U.~Iqbal, E.~Insafutdinov, L.~Pishchulin, A.~Milan, J.~Gall, and
  B.~Schiele, ``{PoseTrack}: A benchmark for human pose estimation and
  tracking,'' in \emph{CVPR}, 2018, pp. 5167--5176.

\bibitem{zhang2019pose2seg}
S.-H. Zhang, R.~Li, X.~Dong, P.~Rosin, Z.~Cai, X.~Han, D.~Yang, H.~Huang, and
  S.-M. Hu, ``{Pose2Seg}: Detection free human instance segmentation,'' in
  \emph{CVPR}, 2019, pp. 889--898.

\bibitem{moon2022neuralannot}
G.~Moon, H.~Choi, and K.~M. Lee, ``{NeuralAnnot}: Neural annotator for {3D}
  human mesh training sets,'' in \emph{CVPRW}, 2022, pp. 2299--2307.

\bibitem{saito2019pifu}
S.~Saito, Z.~Huang, R.~Natsume, S.~Morishima, A.~Kanazawa, and H.~Li, ``{PIFu}:
  Pixel-aligned implicit function for high-resolution clothed human
  digitization,'' in \emph{ICCV}, 2019, pp. 2304--2314.

\bibitem{Yang2023SynBody}
Z.~Yang, Z.~Cai, H.~Mei, S.~Liu, Z.~Chen, W.~Xiao, Y.~Wei, Z.~Qing, C.~Wei,
  B.~Dai, W.~Wu, C.~Qian, D.~Lin, Z.~Liu, and L.~Yang, ``{SynBody}: Synthetic
  dataset with layered human models for {3D} human perception and modeling,''
  in \emph{ICCV}, 2023, pp. 20\,282--20\,292.

\bibitem{paysan20093d}
P.~Paysan, R.~Knothe, B.~Amberg, S.~Romdhani, and T.~Vetter, ``A {3D} face
  model for pose and illumination invariant face recognition,'' in
  \emph{AVSS}.\hskip 1em plus 0.5em minus 0.4em\relax Ieee, 2009, pp. 296--301.

\bibitem{bhatnagar2022behave}
B.~L. Bhatnagar, X.~Xie, I.~A. Petrov, C.~Sminchisescu, C.~Theobalt, and
  G.~Pons-Moll, ``{BEHAVE}: Dataset and method for tracking human object
  interactions,'' in \emph{CVPR}, 2022, pp. 15\,935--15\,946.

\bibitem{taheri2020grab}
O.~Taheri, N.~Ghorbani, M.~J. Black, and D.~Tzionas, ``{GRAB}: A dataset of
  whole-body human grasping of objects,'' in \emph{ECCV}.\hskip 1em plus 0.5em
  minus 0.4em\relax Springer, 2020, pp. 581--600.

\bibitem{huang2022capturing}
C.-H.~P. Huang, H.~Yi, M.~H{\"o}schle, M.~Safroshkin, T.~Alexiadis,
  S.~Polikovsky, D.~Scharstein, and M.~J. Black, ``Capturing and inferring
  dense full-body human-scene contact,'' in \emph{CVPR}, 2022, pp.
  13\,274--13\,285.

\bibitem{dai2023sloper4d}
Y.~Dai, Y.~Lin, X.~Lin, C.~Wen, L.~Xu, H.~Yi, S.~Shen, Y.~Ma, and C.~Wang,
  ``{SLOPER4D}: A scene-aware dataset for global {4D} human pose estimation in
  urban environments,'' in \emph{CVPR}, 2023, pp. 682--692.

\bibitem{karpathy2014large}
A.~Karpathy, G.~Toderici, S.~Shetty, T.~Leung, R.~Sukthankar, and L.~Fei-Fei,
  ``Large-scale video classification with convolutional neural networks,'' in
  \emph{CVPR}, 2014, pp. 1725--1732.

\bibitem{lin2023motionx}
J.~Lin, A.~Zeng, S.~Lu, Y.~Cai, R.~Zhang, H.~Wang, and L.~Zhang, ``{Motion-X}:
  A large-scale {3D} expressive whole-body human motion dataset,''
  \emph{NeurIPS}, 2023.

\bibitem{zeng2021smoothnet}
A.~Zeng, L.~Yang, X.~Ju, J.~Li, J.~Wang, and Q.~Xu, ``{SmoothNet}: A
  plug-and-play network for refining human poses in videos,'' in
  \emph{ECCV}.\hskip 1em plus 0.5em minus 0.4em\relax Springer, 2022, pp.
  625--642.

\bibitem{zhang2020place}
S.~Zhang, Y.~Zhang, Q.~Ma, M.~J. Black, and S.~Tang, ``{PLACE}: Proximity
  learning of articulation and contact in {3D} environments,'' in
  \emph{3DV}.\hskip 1em plus 0.5em minus 0.4em\relax IEEE, 2020, pp. 642--651.

\bibitem{zhang2020generating}
Y.~Zhang, M.~Hassan, H.~Neumann, M.~J. Black, and S.~Tang, ``Generating {3D}
  people in scenes without people,'' in \emph{CVPR}, 2020, pp. 6194--6204.

\bibitem{liu20204d}
M.~Liu, D.~Yang, Y.~Zhang, Z.~Cui, J.~M. Rehg, and S.~Tang, ``{4D} human body
  capture from egocentric video via {3D} scene grounding,'' in
  \emph{3DV}.\hskip 1em plus 0.5em minus 0.4em\relax IEEE, 2021, pp. 930--939.

\bibitem{chen2019unsupervised}
C.-H. Chen, A.~Tyagi, A.~Agrawal, D.~Drover, S.~Stojanov, and J.~M. Rehg,
  ``Unsupervised {3D} pose estimation with geometric self-supervision,'' in
  \emph{CVPR}, 2019, pp. 5714--5724.

\bibitem{rhodin2018unsupervised}
H.~Rhodin, M.~Salzmann, and P.~Fua, ``Unsupervised geometry-aware
  representation for {3D} human pose estimation,'' in \emph{ECCV}, 2018, pp.
  750--767.

\bibitem{kundu2020appearance}
J.~N. Kundu, M.~Rakesh, V.~Jampani, R.~M. Venkatesh, and R.~Venkatesh~Babu,
  ``Appearance consensus driven self-supervised human mesh recovery,'' in
  \emph{ECCV}.\hskip 1em plus 0.5em minus 0.4em\relax Springer, 2020, pp.
  794--812.

\bibitem{zheng2009associating}
W.-S. Zheng, S.~Gong, and T.~Xiang, ``Associating groups of people,'' in
  \emph{BMVC}, vol.~2, no.~6, 2009, pp. 1--11.

\bibitem{lisanti2017group}
G.~Lisanti, N.~Martinel, A.~Del~Bimbo, and G.~Luca~Foresti, ``Group
  re-identification via unsupervised transfer of sparse features encoding,'' in
  \emph{ICCV}, 2017, pp. 2449--2458.

\bibitem{xu2018monoperfcap}
W.~Xu, A.~Chatterjee, M.~Zollh{\"o}fer, H.~Rhodin, D.~Mehta, H.-P. Seidel, and
  C.~Theobalt, ``{MonoPerfCap}: Human performance capture from monocular
  video,'' \emph{TOG}, vol.~37, no.~2, pp. 1--15, 2018.

\bibitem{ma2020cape}
Q.~Ma, J.~Yang, A.~Ranjan, S.~Pujades, G.~Pons{-}Moll, S.~Tang, and M.~J.
  Black, ``Learning to dress {3D} people in generative clothing,'' in
  \emph{CVPR}, 2020, pp. 6468--6477.

\bibitem{zhu2019detailed}
H.~Zhu, X.~Zuo, S.~Wang, X.~Cao, and R.~Yang, ``Detailed human shape estimation
  from a single image by hierarchical mesh deformation,'' in \emph{CVPR}, 2019,
  pp. 4491--4500.

\bibitem{ma2021scale}
Q.~Ma, S.~Saito, J.~Yang, S.~Tang, and M.~J. Black, ``{SCALE}: Modeling clothed
  humans with a surface codec of articulated local elements,'' in \emph{CVPR},
  2021, pp. 16\,082--16\,093.

\bibitem{lin2022learning}
S.~Lin, H.~Zhang, Z.~Zheng, R.~Shao, and Y.~Liu, ``Learning implicit templates
  for point-based clothed human modeling,'' in \emph{ECCV}.\hskip 1em plus
  0.5em minus 0.4em\relax Springer, 2022, pp. 210--228.

\bibitem{zhang2023closet}
H.~Zhang, S.~Lin, R.~Shao, Y.~Zhang, Z.~Zheng, H.~Huang, Y.~Guo, and Y.~Liu,
  ``{CloSET}: Modeling clothed humans on continuous surface with explicit
  template decomposition,'' in \emph{CVPR}, 2023, pp. 501--511.

\bibitem{li2020robust}
Z.~Li, T.~Yu, C.~Pan, Z.~Zheng, and Y.~Liu, ``Robust {3D} self-portraits in
  seconds,'' in \emph{CVPR}, 2020, pp. 1344--1353.

\bibitem{alldieck2022photorealistic}
T.~Alldieck, M.~Zanfir, and C.~Sminchisescu, ``Photorealistic monocular {3D}
  reconstruction of humans wearing clothing,'' in \emph{CVPR}, 2022, pp.
  1506--1515.

\bibitem{bhatnagar2020combining}
B.~L. Bhatnagar, C.~Sminchisescu, C.~Theobalt, and G.~Pons-Moll, ``Combining
  implicit function learning and parametric models for {3D} human
  reconstruction,'' in \emph{ECCV}.\hskip 1em plus 0.5em minus 0.4em\relax
  Springer, 2020, pp. 311--329.

\bibitem{shao2022dbfield}
R.~Shao, H.~Zhang, H.~Zhang, M.~Chen, Y.~Cao, T.~Yu, and Y.~Liu,
  ``{DoubleField}: Bridging the neural surface and radiance fields for
  high-fidelity human reconstruction and rendering,'' in \emph{CVPR}, 2022.

\bibitem{Feng2022scarf}
Y.~Feng, J.~Yang, M.~Pollefeys, M.~J. Black, and T.~Bolkart, ``{SCARF}:
  Capturing and animation of body and clothing from monocular video,'' in
  \emph{SIGGRAPH Asia Conference Papers}, 2022, p.~9.

\bibitem{moon20223d}
G.~Moon, H.~Nam, T.~Shiratori, and K.~M. Lee, ``{3D} clothed human
  reconstruction in the wild,'' in \emph{ECCV}.\hskip 1em plus 0.5em minus
  0.4em\relax Springer, 2022, pp. 184--200.

\end{thebibliography}

\end{document}